\documentclass{article}

     \usepackage[nonatbib, preprint]{neurips_2020}

\usepackage[utf8]{inputenc} 
\usepackage[T1]{fontenc}    
\usepackage{hyperref}       
\usepackage{url}            
\usepackage{booktabs}       
\usepackage{amsfonts}       
\usepackage{nicefrac}       
\usepackage{microtype}      
\usepackage{graphicx}
\usepackage{mathrsfs}
\usepackage{bm}
\usepackage{bbm}
\usepackage{upgreek}
\usepackage{xcolor}
\usepackage{wrapfig}
\usepackage{lipsum} 

\usepackage{amsmath}
\usepackage{amssymb}
\usepackage{amsthm}

\usepackage[round]{natbib}
\usepackage[title]{appendix}

\usepackage{tabularx}
\newcommand\setrow[1]{\gdef\rowmac{#1}#1\ignorespaces}
\newcommand\clearrow{\global\let\rowmac\relax}
\clearrow

\usepackage{subcaption}

\newtheorem{theorem}{Theorem}

\newtheorem{definition}{Definition}

\newtheorem{assumption}{Assumption}

\title{Why to “grow” and “harvest” deep learning models?}

\author{%
 Ilona Kulikovskikh\\
  Information Systems and Technologies Department\\
  Samara University, Samara, Russia \\
  \texttt{kulikovskikh.im@ssau.ru} \\
   \And
   Tarzan Legovi\'c \\
   Data Analysis Laboratory\\
   Institute of Applied Ecology, Oikon Ltd.
   Zagreb, Croatia \\
   \texttt{tlegovic@oikon.hr} \\
}

\begin{document}

\maketitle

\begin{abstract}
	Current expectations from training deep learning models with gradient-based methods include: 1) transparency; 2) high convergence rates; 3) high inductive biases. While the state-of-art methods with adaptive learning rate schedules are fast, they still fail to meet the other two requirements. We suggest reconsidering neural network models in terms of single-species population dynamics where adaptation comes naturally from open-ended processes of ``growth'' and ``harvesting''. We show that the stochastic gradient descent (SGD) with two balanced pre-defined values of \emph{per} capita growth and  harvesting rates outperform the most common adaptive gradient methods in all of the three requirements.
	
\end{abstract}

\section{Introduction}

Deep learning models imitate signal transmission within neurons in the brain with units which are interconnected through weighted links and assembled in layers. This connectionist approach to building the models establishes a general mathematical framework for their simple and effective implementation in parallel and distributed settings~\citep{markus2001}.  
On the one hand, a simplified model of the brain in the form of neural networks makes them popular due to its successful implementation in a wide range of real-world applications \citep{goodfellow2016, lecun2015}. On the other hand, this beneficial simplification requires an enormous number of units and layers to represent, process, and store data. This results in overparametrization that \textbf{makes deep neural networks difficult to interpret}.

Training neural networks implies modifying the weights of connections   according to some learning algorithms. Using gradient methods as such algorithms introduces another oversimplification of the processes in the brain.
One of the main issues in training the models with gradient-based methods is the highest convergence rate to the solution. In addition, it is desirable to guarantee a high inductive bias.
The state-of-art iterative schemes with adaptive learning rate schedules \textbf{converge faster but lead to lower inductive bias} ~\citep{gunasekar2018, hoffer2017, liuC2020, liuL2020, wilson2017, kim2017, arora2020}.  Overparameterization in non-adaptive methods implicitly \textbf{accelerates the training of deep networks}~\citep{allen-zhu2019, arora2018, li2018} \textbf{but, again, heavily reduces their transparency}.

We reconsider neural networks in terms of single-species population dynamics to  break this vicious circle. Even though the connectionist model of the brain seems limited to fully represent the dynamics of populations of highly interconnected units, it brings several benefits. 

\begin{wrapfigure}{R}{0.55\textwidth}
	\vspace{-2mm}
	\centering
	\includegraphics[width=0.55\columnwidth]{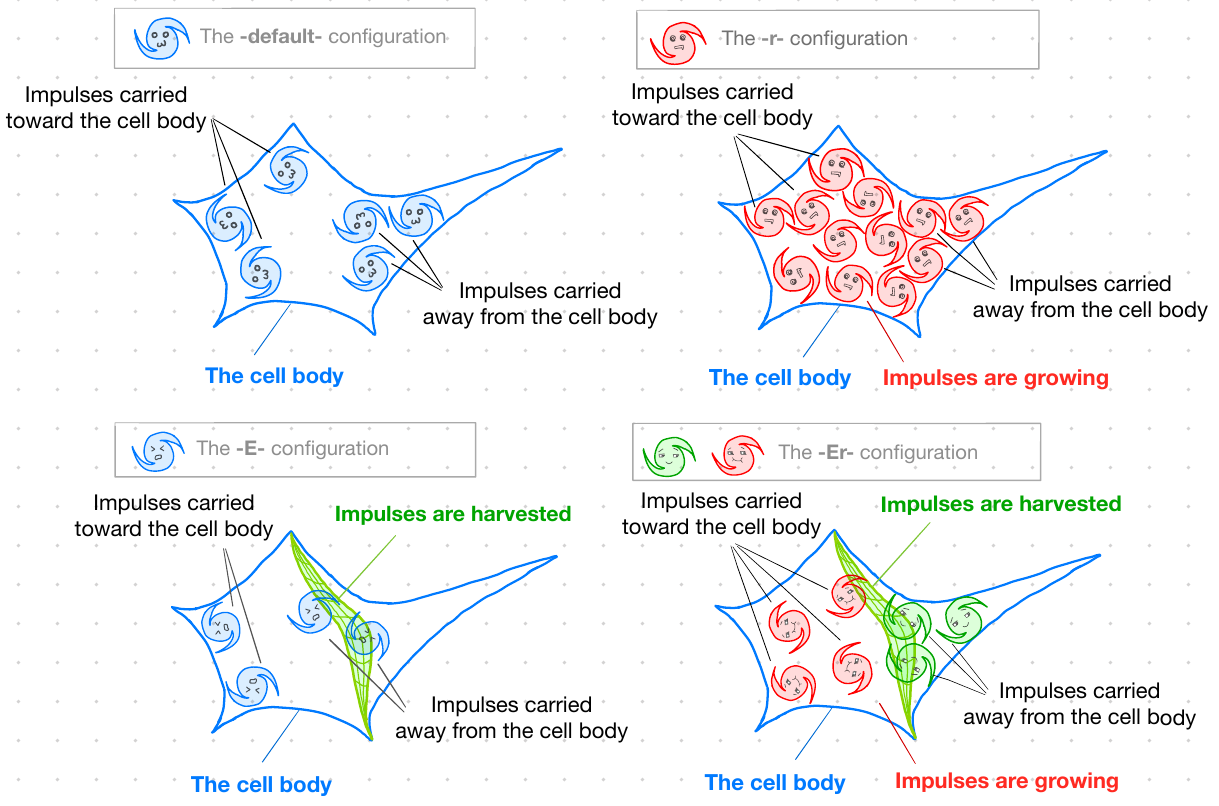}
	\caption{The LIGHT neuron}
	\label{light_neuron}
	\vspace{-5mm}
\end{wrapfigure}

\vspace{-3mm}
\paragraph{Deep learning theory} Integrating deep learning and population dynamics complements the state-of-art perspective of how neurons receive electrical impulses from other cells, accumulate them, and generate an action potential spike if a threshold value is exceeded. We introduce a new function, the LIGHT (\textbf{L}og\textbf{I}stic \textbf{G}rowth with \textbf{H}arves\textbf{T}ing), to model inner processes inside neurons with four different configurations (see Figure \ref{light_neuron}): 1) the \textbf{-default-} configuration where the LIGHT function reduces to a sigmoid activation function; 2) the \textbf{-r-} configuration where impulses are growing with a constant \emph{per} capita rate $r$; 3) the \textbf{-E-} configuration where impulses are harvested with a constant \emph{per} capita rate $E$; 4) the \textbf{-Er-} configuration where impulses are growing and harvested with constant \emph{per} capita rates $r$ and $E$ simultaneously.

In population dynamics, the effect of harvesting is one of the major concerns \citep{brauer2012,legovic2016}. Harvesting represents the reduction of the population due to hunting or capturing individuals, which removes them from the population. It holds the potential to preclude the possibility of overshoot when the population temporarily exceeds its long-term carrying capacity - the maximum population size of the individuals that the environment can sustain indefinitely, given the food, water, and other necessities available in the environment. 
\begin{center}
	\emph{ Why to ``grow'' and ``harvest'' deep learning models?}
\end{center}

\vspace{-3mm}
\paragraph{Transparency} 
Drawing the similarity between overparametrized models and overloaded population, we show that the LIGHT function with the -Er- configuration can guarantee a higher convergence rate and inductive bias with the reduced complexity of neural networks in comparison with the -default- configuration.  
 
\begin{wrapfigure}{R}{0.55\textwidth}
	\vspace{-15mm}
	\begin{subfigure}{.27\textwidth}
		\centering
		\includegraphics[width=0.98\linewidth]{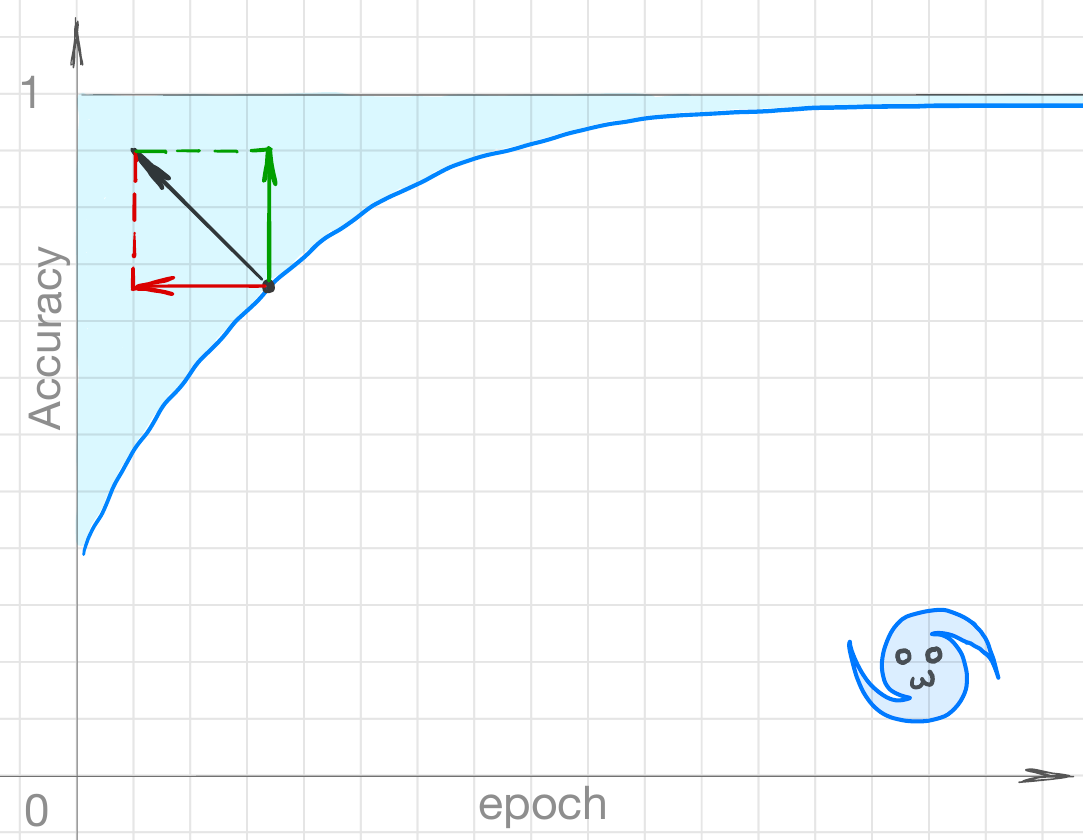}  
		\caption{-default-}
		\label{fig:def}
	\end{subfigure}
	\begin{subfigure}{.27\textwidth}
		\centering
		\includegraphics[width=0.98\linewidth]{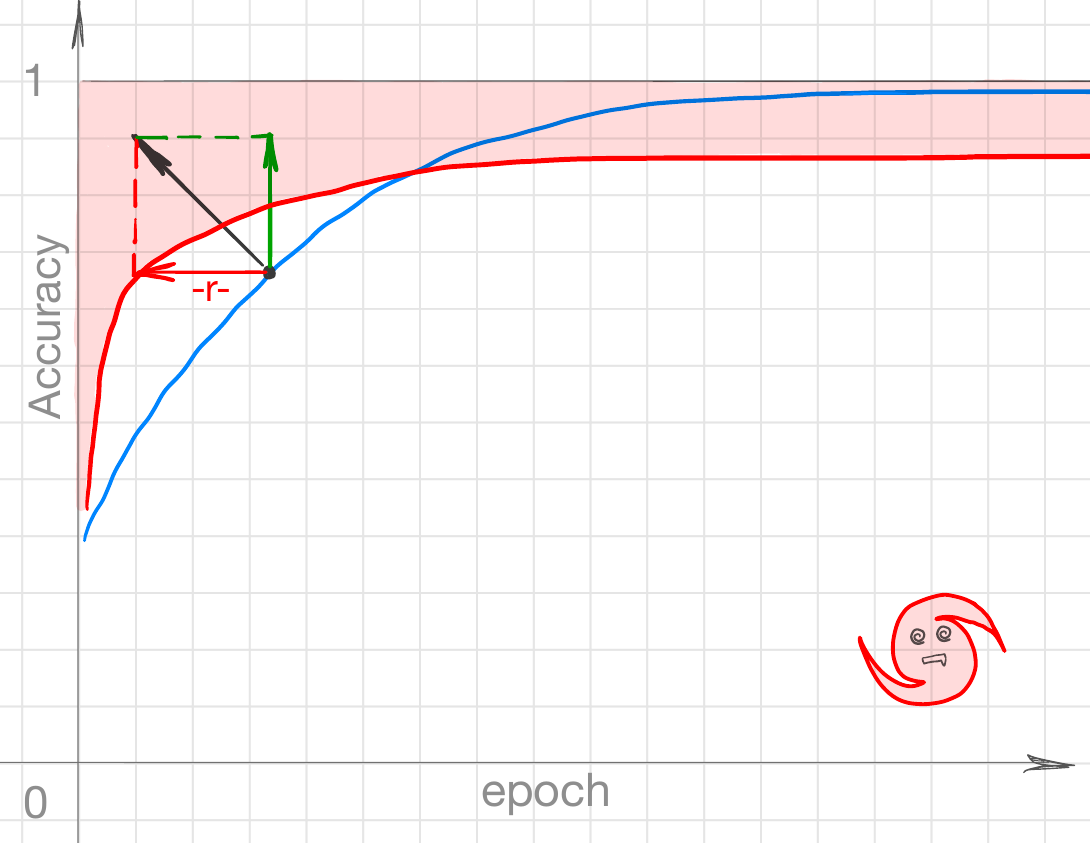}  
		\caption{-r-}
		\label{fig:r}
	\end{subfigure}
	\hfill
	\begin{subfigure}{.27\textwidth}
		\centering
		\includegraphics[width=0.98\linewidth]{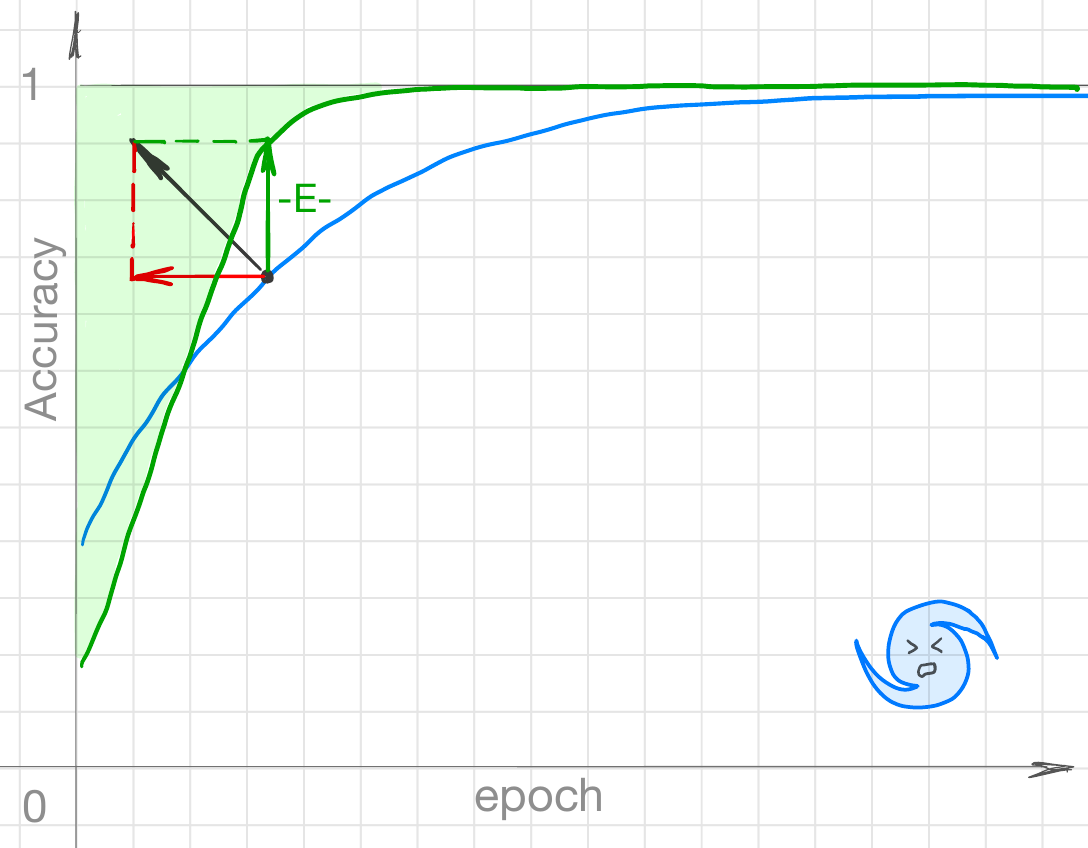}  
		\caption{-E- }
		\label{fig:E}
	\end{subfigure}
	\begin{subfigure}{.27\textwidth}
		\centering
		\includegraphics[width=0.98\linewidth]{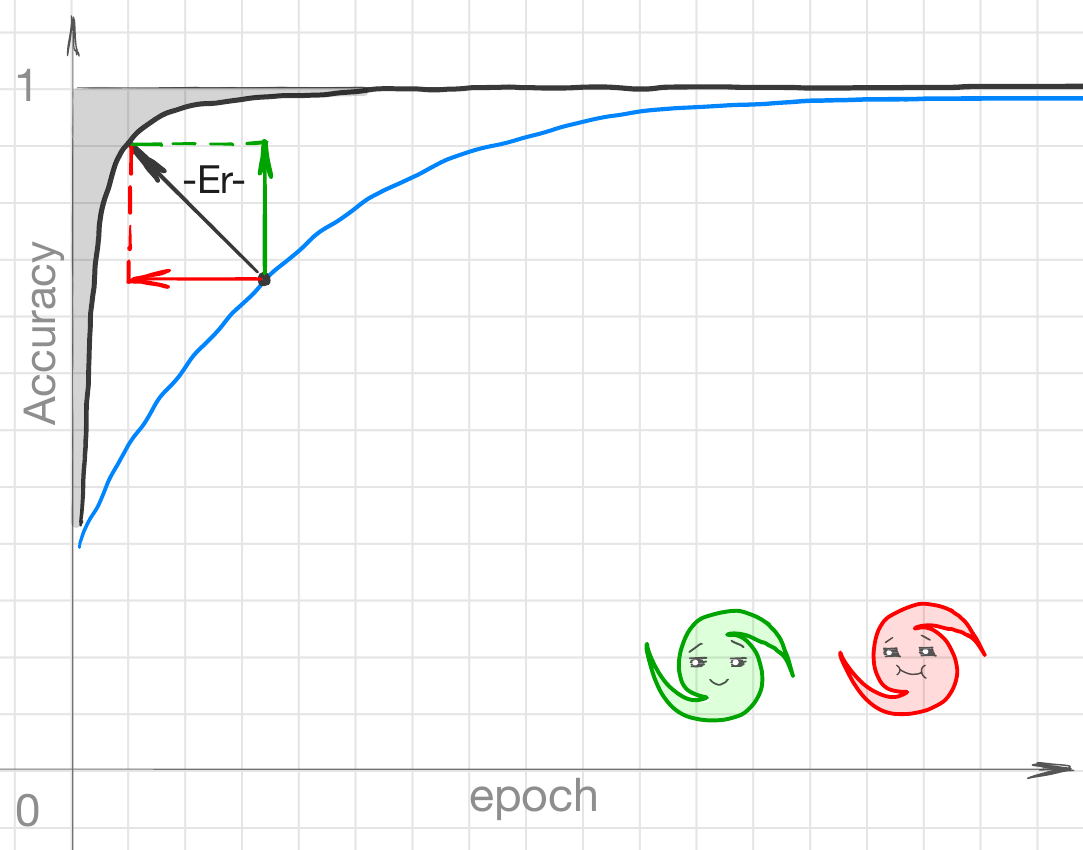}  
		\caption{-Er-}
		\label{fig:Er}
	\end{subfigure}
	\caption{Different configurations of the LIGHT neuron}
	\label{fig:config}
	\vspace{-20mm}
\end{wrapfigure}

 \vspace{-3mm} 
\paragraph{Convergence rate}  
The -default- configuration requires network architectures that reduce the area under an accuracy learning curve by squeezing it to the left horizontally and to the top vertically (see Figure \ref{fig:config} (a)). The -r- configuration adopts a pre-defined \emph{per} capita growth rate $r$ that results in a  higher convergence rate but lower inductive bias by squeezing the curve horizontally (see Figure \ref{fig:config} (b)).

\vspace{-3mm}
\paragraph{Inductive bias}
The -E- configuration includes a pre-defined \emph{per} capita harvesting rate $E$ that leads to a lower convergence rate but higher inductive bias by squeezing the curve vertically  (see Figure \ref{fig:config} (c)). The -Er- configuration involves two balanced pre-defined values of \emph{per} capita growth and  harvesting rates which increase both convergence rate and inductive bias (see Figure \ref{fig:config} (d)).

\section{Preliminaries}

We consider a dataset $\{\mathrm{x}_i,y_i\}_{i=1}^m$ with $\mathrm{x}_i\in\mathrm{R}^n$, $y_i\in\{-1,1\}$ and minimize an empirical loss function 
\begin{equation}
\label{eq::01}
\mathscr{L}(\bm{\uptheta}) = \sum_{i=1}^m\ell(y_i\bm{\uptheta}^\mathrm{T}\mathrm{x}_i)
\end{equation}
with a weight vector $\bm{\uptheta}\in\mathrm{R}^n$, $\ell$ is a smooth monotone strictly decreasing, $\beta$-smooth, and non-negative loss function.

We minimize \eqref{eq::01} using gradient descent (GD) with a fixed learning rate $\eta$:
\begin{multline}
\label{eq::02}
\bm{\uptheta}(t+1) = \bm{\uptheta}(t)-\eta\nabla\mathscr{L}(\bm{\uptheta}(t)) = \\
\bm{\uptheta}(t) -\eta\sum_{i=1}^m\ell^{\prime}(\bm{\uptheta}(t)^\mathrm{T}\mathrm{x}_i)\mathrm{x}_i.
\end{multline}

For the case of fully connected multi-layer linear networks, the equation \eqref{eq::02} can be presented as:
\begin{equation}
\label{eq::01mlp}
\bm{\Uptheta}_l(t+1) = \bm{\Uptheta}_l(t)-\eta\nabla_{\uptheta_l}\mathcal{L} (\Theta), \hspace{5mm} 
\mathcal{L} (\Theta) = \sum_{i=1}^m\ell (
\left\langle \Pi(\Theta),\mathrm{x}_i\right\rangle),
\end{equation}
where  $\Pi(\Theta) = \bm{\Uptheta}_1\times\bm{\Uptheta}_2\times\dots\times\bm{\Uptheta}_L$, 
$\Theta = \{\bm{\Uptheta}_l\in\mathrm{R}^{d_{l-1}\times d_l}: l = 1, 2,  \dots, L\}$, $L$ in the number of layers, $d_l$ is the number of nodes in the layer $l$.

In the stochastic setting, GD updates \eqref{eq::01mlp} for each mini-batch dataset $B(t)\subseteq \{1, \dots, m\}$ as:
\begin{equation}
\label{eq::03mlp}
\mathcal{L} (\Theta) = \sum_{i\in B(t)}\ell(\left\langle \Pi(\Theta),\mathrm{x}_i\right\rangle).
\end{equation}

We are particularly interested in modeling the loss/activation function $\ell$ of the last classification layer with population dynamics.

\section{Population growth with harvesting}

In the theory of natural selection, populations with unlimited natural resources grow exponentially. Exponential growth may occur in environments where there are few individuals and plentiful resources, but when the number of individuals becomes large enough, resources become depleted, slowing the growth rate. Eventually, the growth rate stops at the population size that a particular environment can support, which is called the carrying capacity. This scenario includes Verhulst and Gompertz population growth. Harvesting could be considered as an efficient way of maintaining the growth rate while ensuring a sustainable population size. 

\subsection{Verhulst model}
\label{verh}
Let us consider the population $N(t)$ which grows according to the Verhulst's logistic law ~\citep{verhulst1838}. In addition, we impose a constant \emph{per} capita harvesting rate $E$ according to the harvesting strategy where a harvesting rate $H(t)$ is proportional to the number of individuals present. This type of harvesting is called \emph{proportional}  ~\citep{legovic2016, schaefer1954}:
\begin{equation}
\scalebox{0.99}[1]{$
\frac{\mathrm{d}N(t)}{\mathrm{d}t} = rN(t)\left(1-\frac{N(t)}{K}\right)-H(t),
	$}
\label{eqlog:1}
\end{equation}
where harvesting starts at time $T$ with the rate $H(t) = EN(t)$, $r$ is the \emph{per} capita rate of population growth, $K$ is the carrying capacity which stands for the maximum sustainable size of the population. The solution to \eqref{eqlog:1} is:
\begin{equation}
\label{eqlog:1s}
\scalebox{0.99}[1]{$
N(t) = \displaystyle\frac{s}{1-(1-\frac{s}{N_T})\mathrm{e}^{-(r-E)(t-T)}},
	$}
\end{equation}    
where $s = K\left(1-\frac{E}{r}\right)$, $N_T$ is the size of the population at the time $T$.

If $H(t) = 0$, the equation \eqref{eqlog:1} has two equilibria at $N^* = 0$ and $N^* = K$, where $K$ defines the non-extinction equilibrium point. While the first equilibrium is unstable, the second one is asymptotically stable $\forall E\in[0,r)$. As $E$ increases from zero to $r$, the equilibrium decreases from $K$ to zero. 

If $H(t) > 0$, the harvested model has the equilibria at $N^* = 0$ and $N^* = K\left(1-\frac{E}{r}\right)$. For a given $E$, the value $H = EN^*$ defines the harvesting rate, which attains the maximum $H^* = \frac{rK}{4}$ for $N^* = \frac{K}{2}$ and $E^* = \frac{r}{2}$. 
When the harvesting persists at $H^* > \frac{rK}{4}$, the two equilibria points become one, $N^* = 0$, and the entire system collapses \citep{brauer2012,legovic2016}.

\subsection{Gompertz model}

\label{gomp}
The Gompertz equation is an alternative to the logistic growth model that has been successfully used to describe the growth of animals, plants,  bacteria, and cancer cells~\citep{gompertz1825,tjorve2017,winsor1932}. With the proportional harvesting, the Gompertz growth becomes: 
\begin{equation}
\scalebox{0.99}[1]{$
\frac{\mathrm{d}N(t)}{\mathrm{d}t} =
	rN(t)\ln\left(\frac{K}{N(t)}\right)-H(t). 
	$}
\label{eqlog:2}
\end{equation}
This equation with $H(t) = 0$ is a special case of the Richards model, and, thus,  belongs to the Richards family of sigmoidal growth models ~\citep{tjorve2017}.
The solution is equal to:
\begin{equation}
\label{eqlog:2s}
\scalebox{0.99}[1]{$
N(t) = 
K\mathrm{e}^{-\frac{E}{r} + s\mathrm{e}^{-r(t-T)}}
	$}
\end{equation}
where $s = \ln\left(\frac{N_T}{K}\right)+\frac{E}{r}$.
If $H(t) = 0$, the equation \eqref{eqlog:2} has the same equilibria at $N^* = 0$ and $N^* = K$ as the equation \eqref{eqlog:1}. If $H(t) > 0$, the second equilibrium is different $N^* = \frac{K}{\exp\left(\frac{E}{r}\right)}$. Then, the maximum harvesting rate $H^* = \frac{rK}{e}$ for $N^* = \frac{K}{e}$ and $E^* = r$.


\section{LIGHT}

\subsection{Definition}

Reconsidering the models of logistic population growth with harvesting for real values of $t$:
$\lim_{t\rightarrow-\infty}\ell(t; r, E, K, T, N_T) = 0,$
$\lim_{t\rightarrow\infty}\ell (t; r, E, K, T, N_T) = K,$
where $K \leq 1$, we propose two versions of the LIGHT function $\ell(t; r, E, K, T, N_T)$:\\
-- \textbf{LIGHT-V}: the function of growth by the Verhulst law $\ell^{\mathrm{V}}(t; r, E, K, T, N_T)$ based on \eqref{eqlog:1s};\\
-- \textbf{LIGHT-G}: the function of growth by the Gompertz law $\ell^{\mathrm{G}}(t; r, E, K, T, N_T)$  based on \eqref{eqlog:2s}.

\begin{figure}[h!]
	\begin{center}
		\centering
		\includegraphics[width=\columnwidth]{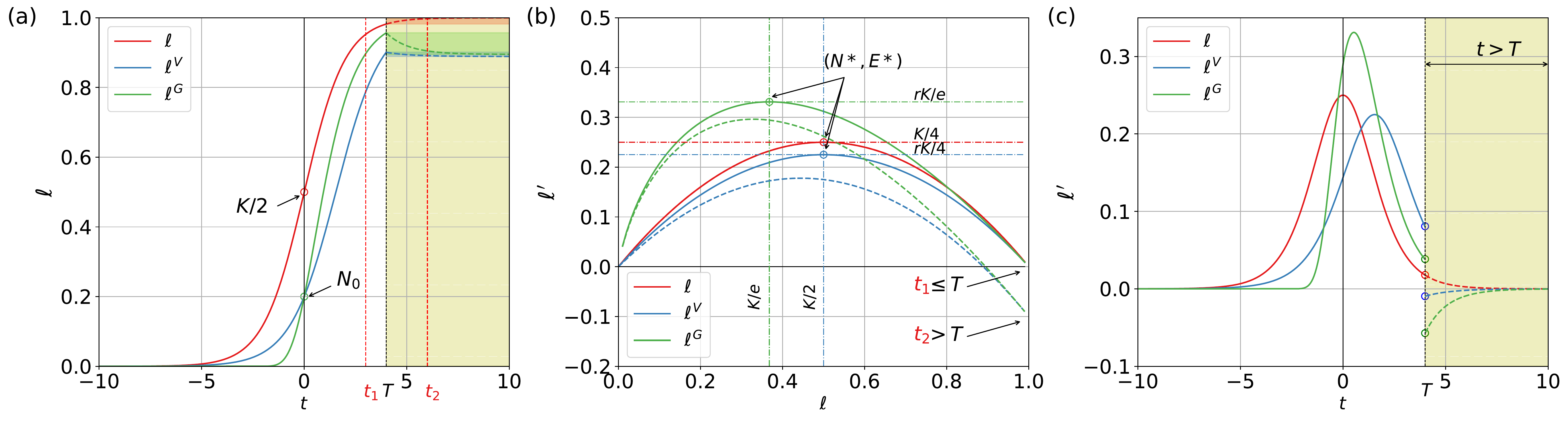}
		\caption{The LIGHT neuron starts harvesting at the fixed time $T = 4$ with two timestamps before $t_1 = 3$ and after $t_2 = 6$: $r = 1$, $E = 0$, $K = 1$, $T = 0$, $N_T = 1/2$ ($\ell$: colored in red);  $r = 0.9$, $E = 0.1$, $K=1$, $T = 4$, $N_T = 0.2$ ($\ell^{\mathrm{V}}$: colored in blue, $\ell^{\mathrm{G}}$: colored in green): (a) $\ell$ over $t$ ; (b) $\ell^{\prime}(\ell)$; (c) $\ell^{\prime}(t)$.}
		\label{fig1}
	\end{center}
\end{figure}

Each of the versions builds the LIGHT neuron with four configurations (see Figure \ref{fig:config}):  
\textbf{-default-}: $r = 1$ and $E = 0$; 
\textbf{-r-}: $r = \rm{const}$ and $E = 0$; 
\textbf{-E-}: $r = 1$ and $E = \rm{const}$; 
\textbf{-Er-}: $r = \rm{const}$ and $E = \rm{const}$.  
For the first configuration, if $K = 1$, $T = 0$, $N_T = 1/2$, the LIGHT function reduces to the sigmoid function (see Figure~\ref{fig1} (a), $\ell$: colored in red). For the last three configurations, $K = 1$ and $T, N_T = \rm{const}$ (see Figure~\ref{fig1} (a), $\ell^{\mathrm{V}}$: colored in blue, $\ell^{\mathrm{G}}$: colored in green). Figure~\ref{fig1} (b) provide an extra interpretation on the equilibria points for LIGHT-V  (see Section \ref{verh}) and LIGHT-G (see Section \ref{gomp}). Figure~\ref{fig1} (c) depicts the derivatives we adopt to modify \eqref{eq::02}. 

The key difference between LIGHT-V and LIGHT-G is that the latter grows faster when the population of impulses in the LIGHT neuron is smaller. But, the same as LIGHT-V, LIGHT-G drives the growth rate to zero when the capacity $K$ is approached. Moving away from the carrying capacity, harvesting accelerates the population growth.

Taking into account a generalizing parameter $q$, responsible for how fast the population grows at smaller population, we extend a traditional mathematical framework with quantum calculus~\citep{ernst2003, jackson1908, tsallis1988,  tsallis1994}. Q-calculus is equivalent to traditional infinitesimal calculus but without the concept of limits.

Combining two definitions $\ell^{\mathrm{V}}(t; r, E, K, T, N_T)$ and $\ell^{\mathrm{G}}(t; r, E, K, T, N_T)$, we propose a generalized function $\ell (t;  r, E, K, T, N_T, q)$. 

\begin{definition}
	\label{def::1}
	The generalized LIGHT function is equal to: 
	\begin{equation}
	\label{eqlog:4}
	\ell (t; r, E, K, T, N_T, q) = 
	\varepsilon K\mathrm{e}^{\left(\ln_q\left(\frac{N_T}{K}\right)+\frac{E}{r}\right)\mathrm{e}^{-r(t-T)}},
	\end{equation}
	where $\ln_q(x)$ is the \emph{q}-logarithm, $q$ is a rate with which the population grows when smaller, $\varepsilon$ is the extent to which the \emph{per} capita growth rate $r$ is impacted by the \emph{per} capita harvesting rate $E$.
\end{definition}
The detailed derivation of the equation \eqref{eqlog:4} is given in Appendix \ref{proof::def}.

From Definition \ref{def::1}, 
$\ell (t; r, E, K, T, N_T, q) = \ell^{\mathrm{V}}(t; r, E, K, T, N_T)$ if $q = 1$;  $\ell (t; r, E, K, T, N_T, q)\\ = \ell^{\mathrm{G}}(t; r, E, K, T, N_T)$ with regard to $\lim_{q\rightarrow \infty}\ln_q(x) =\ln(x)$. 

\vspace{-3mm}
\paragraph{Remark} 
While logistic growth models resemble the sigmoid function, the LIGHT function is not restricted to this similarity. In a certain space of the hyperparameters, LIGHT can present the smooth versions of ReLU such as SiLU or ELU if $K\rightarrow\infty$ and Swish if $K\rightarrow\infty$ and harvesting starts when the population size is minimal. For the latter, LIGHT also needs redefining exponents with q-calculus to $\exp_q(x)$.

\subsection{Convergence rate analysis}

\citet{soudry2018} disclosed a spectacular feature of gradient descent on separable data in both the default (GD) and stochastic (SGD) settings. The rate of convergence of a loss function with a fixed step size is linear $\mathcal{O}\left(\frac{1}{t}\right)$ while the rate of convergence to $L_2$ maximum margin is only logarithmic $\mathcal{O}\left(\frac{1}{\ln t}\right)$ in the number of iterations on not degenerate datasets. \citet{allen-zhu2019} proved that SGD can find global minima on the training objective in polynomial time under similar assumptions. 

We intend to analyze how much growing and harvesting contribute to accelerating the reported convergence rate in the setting given by \citet{soudry2018} (see Assumption \ref{ass::1}, \ref{ass::2}).

\begin{assumption}
	\label{ass::1}
	The dataset is linearly separable: $\exists\hspace{1mm}\bm{\uptheta^*}$ such that $\forall i: y_i\bm{\uptheta}^{*\mathrm{T}}\mathrm{x}_i > 0$.
\end{assumption}
\begin{assumption}
	\label{ass::2}
	$\forall t\in\mathrm{R}$: $\ell(t)$ is a differentiable, monotonically decreasing function bounded from below: $\ell(t) > 0$, $\ell^{\prime}(t)<0$, $\lim_{t\rightarrow -\infty}\ell^{\prime}(t) \neq 0$, $\lim_{t\rightarrow\infty}\ell(t) = \lim_{t\rightarrow\infty}\ell^{\prime}(t) = 0$, and its derivative is $\beta$-Lipschitz: 
	$\ell(t')\leq \ell(t) + \langle\nabla\ell(t),t'-t\rangle + \frac{\beta}{2}\|t'-t\|^2,\hspace{3mm} \beta > 0$ such that  $-\ell^{\prime}(t) = \exp(-f(t))$, where $f(t) = \omega(\ln(t))$ \emph{\citep{nacson2019}}
\end{assumption}

For the sake of simplicity, we also assume that $\forall i\in\{1,...,m\}: y_i = 1, \|x_i\| < 1$. 

Using Definition \ref{def::1}, we can rewrite the updates of GD in \eqref{eq::02} as follows:
\begin{equation}
\label{eqlog:3}
\bm{\uptheta}(t+1) = 
\bm{\uptheta}(t) -\eta\sum_{i=1}^m\ell^{\prime}(\bm{\uptheta}(t)^\mathrm{T}\mathrm{x}_i; r, E, K, T, N_T, q)\mathrm{x}_i,
\end{equation}
\begin{equation}
\label{eqlog:5}
\ell^{\prime}(t; r, E, K, T, N_T, q) = 
-\varepsilon Kr\left(\ln_q\left(\frac{N_T}{K}\right)+\frac{E}{r}\right)\mathrm{e}^{-r(t-T)+\left(\frac{E}{r}+\ln_q\left(\frac{N_T}{K}\right)\right)\mathrm{e}^{-rt}-\frac{E}{r}}.
\end{equation}

\begin{theorem}
	\label{thm::1}
	Let the LIGHT neuron in the layer $d_L$ of a network with a number of layers $L=1$ start harvesting the population of impulses, which grow from $N_T$ to $K$ with the \emph{per} capita rate $r$, at the time $T$ with the \emph{per} capita rate $E$. For any dataset (Assumption \ref{ass::1}), any configuration of the LIGHT neuron (Assumption \ref{ass::2}) the updates of gradient descent \eqref{eqlog:3} at any starting point $\bm{\uptheta}_0$ converge towards the max margin with a fixed step size $\eta < \frac{2}{\beta}$ as $\min_i\frac{\bm{\uptheta}_t^\mathrm{T}\mathrm{x}_i}{\|\bm{\uptheta}_t\|} = d - \mathcal{O}\left(\frac{1}{g(t)}\right)$, 
	$d = \max_{\bm{\uptheta}}\min_i\frac{\bm{\uptheta}^\mathrm{T}\mathrm{x}_i}{\|\bm{\uptheta}\|} = \frac{1}{\hat{\bm{\uptheta}}}$ with the rate:
	\begin{equation}
	\label{eq::thm}
	g(t) = \frac{E}{r^2}+T+\frac{\ln(t)+W_0\left(\left(\frac{E}{r} + \ln_q\left(\frac{N_T}{K}\right)\right)\mathrm{e}^{-\frac{E}{r}-Tr-\ln(t)}\right)}{r}, 
	\end{equation}
where $W_0$ defines the principal branch of the Lambert function $W$~\emph{\citep{lambert1758}}.
\end{theorem}
Theorem \ref{thm::1} includes the analysis of GD if the number of layers $L = 1$, but it can be easily extended to GD in the stochastic setting when $L > 1$. The proof of Theorem \ref{thm::1} is given in Appendix \ref{proof::thm}.

The explicit analytical estimate $g(t)$ justifies that the LIGHT function allows for faster convergence in comparison with the reported rate $g(t) = \ln(t)$ under the same assumptions. In addition, it mostly depends on the relation between $E$ and $r$.

\section{Experiments}

Having established that there is a beneficial interplay between $E$ and $r$, we now turn to an empirical study of different configurations of the LIGHT neuron to see whether we observe an increase in convergence rate and inductive bias according to the expectations (see Figure~\ref{fig:config}).
We compare the three non-adaptive methods - SGD with the -default- configuration (sigmoid-sgd) and SGD with the -r-, -E-, -Er- configurations (LIGHT-V  and LIGHT-G)  -  to two popular adaptive methods with the -default- configuration - Adam (sigmoid-adam) and AdaGrad (sigmoid-adagrad). We used the default parameters for all the optimizers, with the mini-batch size $|B(t)| = 75$ and $n_\mathrm{epoch} = 1500$. 
We first study performance on synthetic datasets for different network architectures to derive the strategy for setting pre-defined values of the rates $E$ and $r$.
Then, we validate the strategy on experimental datasets.  

The LIGHT function was implemented as a custom output activation layer with Keras \texttt{class LIGHT(Layer)}. The layer controls a population of impulses passing through the LIGHT neuron.

\subsection{Synthetic datasets}

\paragraph{Design of experiments}
We generated a set of synthetic datasets for  different $m = \{100, 1000, 5000, 10000\}$ and $n = \{2, 20, 200, 2000\}$. The centers of clusters for a binary classification task were chosen at (-0.75, 2.25) and (1, 2) with the standard deviation \emph{cluster std} = \{0.25, 0.5, 0.75, 1\}. 
Network architectures were constructed from $L = \{0, 1, 2, 3\}$ layers and $d_l = \{1, 10, 100, 1000\}$ neurons in each hidden layer. 
The datasets were randomly split into training (80\%) and testing (20\%) subsets.
For training, we used the default values for all the optimizers, with the mini-batch size $|B(t)| = 75$ and $n_\mathrm{epoch} = 1500$. The number of runs is equal to 10.

\vspace{-3mm}
\paragraph{Hyperparameter optimization}
The hyperparameters of LIGHT-V and LIGHT-G  were optimized with a random search~\citep{bergstra2012} with a 2.5\% random pick of all possible parameters combinations from the full grid space within the following ranges: $r\in[0.1, 20]$ with the number of points $n_r = 5$; $E\in[0.1, 20)$, $n_E = 5$; $T\in[0.1,20]$, $n_T = 3$; $N_T\in[0.2, 0.8]$, $n_{N_T} = 3$. We set $K = 1$ to ensure a range of output values equal to [0,1]. We initially tried number of epochs for the hyperparameters search $h_{\rm{epoch}} = \{1, 10, 100\}$ and found that $h_{\rm{epoch}} = 1$ was the best performing. Increasing $h_{\rm{epoch}}$ did not yield much improvement.  

\begin{figure}[h!]
	\centering
	\includegraphics[width=0.96\columnwidth]{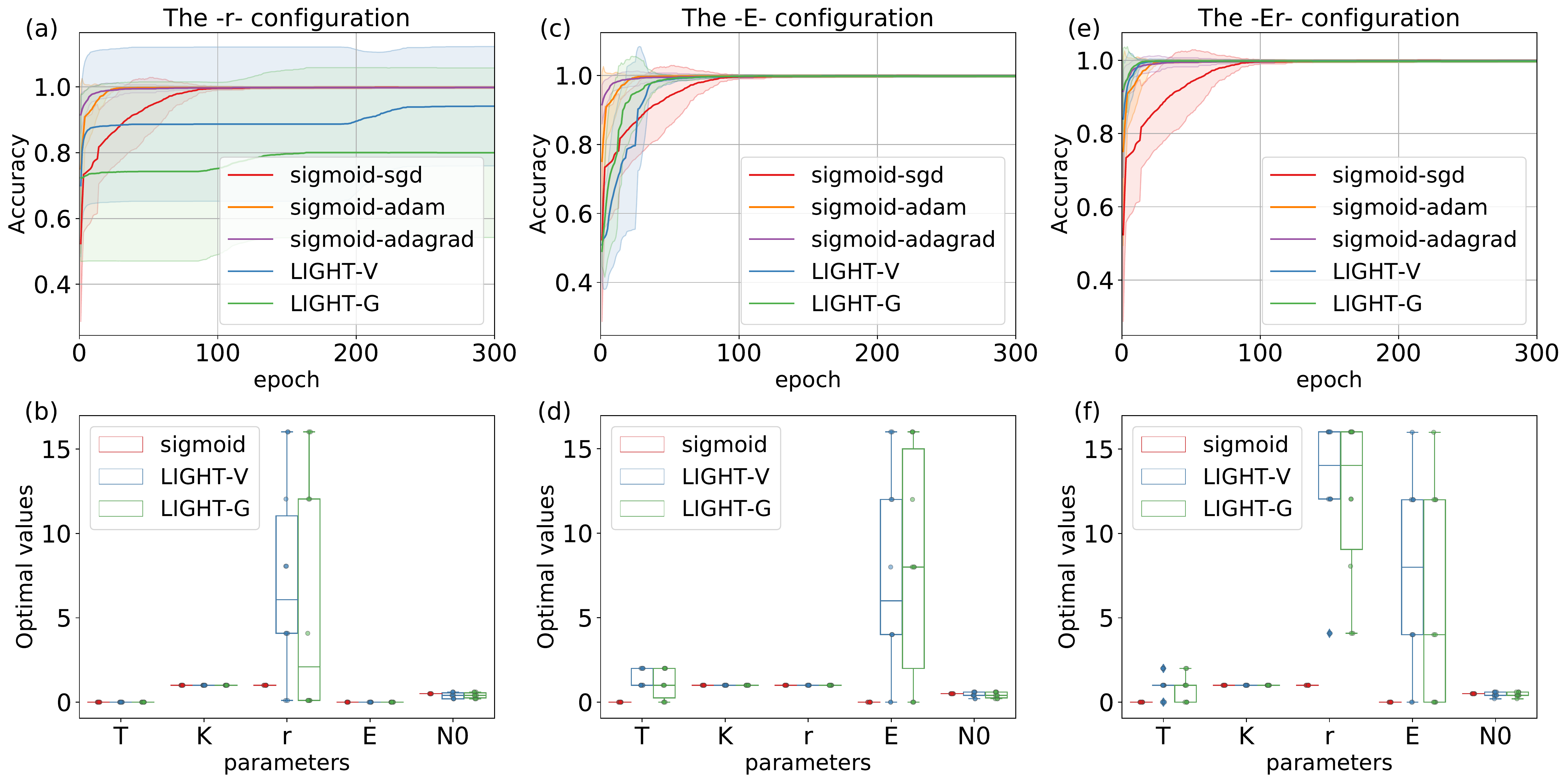}
	\caption{The accuracy curves on testing for $L = 1$, $d_l = 100$, $m = 1000$, $n = 2$, \emph{cluster std} = 0.25}
	\label{fig::syn:1}
	\vspace{-5mm}
\end{figure}

\vspace{-3mm}
\paragraph{Results}
Figures~\ref{fig::syn:1} (a), (c), (e) demonstrate the accuracy curves on testing for the -r-, -E-, -Er- configurations in case of $L = 1$, $d_l = 100$,  $m = 1000$, $n = 2$, and \emph{cluster std} = 0.25 that generates a linearly separable dataset. As we can see, the presented results comply with the expected behavior of the curves given in Figure~\ref{fig:config}. In the -Er- configuration,  we see that introducing harvesting increases the \emph{per} capita growth rate $r$ (see Figure~\ref{fig::syn:1} (f), Table~\ref{tab::L1std25}, -Er-) compared to the value $r$ in the -r- configuration (see Figure~\ref{fig::syn:1} (b), Table~\ref{tab::L1std25}, -r-).  We computed the values $H$, $E^*$ and $H^*$ for both LIGHT-V and LIGHT-G with regard to the definitions given in Sections~\ref{verh} and \ref{gomp}. 
For conciseness, the comprehensive empirical analysis on different combinations of configurations, network architectures, and datasets is deferred to Appendix~\ref{exp::syn}. Figure~\ref{fig:sum:er} shows the summary plot in the -Er- configuration for the following fixed values: $L = 0$, $d_l = 100$, $m = 1000$, $n = 2$, \emph{cluster std} = 0.25, $h_{\rm{epoch}} = 1$. We can see here that LIGHT-V and LIGHT-G outperform adaptive and non-adaptive default methods on the test subsets across all evaluated models and tasks. For the sake of comparison, the summary plots in two other configurations are given in Appendix~\ref{exp::sum}.

The denotations marked with an asterisk stand for the maximum harvesting rate for a given value of the growth rate $r$. 
From Table ~\ref{tab::L1std25} we concluded that the rates can be chosen within a wide range: m $\pm$ sd. It allowed us to  suggest the following simple strategy for predefining the values of $r$ and $E$: first, set $r$ and, then, compute $E^*$ to maximize $H^*$ based on the chosen value $r$.   

\begin{table}[h!]
	\caption{The estimates of the rates for $L = 1$,  $m = 1000$, $n = 2$, \emph{cluster std }= 0.25}
	\label{tab::L1std25}
	\centering
	\begin{tabular}{>{\rowmac}l>{\rowmac}l>{\rowmac}l>{\rowmac}l>{\rowmac}l>{\rowmac}l>{\rowmac}l<{\clearrow}}
		\toprule
		& & \multicolumn{3}{l}{Optimal values}              
		& \multicolumn{2}{l}{Pre-defined values} \\
		\cmidrule(r){2-7}
		LIGHT & Configuration   & m($r$)$\pm$ sd($r$)   & m($E$)$\pm$ sd($E$)  & $H$ & $E^*$ & $H^*$\\
		\midrule
		-V & -r- & 7.26$\pm 5.87$  & 0.0$\pm 0.0$  &  0.0 & 3.63 &  1.82\\
		& -E-  & 1.0$\pm 0.0$  & 7.6$\pm 6.1$  & 0.0 &  0.5 & 0.25\\
		& \setrow{\bfseries}-Er- & 13.23$\pm 3.78$  & 7.6$\pm 5.8$  &  3.23 & 6.6 & 3.3\\
		
		\midrule
		-G & -r- & 6.07$\pm 4.3$  & 0.0$\pm 0.0$  &  0.0 & 6.07 & 2.23\\
		& -E-  & 1.0$\pm 0.0$  & 8.4$\pm 6.66$  &   0.0 & 1.0 & 0.37\\
		& \setrow{\bfseries}-Er- & 12.04$\pm 4.96$  & 6$\pm 6.32$  &  3.65 & 12.04 & 4.43\\
		\bottomrule
	\end{tabular}
\end{table}

\begin{figure}[h!]
	\centering
	\includegraphics[width=0.96\columnwidth]{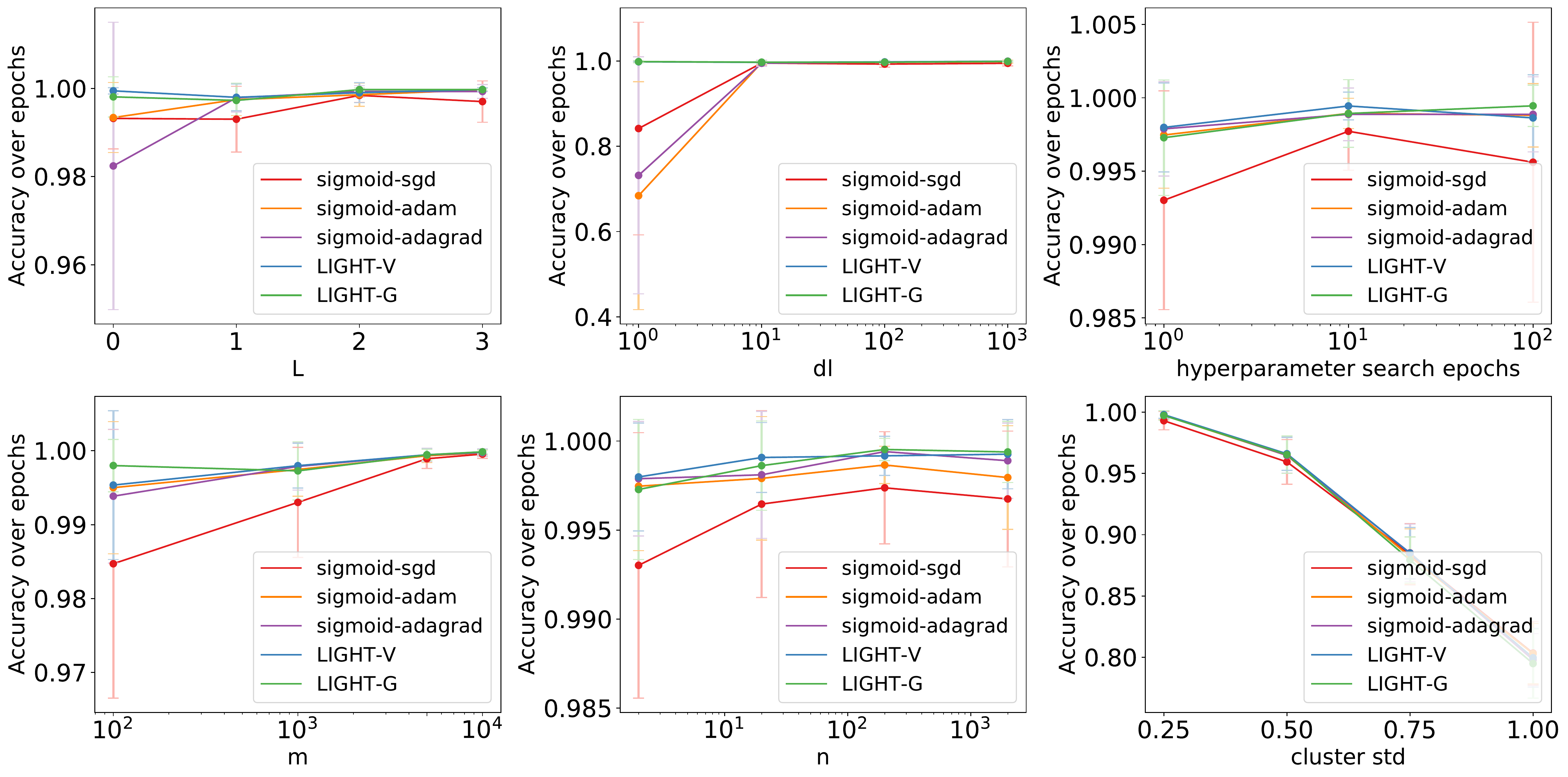}
	\caption{The values of accuracy on testing over epochs: The \textbf{-Er-} configuration}
	\label{fig::sum:er}
	\vspace{-5mm}
\end{figure}

\subsection{Experimental datasets}

\paragraph{Design of experiments}
We tested the LIGHT function on the six well-known experimental datasets: \emph{breast cancer win}, \emph{heart statlog}, \emph{pima indians}, \emph{mnist}, \emph{fashion mnist}, \emph{cifar10} from UCI Machine Learning Repository. 
The labels of the last three image classification datasets were binarized. We randomly extracted samples $m=1000$ from each of them and split into training (80\%) and testing (20\%) subsets.

\vspace{-3mm}
\paragraph{Hyperparameter optimization}
We followed the identical procedure to optimize hyperparameters on the experimental datasets.
The results were compared with those for the predefined values on a network architecture with $L = 1$ and $d_l = 10$. Each experiment was conducted 10 times.

\begin{figure}[h!]
	\centering
	\includegraphics[width=0.96\columnwidth]{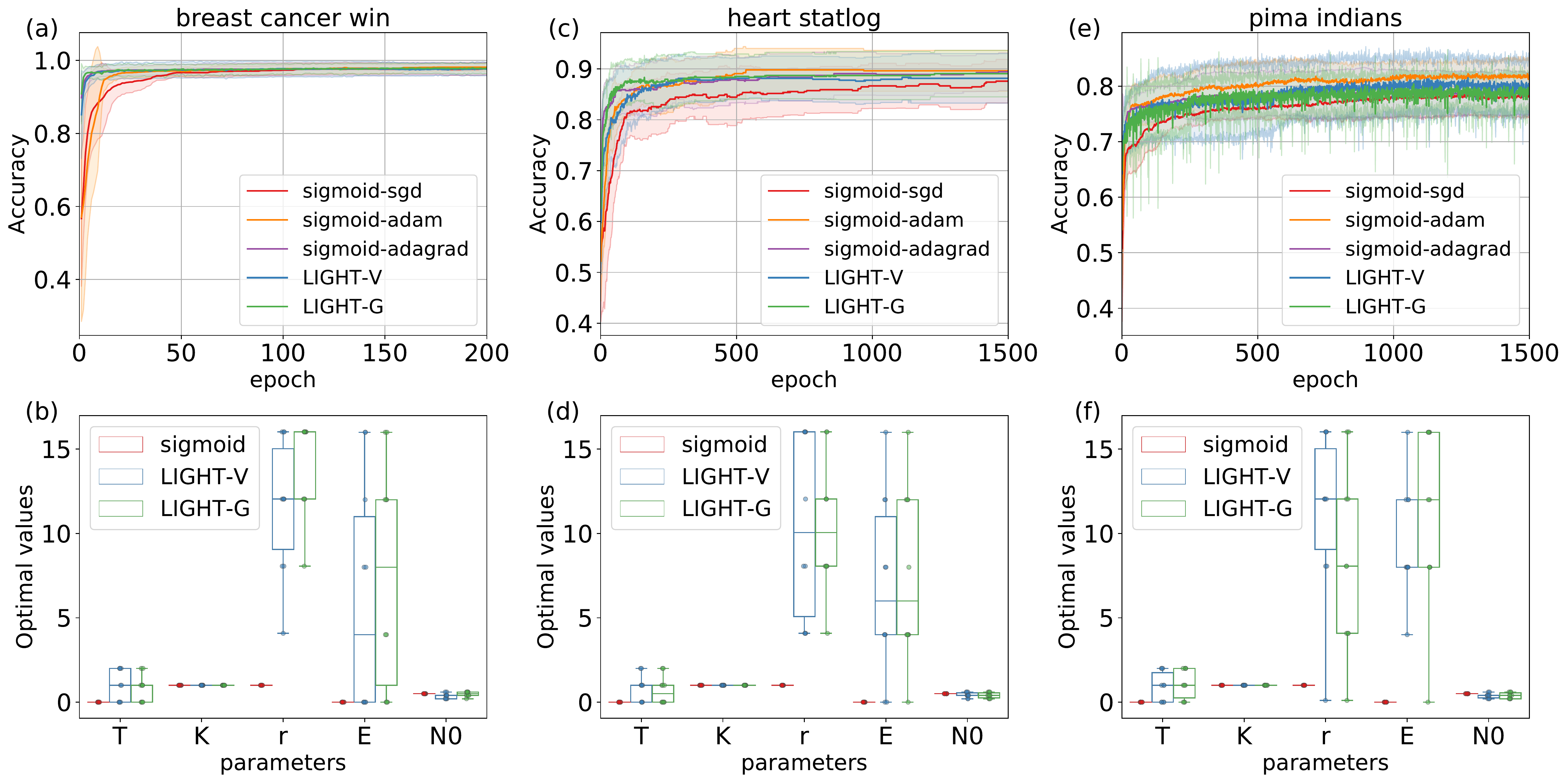}
	\caption{The accuracy curves on testing for \emph{pima indians}, \emph{breast cancer wisc}, \emph{heart statlog}}
	\label{fig:exp:1}
	\vspace{-3mm}
\end{figure}

\begin{figure}[h!]
	\centering
	\includegraphics[width=0.96\columnwidth]{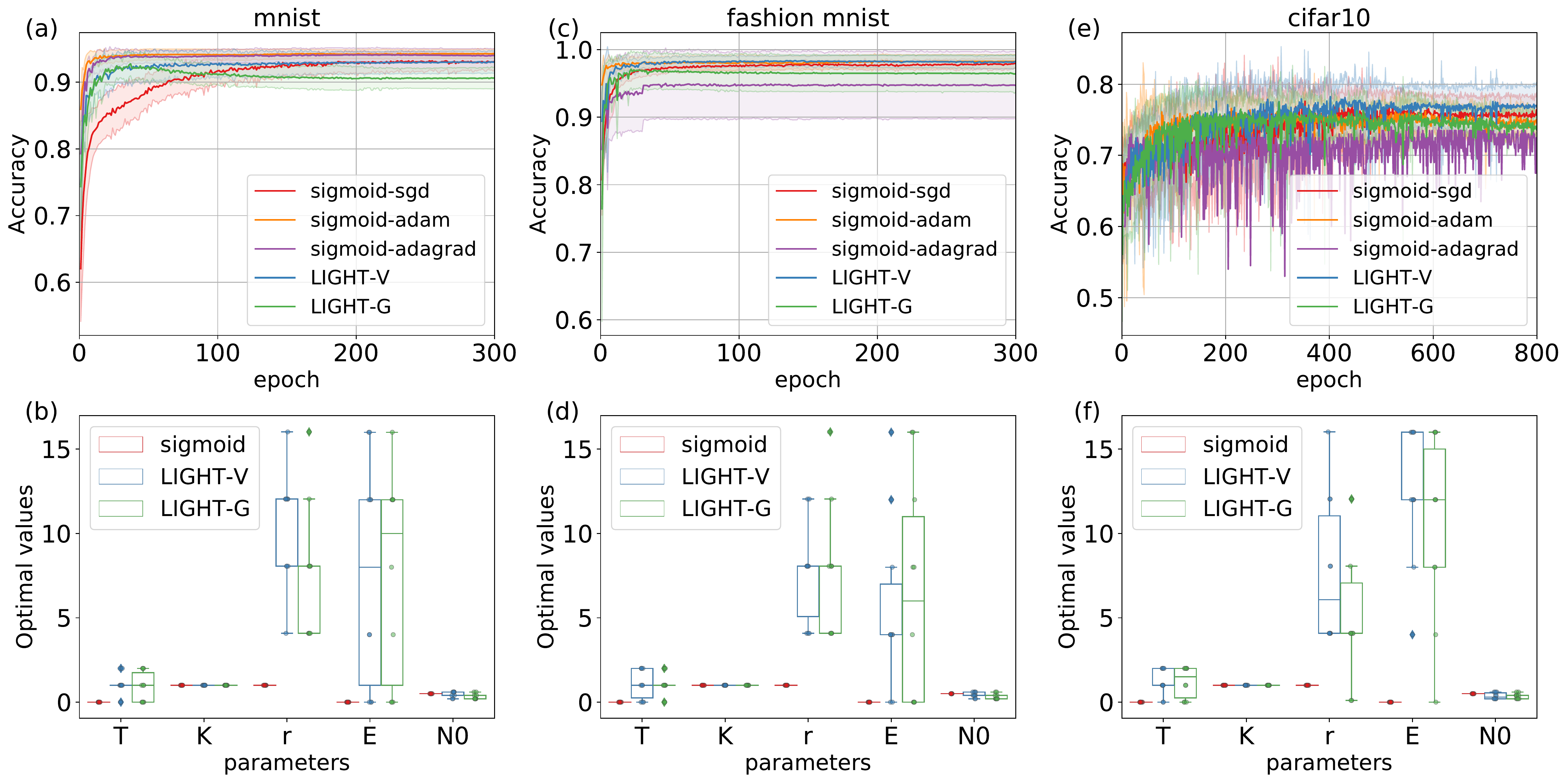}
	\caption{The accuracy curves on testing for \emph{mnist}, \emph{fashion mnist}, \emph{cifar10}}
	\label{fig:exp:2}
	\vspace{-5mm}
\end{figure}

\vspace{-3mm}
\paragraph{Results}
Figures~\ref{fig:exp:1} and \ref{fig:exp:2} show the accuracy curves on the test subset for the -Er- configuration. A rough estimate suggests that the dataset \emph{breast cancer win} with the most balanced combination $r$ and $E$  ($E \leq \frac{r}{2}$ for LIGHT-V and $E \leq  r$ for LIGHT-G) demonstrates the best results. The optimal hyperparameters are also the closest to the pre-defined values $E^*$ and $H^*$. The other datasets mostly require a larger pre-defined value $r$ to compensate for a higher harvesting rate $E$. In Appendix~\ref{exp::exp}, we provided a more detailed empirical study with all quantitative results.

\section{Related work}

\paragraph{Adaptive optimization}
The common approach to increasing the convergence rate is adopting optimization methods with a variable step size such as Adam ~\citep{kingma2015}, Adagrad~\citep{duchi2011}, Adadelta~\citep{zeiler2012} and etc.~\citep{kim2017, ruder2016}.  Using adaptive learning rate methods results in worse generalization~\citep{hoffer2017, wilson2017,  kim2017} as the limit direction of adaptive optimization methods is less predictable and stable compared to non-adaptive methods  \citet{gunasekar2018}. 
\citet{allen-zhu2019, arora2018, li2018} demonstrated that non-adaptive SGD learns an overparameterized model with random initialization and small generalization errors. We show that SGD with two balanced pre-defined values of \emph{per} capita growth and harvesting rates outperform the most common adaptive gradient methods without overparametrization.

\vspace{-3mm}
\paragraph{Hyperparameter optimization}
The results of this study comply with the recent research in deep learning on hyperparameter optimization. 
\citet{hayou2019, schoenholz2017} revealed the importance of specifically chosen hyperparameters, known as the ``Edge of Chaos'', for good performance and fast convergence. Addressing similar problems, we reconsider neural networks from a population dynamics perspective to replace the optimization procedure with a simple strategy for setting pre-defined parameters.  

\vspace{-3mm}
\paragraph{Adaptive learning and  evolution}
The processes inside the LIGHT neuron share some similarity with the Baldwin effect which was successfully adopted in bias shifting algorithms~\citep{downing2010, hinton1987, fernando2018}. The reported studies mostly focused on increasing the inductive bias with adaptively evolving optimization schemes. In contrast, we intend to simplify learning algorithms in deep networks while fulfilling the current expectation from the gradient-based methods.

\vspace{-1mm}
\section{Conclusion}

We introduced the LIGHT function to complement inner processes inside neurons in deep networks with growing and harvesting borrowed from population dynamics. This function allows explicit control of the trade-off between inductive biases and convergence rates with two balanced pre-defined values of \emph{per} capita growth and harvesting rates. The proposed LIGHT function increases the transparency in deep learning by increasing both convergence rate and inductive bias without overcomplicating optimization processes and overparametrizing models.






\section*{Broader Impact}
Deep learning methods are widely used in areas of high societal significance such as health, police, mobility or education 
but still belie a lack of transparency that is vital for their adoption. This fact reveals even more impactful trade-off we would like to accentuate: A human society that supports profit-driven business models often overlooks transparency in favor of frictionless functionality. 

Consequently, even if the current trends in deep learning have spurred numerous studies in transparency and interpretability, society may slow the progress down with its priorities and principles. By looking into a key trade-off between inductive bias and convergence rate in deep learning, we would like to address the problem of finding a more meaningful balance between profit and social welfare which is a necessary condition for a sustainable society.




\bibliography{nips2020}




\clearpage

\begin{appendices}

\setcounter{section}{0}
\section{On Theorem~\ref{thm::1}}
\subsection{Definition \ref{def::1}}
\label{proof::def}
Let us introduce $f_1(t) = \left(\frac{E}{r} + \ln_1\left(\frac{N_T}{K}\right)\right)\mathrm{e}^{-r(t-T)}$. 
We show that the model \eqref{eqlog:1s} with regard to $f_1(t)$ can be presented as: 
\begin{equation}
\label{eq::app:1}
\ell^{\mathrm{V}}(t; r, E, K, T, N_T) = \displaystyle\frac{s}{1+f_1(t)\varepsilon_1},
\end{equation}
where $s = K\left(1-\frac{E}{r}\right)$, $\varepsilon_1 = -\frac{N_T}{K}\mathrm{e}^{1-\frac{E}{r}}$.
With regard to the definition $f_1(t)$, the equation \eqref{eq::app:1} gives:
\begin{equation}
\ell^{\mathrm{V}}(t; r, E, K, T, N_T)= \displaystyle\frac{s}{1+\left(\frac{N_T}{K\left(1-\frac{E}{r}\right)}-1\right)\left(1-\frac{E}{r}\right)\varepsilon_1\mathrm{e}^{-r(t-T)}} =
\displaystyle\frac{s}{1-\left(1-\frac{s}{N_T}\right)\mathrm{e}^{-(r-E)(t-T)}}
\nonumber
\end{equation} 
that is equal to \eqref{eqlog:1s}.

Let us now introduce $f_2(t) = \left(\frac{E}{r} + \ln\left(\frac{N_T}{K}\right)\right)\mathrm{e}^{-r(t-T)}$. 
By analogy, we show that the model \eqref{eqlog:2s} with regard to $f_2(t)$ can be presented as: 
\begin{equation}
\label{eq::app:2}
\ell^{\mathrm{G}}(t; r, E, K, T, N_T) = K\mathrm{e}^{f_2(t)\varepsilon_2},
\end{equation}
where $\varepsilon_2 = -\frac{E}{r}$.
Taking into account the definition $f_2(t)$, \eqref{eq::app:2} reduces to:
\begin{equation}
\ell^{\mathrm{G}}(t; r, E, K, T, N_T)= K\mathrm{e}^{\left(\frac{E}{r}+\ln\left(\frac{N_T}{K}\right)\right)\mathrm{e}^{-r(t-T)}\varepsilon_2}
=
K\mathrm{e}^{-\frac{E}{r}+\left(\frac{E}{r}+\ln\left(\frac{N_T}{K}\right)\right)\mathrm{e}^{-r(t-T)}},
\nonumber
\end{equation} 
that gives \eqref{eqlog:2s}.

Assuming $\displaystyle\frac{1}{1+f_1(t)\varepsilon_1} \approx \mathrm{e}^{f_2(t)\varepsilon_2}$, we define the generalized function \eqref{eqlog:4} with some $\varepsilon$.

The underlying reasoning behind a generalizing parameter $q$ with regard to the Gompertz model is explained by \citet{gray2017} in Section 7.2.3.


\subsection{Theorem~\ref{thm::1}}
\label{proof::thm}
\begin{proof}[Sketch of the proof]
According to \citep{nacson2018, soudry2018}, under Assumptions \ref{ass::1} and \ref{ass::2}, GD finds the global minimum  even if the loss function $\mathscr{L}(\bm{\uptheta})$ is non-convex. 
Since $\forall i: \bm{\uptheta}^{*\mathrm{T}}\mathrm{x}_i > 0$ and $-\ell '(t)>0$ for any finite t, with regard to the Cauchy–Schwarz inequality
$$\|\bm{\uptheta}(t)\|  \geq \frac{\|\bm{\uptheta}^{*\mathrm{T}}\bm{\uptheta}(t)\|}{\|\bm{\uptheta}^{*}\|},$$ 
$\lim_{t\rightarrow\infty}\|\bm{\uptheta}(t)\| = \infty$, $\forall i: \lim_{t\rightarrow\infty}  \bm{\uptheta}(t)^\mathrm{T}\mathrm{x}_i = \infty$.

Under the given conditions, the normalized weight vector converges to the normalized max margin vector in $L_2$ norm \citep{nacson2018, soudry2018}: 
\begin{equation}
\label{eq::03}
\left\|\frac{\bm{\uptheta}(t)}{\|\bm{\uptheta}(t)\|}-\frac{\hat{\bm{\uptheta}}}{\|\hat{\bm{\uptheta}}\|}\right\|
=\mathcal{O}\left(\frac{1}{g(t)}\right),
\nonumber
\end{equation}

where the margin converges as  $\min_i\frac{\bm{\uptheta}_t^\mathrm{T}\mathrm{x}_i}{\|\bm{\uptheta}_t\|}
= d - \mathcal{O}\left(\frac{1}{g(t)}\right)$, 
$d = \max_{\bm{\uptheta}}\min_i\frac{\bm{\uptheta}^\mathrm{T}\mathrm{x}_i}{\|\bm{\uptheta}\|} = \frac{1}{\hat{\bm{\uptheta}}}$.

To simplify the convergence analysis, we consider the continuous version of GD:
$$\bm{\uptheta}' (t) = - \nabla \mathscr{L}(\bm{\uptheta}(t), r, E, K, T, N_T)).$$
Assume
$$\bm{\uptheta}' (t) = \sum_{i=1}^m\ell '(\bm{\uptheta}(t)^\mathrm{T}\mathrm{x}_i; r, E, K, T, N_T)\mathrm{x}_i,$$
where \\
$\ell '(t; r, E, K, T, N_T) = -a\mathrm{e}^{-f(t)}$, $f(t) = r(t-T)-\left(\frac{E}{r}+\ln_q\left(\frac{N_T}{K}\right)\right)\mathrm{e}^{-rt}+\frac{E}{r}\mathrm{e}^{-rt}$,\\ $a = \varepsilon Kr\left(\ln_q\left(\frac{N_T}{K}\right)+\frac{E}{r}\right)$. According to \citep{soudry2018, nacson2018}, the weight vector can be presented asymptotically as
$$\bm{\uptheta} (t) = g(t)\hat{\bm{\uptheta}}+o(g(t)).$$
Skipping $a$ for simplicity and finding the derivative  $\bm{\uptheta}' (t)$, we require 
$g'(t) = \mathrm{e}^{-f(g(t))}$, $o(g(t)) = \frac{1}{f'(g(t))}$, from where, as $\lim_{t\rightarrow\infty}g(t)f'(g(t)) = \infty$, we have $f'(t) = \omega(\frac{1}{t})$ and  $f(t) = \omega(\ln(t))$ (see Assumption \ref{ass::2}). Then, we can approximate 
$g'(t) \approx \mathrm{e}^{-f(g(t)) - \ln(f'(g(t)))}$ which has a closed form solution as stated by \citet{nacson2018}:
$g(t) = f^{-1}(\ln(t+C))$. Finding the inverse function gives:
\begin{equation}
\label{eq::thm}
g(t) = \frac{E}{r^2}+T+\frac{\ln(t)+W_0\left(\frac{\left(\frac{E}{r} + \ln_q\left(\frac{N_T}{K}\right)\right)\mathrm{e}^{-\frac{E}{r}-Tr}}{t}\right)}{r} 
\nonumber
\end{equation}
i.e.
\begin{equation}
g(t) = \frac{E}{r^2}+T+\frac{\ln(t)+W_0\left(\left(\frac{E}{r} + \ln_q\left(\frac{N_T}{K}\right)\right)\mathrm{e}^{-\frac{E}{r}-Tr-\ln(t)}\right)}{r} 
\nonumber
\end{equation}
which proves the validity of Theorem \ref{thm::1}.
\end{proof}

\renewcommand\thefigure{\thesection.\arabic{figure}}  
\renewcommand\thetable{\thesection.\arabic{table}}  
\section{Synthetic datasets}  
\setcounter{figure}{0}
\setcounter{table}{0}
\label{exp::syn}
We ran the experiments for different combinations
$L = \{0, 1, 2, 3\}$,  \emph{cluster std} = \{0.25, 0.5, 0.75, 1\} and fixed values   $d_l = 100$, $m = 1000$, $n = 2$, $h_{\rm{epoch}} = 1$ for plotting 2D graphs (see Figures \ref{fig::1}-\ref{fig::11}). Tables \ref{tab::L0std25}-\ref{tab::L3std75} provide the quantitative results on hyperparameter optimization. The values biased toward the correct proportion between $r$  and $E$ ($E \leq \frac{r}{2}$ for LIGHT-V and $E \leq  r$ for LIGHT-G) are highlighted in bold. 
As can be seen, these values guarantee the higher harvesting rate $H^*$.
In some cases, we can also see a favourable situation when $r < E$ but  harvesting still increases the growth rate $r$ that also results in higher $H^*$.
Considering high sd for both $r$ and $E$, we highlighted these values as well.

For completeness, Figures \ref{fig::L0std-r} -\ref{fig::L3std-Er} demonstrate the accuracy curves for each LIGHT configuration: -r-, -E-, and -Er-. The accuracy plot for  $L = 1$, $d_l = 1000$, $m = 1000$, $n = 2$, and \emph{cluster std} = 0.25 is given in the paper (see Figure~\ref{fig::syn:1}). 

Finally, we conducted  a series of experiments, varying $d_l = \{1, 10, 100, 1000\}$ (see Figures \ref{fig::dl-r}-\ref{fig::dl-Er}), $m = \{100, 1000, 5000, 10000\}$ (see Figures \ref{fig::N-r}-\ref{fig::N-Er}), $n = \{2, 20, 200, 2000\}$ (see Figures \ref{fig::M-r}-\ref{fig::M-Er}), $h_{\rm{epoch}} = \{1, 10, 100\}$ (see Figures \ref{fig::hyper-r}-\ref{fig::hyper-Er}) for fixed $L = 1$, \emph{cluster std} = 0.25.
All the quantitative results were used to create the summary plots depicted in Appendix~\ref{exp::sum}. 

\begin{figure}[h!]
	\centering
	\includegraphics[width=0.95\columnwidth]{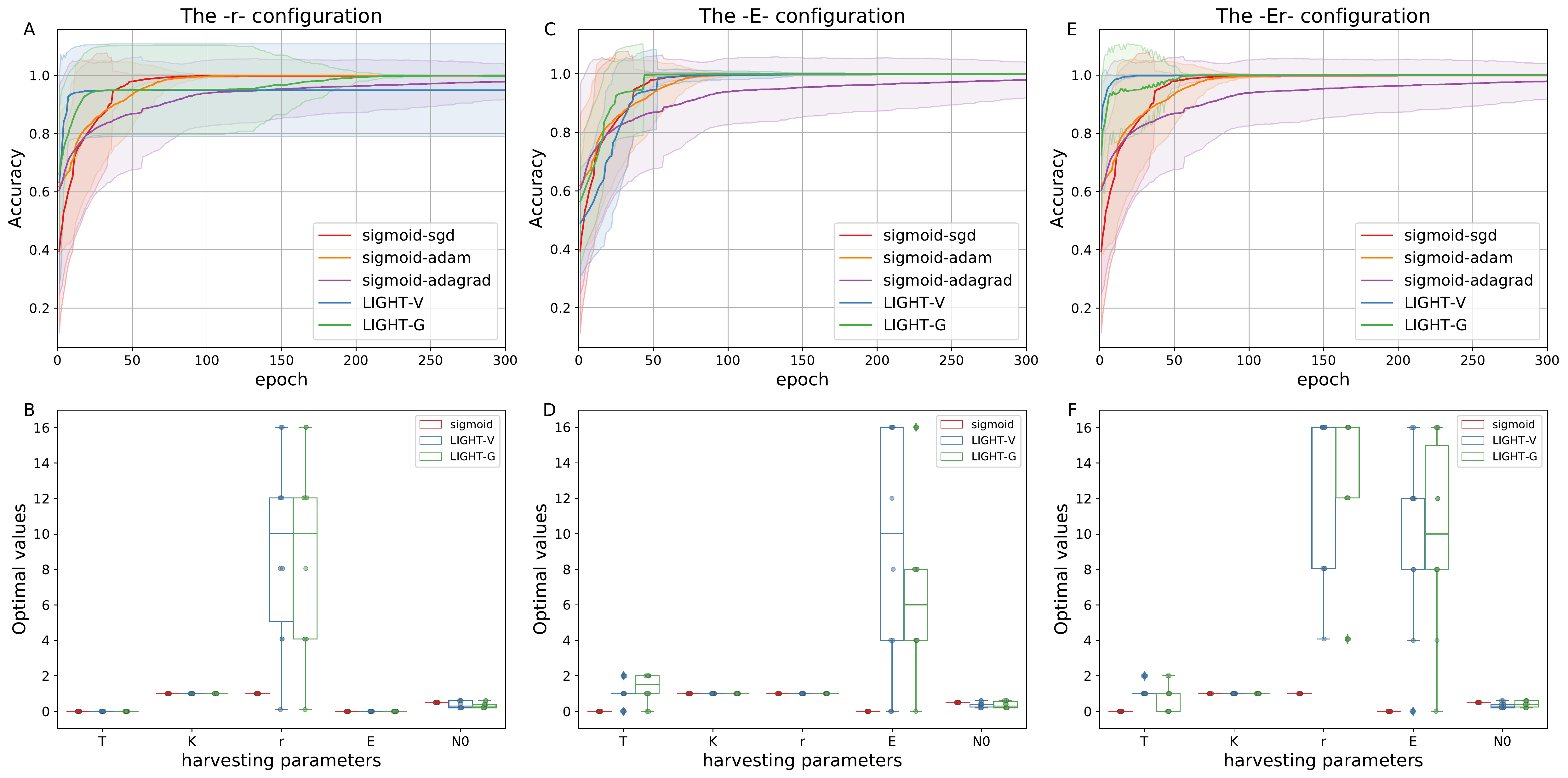}
	\caption{The accuracy curves on testing for $L = 0$, $d_l = 100$, $m = 1000$, $n = 2$, $h_{\rm{epoch}} = 1$, and \emph{cluster std }= 0.25}
	\label{fig::1}
\end{figure}

\begin{figure}[h!]
	\centering
	\includegraphics[width=0.95\columnwidth]{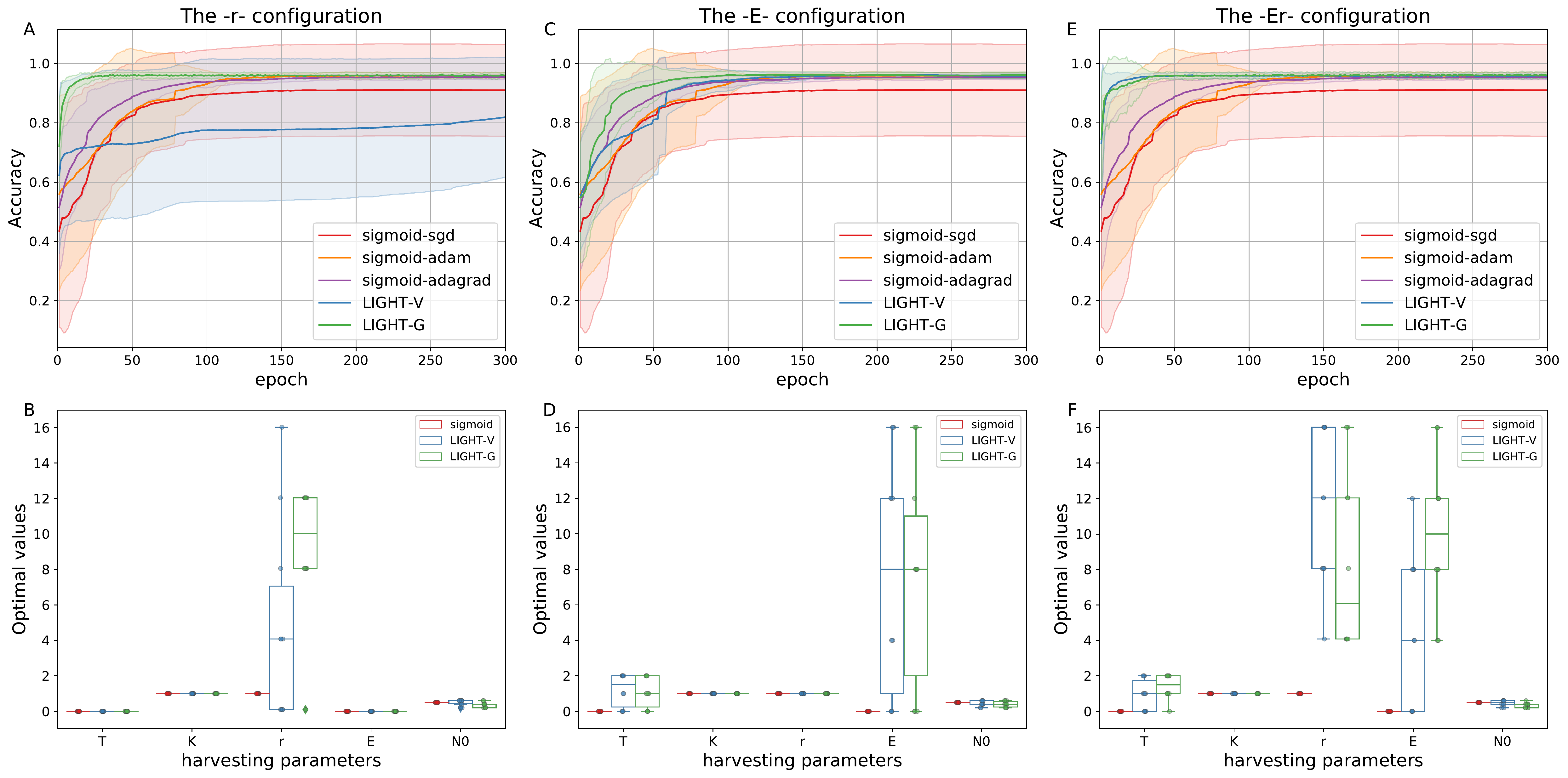}
	\caption{Accuracy for $L = 0$, $d_l = 100$, $m = 1000$, $n = 2$, $h_{\rm{epoch}} = 1$, and \emph{cluster std }= 0.5}
	\label{fig::2}
\end{figure}

\begin{figure}[h!]
	\centering
	\includegraphics[width=0.95\columnwidth]{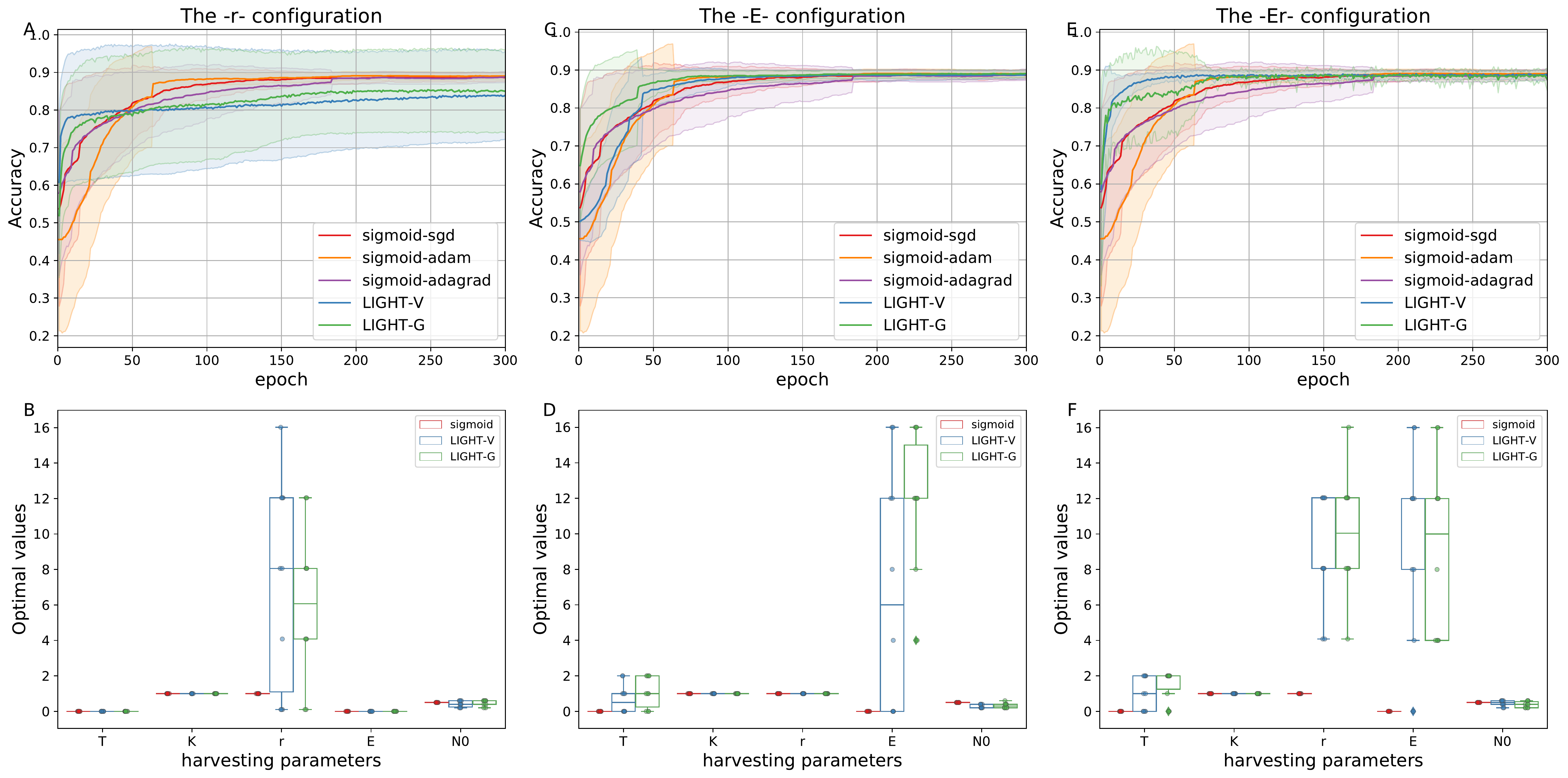}
	\caption{The accuracy curves on testing for $L = 0$, $d_l = 100$, $m = 1000$, $n = 2$, $h_{\rm{epoch}} = 1$, and \emph{cluster std} = 0.75}
	\label{fig::3}
\end{figure}


\begin{figure}[h!]
	\centering
	\includegraphics[width=0.95\columnwidth]{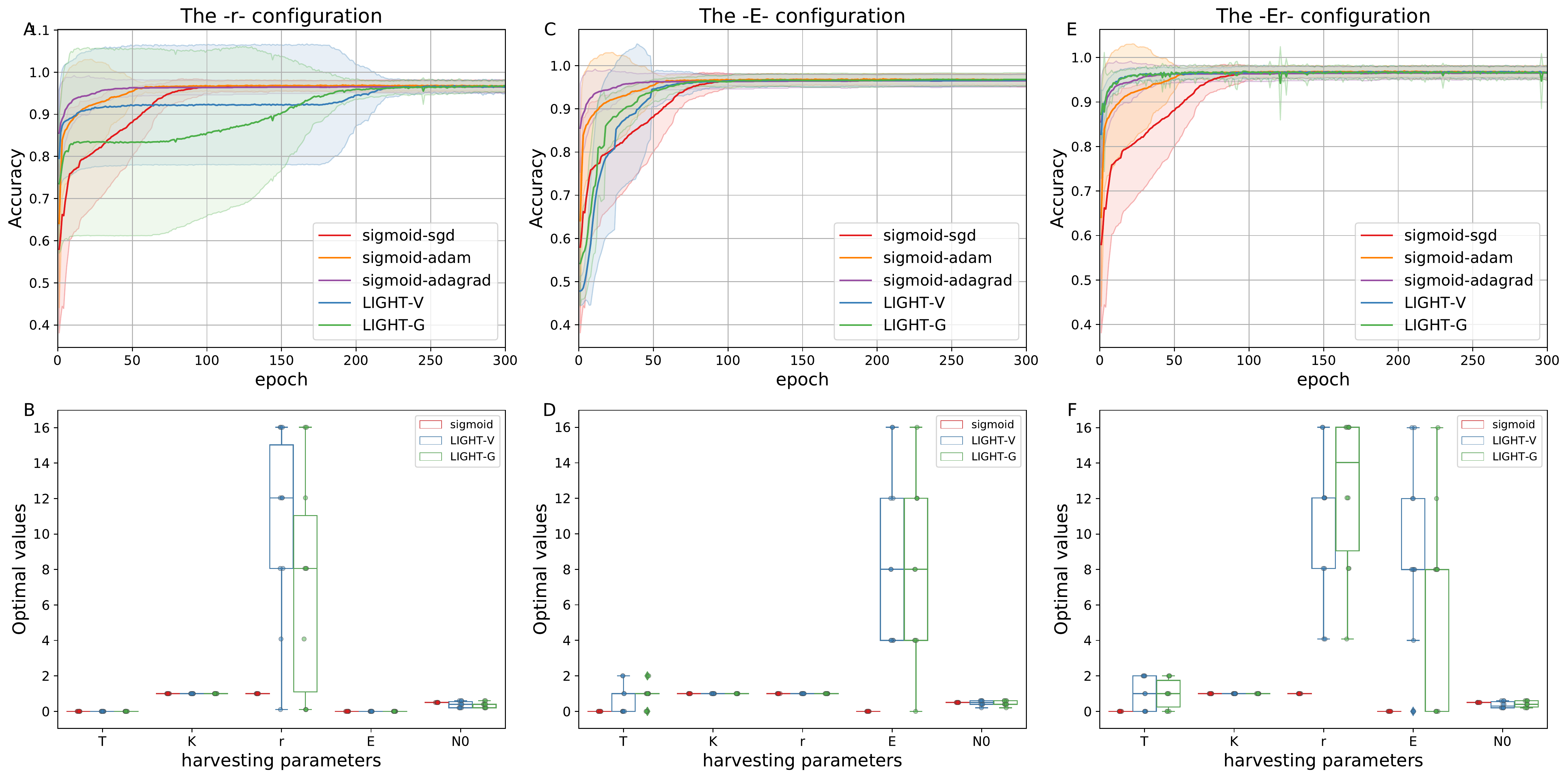}
	\caption{The accuracy curves on testing for $L = 1$, $d_l = 100$, $m = 1000$, $n = 2$, $h_{\rm{epoch}} = 1$, and \emph{cluster std} = 0.5}
	\label{fig::4}
\end{figure}

\begin{figure}[h!]
	\centering
	\includegraphics[width=0.95\columnwidth]{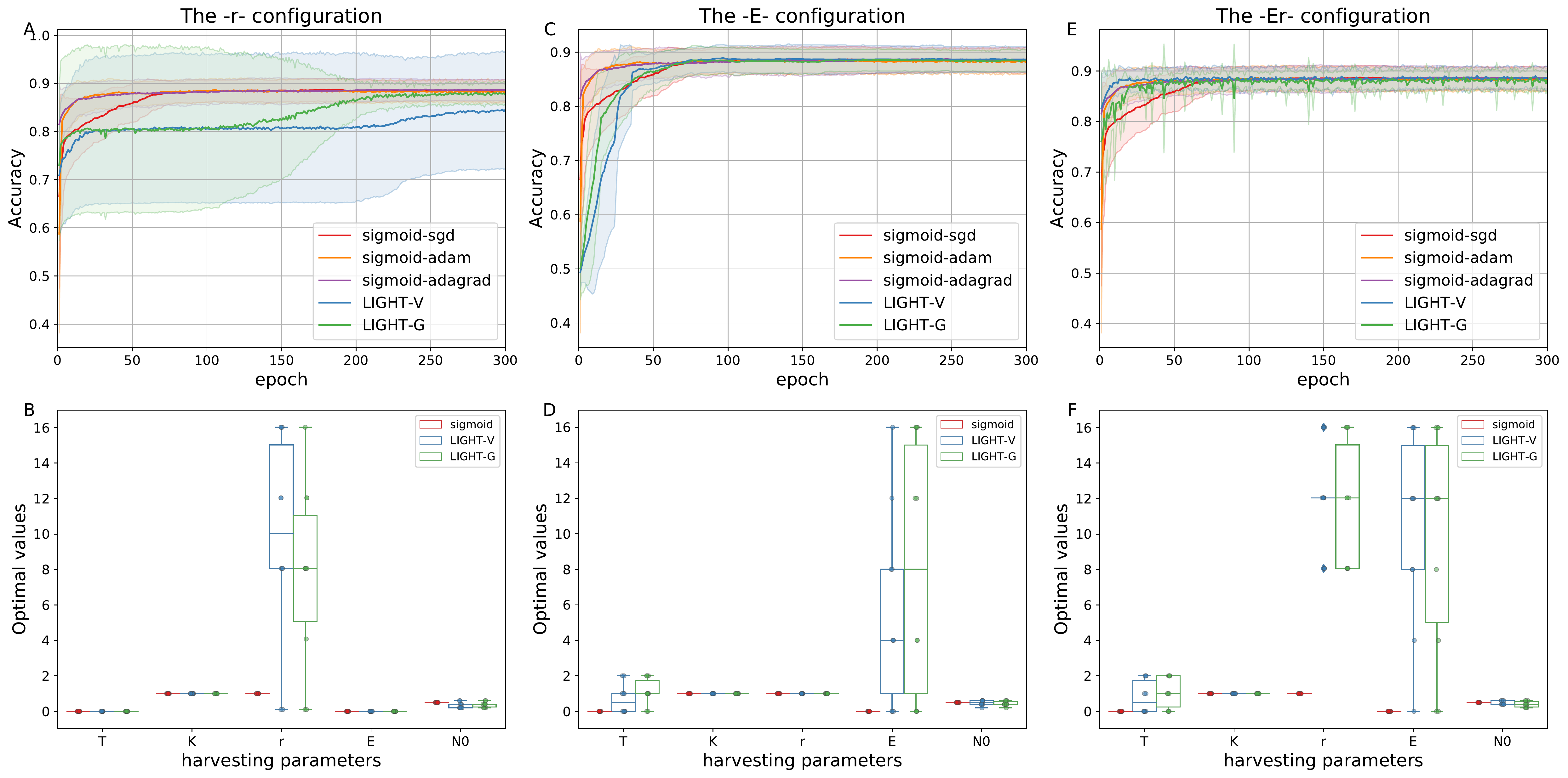}
	\caption{The accuracy curves on testing for $L = 1$, $d_l = 100$, $m = 1000$, $n = 2$, $h_{\rm{epoch}} = 1$, and \emph{cluster std }= 0.75}
	\label{fig:5}
\end{figure}

\begin{figure}[h!]
	\centering
	\includegraphics[width=0.95\columnwidth]{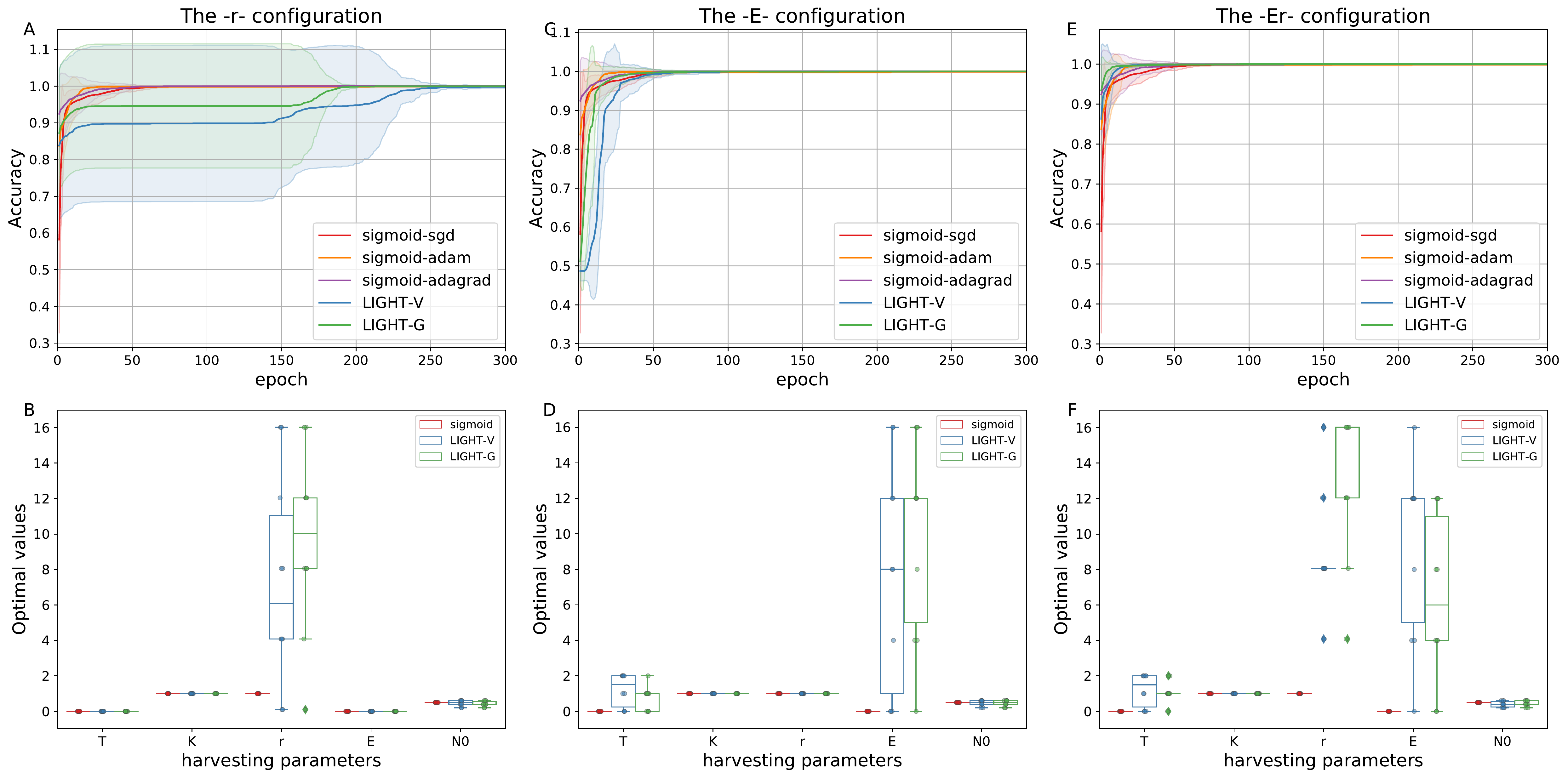}
	\caption{The accuracy curves on testing for $L = 2$, $d_l = 100$, $m = 1000$, $n = 2$, $h_{\rm{epoch}} = 1$, and \emph{cluster std }= 0.25}
	\label{fig::6}
\end{figure}

\begin{figure}[h!]
	\centering
	\includegraphics[width=0.95\columnwidth]{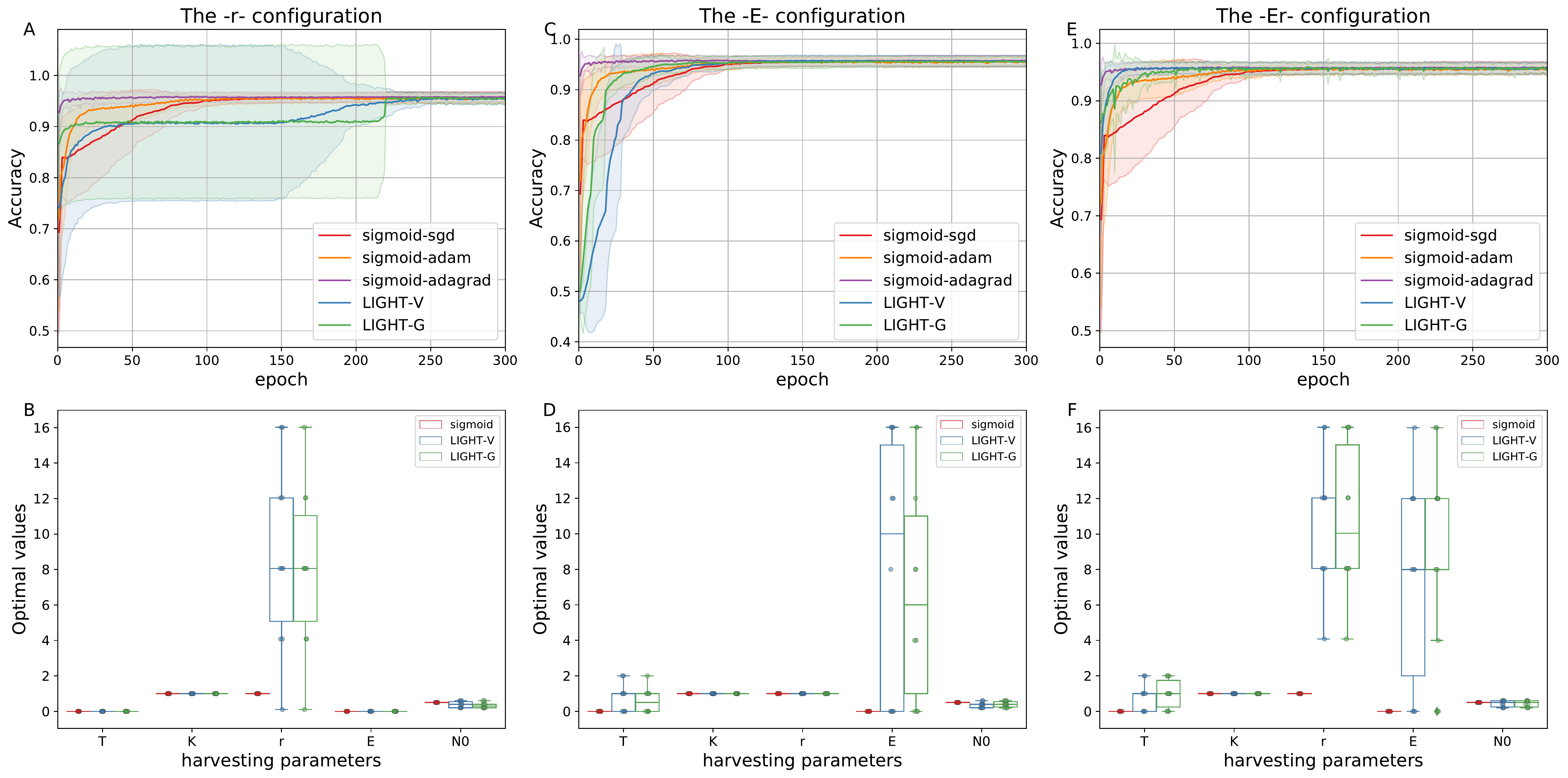}
	\caption{The accuracy curves on testing for $L = 2$, $d_l = 100$, $m = 1000$, $n = 2$, $h_{\rm{epoch}} = 1$, and \emph{cluster std }= 0.5}
	\label{fig::7}
\end{figure}

\begin{figure}[h!]
	\centering
	\includegraphics[width=0.95\columnwidth]{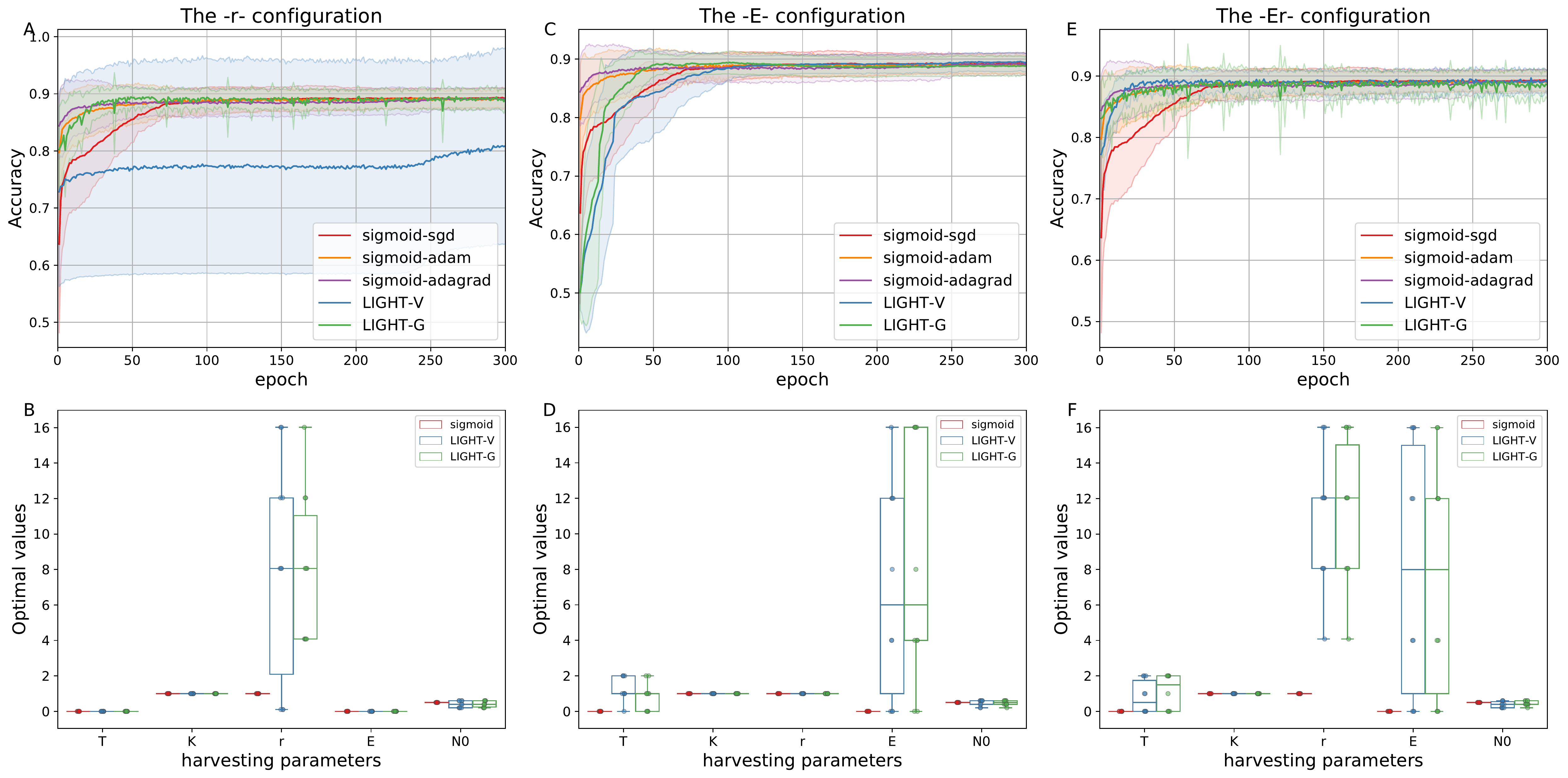}
	\caption{The accuracy curves on testing for $L = 2$, $d_l = 100$, $m = 1000$, $n = 2$, $h_{\rm{epoch}} = 1$, and \emph{cluster std}= 0.75}
	\label{fig::8}
\end{figure}

\begin{figure}[h!]
	\centering
	\includegraphics[width=0.95\columnwidth]{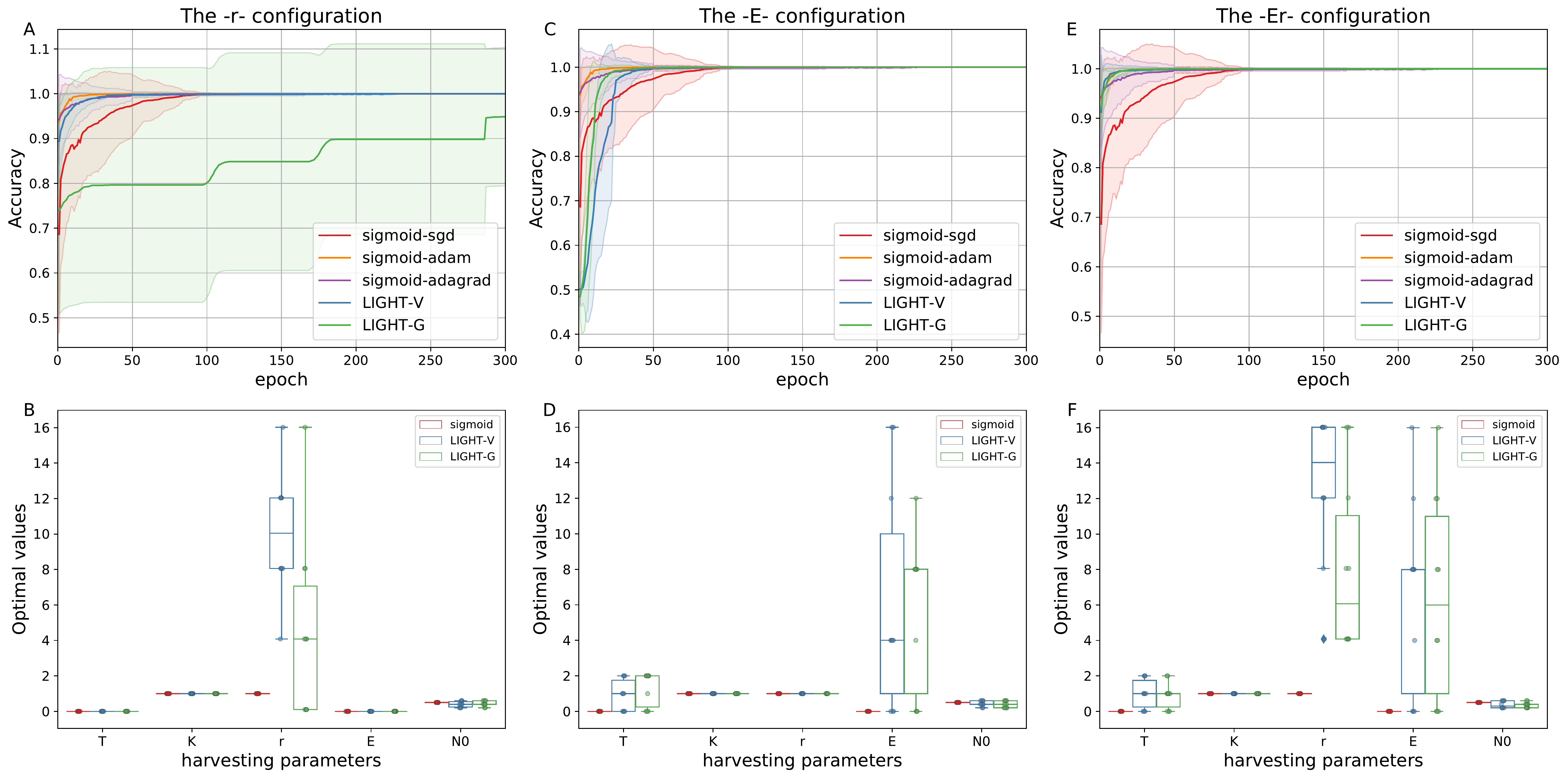}
	\caption{The accuracy curves on testing for $L = 3$, $d_l = 100$, $m = 1000$, $n = 2$, $h_{\rm{epoch}} = 1$, and \emph{cluster std} = 0.25}
	\label{fig::9}
\end{figure}

\begin{figure}[h!]
	\centering
	\includegraphics[width=0.95\columnwidth]{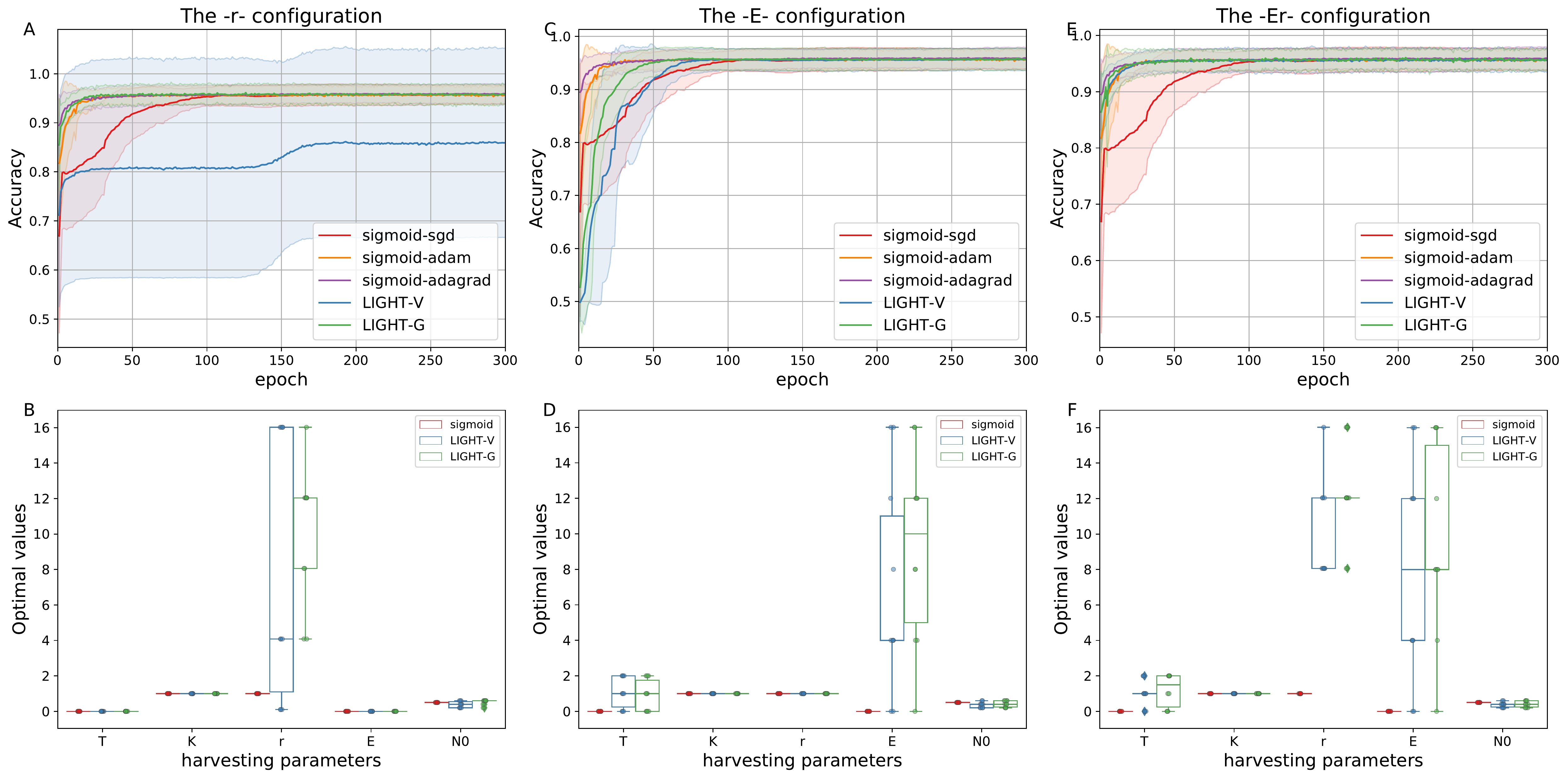}
	\caption{The accuracy curves on testing for $L = 3$, $d_l = 100$, $m = 1000$, $n = 2$, $h_{\rm{epoch}} = 1$, and \emph{cluster std} = 0.5}
	\label{fig::10}
\end{figure}

\begin{figure}[h!]
	\centering
	\includegraphics[width=0.95\columnwidth]{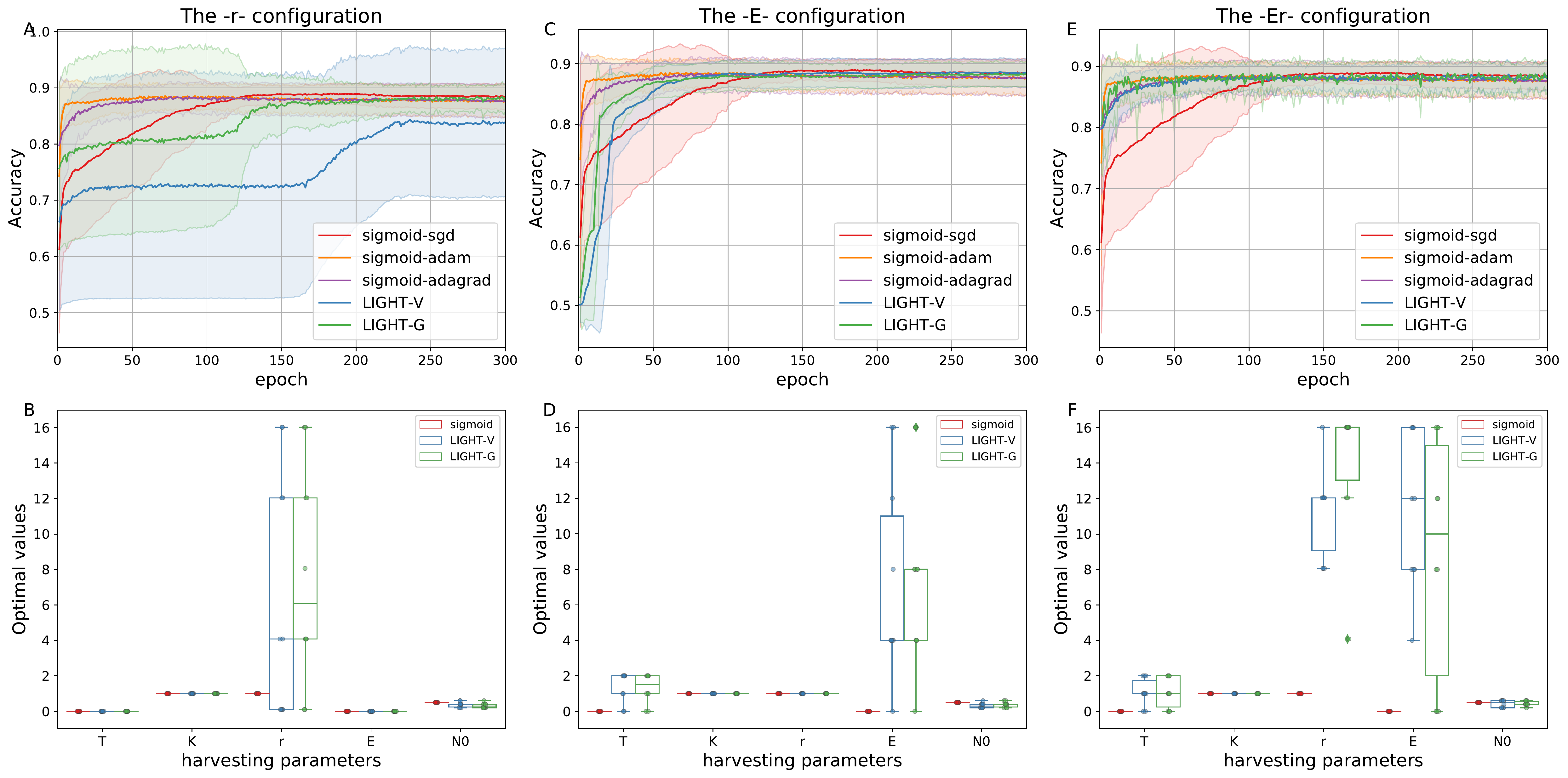}
	\caption{The accuracy curves on testing for $L = 3$, $d_l = 100$, $m = 1000$, $n = 2$, $h_{\rm{epoch}} = 1$, and \emph{cluster std} = 0.75}
	\label{fig::11}
\end{figure}

\begin{table}[h!]
	\caption{The estimates of the rates for $L = 0$, $d_l = 100$, $m = 1000$, $n = 2$, $h_{\rm{epoch}} = 1$, and \emph{cluster std }= 0.25}
	\label{tab::L0std25}
	\centering
\begin{tabular}{>{\rowmac}l>{\rowmac}l>{\rowmac}l>{\rowmac}l>{\rowmac}l>{\rowmac}l>{\rowmac}l<{\clearrow}}
	\toprule
	& & \multicolumn{3}{l}{Optimal values}              
	& \multicolumn{2}{l}{Pre-defined values} \\
	\cmidrule(r){2-7}
	LIGHT & Configuration   & m($r$)$\pm$ sd($r$)   & m($E$)$\pm$ sd($E$)  & $H$ & $E^*$ & $H^*$\\
		\midrule
		-V & -r- & 9.25$\pm 5.32$  & 0.0$\pm 0.0$  &  0.0 & 4.63 &  2.31\\
		& -E-  & $1.0\pm 0.0$  & 9.2$\pm6.81$  &  0.0 &  0.5 & 0.25\\
		& \setrow{\bfseries} -Er- & 12.44$\pm 4.76$  & 10$\pm 5.08$  & 1.96 & 6.22 & 3.1\\
		\midrule
		-G & -r- & 8.86$\pm 5.56$  & 0.0$\pm 0.0$  &  0.0 & 8.86 & 3.26\\
		& -E-  & 1.0$\pm 0.0$  & 6.4$\pm4.3$  &  0.01 & 1.0 & 0.37\\
		& \setrow{\bfseries}-Er- & 12$\pm 4.6$  & 10$\pm 5.42$  &  4.36 & 12.04 & 4.43\\
		\bottomrule
	\end{tabular}
\end{table}

\begin{table}[h!]
	\caption{The estimates of the rates for $L = 0$, $d_l = 100$, $m = 1000$, $n = 2$, $h_{\rm{epoch}} = 1$, and \emph{cluster std }= 0.5}
	\label{tab::L0std50}
	\centering
	\begin{tabular}{>{\rowmac}l>{\rowmac}l>{\rowmac}l>{\rowmac}l>{\rowmac}l>{\rowmac}l>{\rowmac}l<{\clearrow}}
		\toprule
		& & \multicolumn{3}{l}{Optimal values}              
		& \multicolumn{2}{l}{Pre-defined values} \\
		\cmidrule(r){2-7}
		LIGHT & Configuration   & m($r$)$\pm$ sd($r$)   & m($E$)$\pm$ sd($E$)  & $H$ & $E^*$ & $H^*$\\
		\midrule
		-V & -r- & 4.88$\pm 5.57$  & 0.0$\pm 0.0$  &  0.0 & 2.44 &  1.22\\
		& -E-  & 1.0$\pm 0.0$  & 7.6$\pm 6.65$  & 0.0  & 0.5 & 0.25\\
		& \setrow{\bfseries}-Er- & 11.64$\pm 4.38$  & 4.4$\pm 4.4$  &  2.74 & 5.82 & 2.91\\
		\midrule
		-G & \setrow{\bfseries}-r- & 9.25$\pm 3.77$  & 0.0$\pm 0.0$  &  0.0 & 9.25 & 3.4\\
		& -E-  & 1.0$\pm 0.0$  & 7.6$\pm 6.1$  &  0.03 & 1.0 & 0.37\\
		& -Er- & 8.46$\pm 5.12$  & 10$\pm 4.32$  &  3.07 & 8.46 & 3.11\\
		\bottomrule
	\end{tabular}
\end{table}

\begin{table}[h!]
	\caption{The estimates of the rates for $L = 0$, $d_l = 100$, $m = 1000$, $n = 2$, $h_{\rm{epoch}} = 1$, and \emph{cluster std }= 0.75}
	\label{tab::L0std75}
	\centering
	\begin{tabular}{>{\rowmac}l>{\rowmac}l>{\rowmac}l>{\rowmac}l>{\rowmac}l>{\rowmac}l>{\rowmac}l<{\clearrow}}
	\toprule
	& & \multicolumn{3}{l}{Optimal values}              
	& \multicolumn{2}{l}{Pre-defined values} \\
	\cmidrule(r){2-7}
	LIGHT & Configuration   & m($r$)$\pm$ sd($r$)   & m($E$)$\pm$ sd($E$)  & $H$ & $E^*$ & $H^*$\\
		\midrule
		-V & -r- & 7.26$\pm 5.87$  & 0.0$\pm 0.0$  &  0.0 & 3.63 &  1.82\\
		& -E-  & 1.0$\pm 0.0$  & 6.8$\pm 6.81$  & 0.0 &  0.5 & 0.25\\
		& \setrow{\bfseries}-Er- & 8.86$\pm 3.14$  & 10$\pm 5$  &  0.0 & 4.43 & 2.21\\
		\midrule
		-G & -r- & 6.07$\pm 4.3$  & 0.0$\pm 0.0$  &  0.0 & 6.07 & 2.23\\
		& -E-  & 1.0$\pm 0.0$  & 12$\pm 3.77$  &  0.0 & 1.0 & 0.37\\
		& \setrow{\bfseries}-Er- & 10.05$\pm 3.38$  & 9.2$\pm 5$  & 3.68 & 10.05 & 3.7\\
		\bottomrule
	\end{tabular}
\end{table}

\begin{table}[h!]
	\caption{The estimates of the rates for $L = 1$, $d_l = 100$, $m = 1000$, $n = 2$, $h_{\rm{epoch}} = 1$, and \emph{cluster std }= 0.5}
	\label{tab::L1std50}
	\centering
	\begin{tabular}{>{\rowmac}l>{\rowmac}l>{\rowmac}l>{\rowmac}l>{\rowmac}l>{\rowmac}l>{\rowmac}l<{\clearrow}}
		\toprule
		& & \multicolumn{3}{l}{Optimal values}              
		& \multicolumn{2}{l}{Pre-defined values} \\
		\cmidrule(r){2-7}
		LIGHT & Configuration   & m($r$)$\pm$ sd($r$)   & m($E$)$\pm$ sd($E$)  & $H$ & $E^*$ & $H^*$\\
		\midrule
		-V & -r- & 10.45$\pm 5.37$  & 0.0$\pm 0.0$  &  0.0 & 0.5 &  2.61\\
		& -E-  & 1.0$\pm 0.0$  & 8.8$\pm 4.92$  & 0.0 &  0.5 & 0.25\\
		& \setrow{\bfseries}-Er- & 10.45$\pm 4.28$  & 9.2$\pm 5$  & 1.1 & 5.22 & 2.61\\
		\midrule
		-G & -r- & 7.26$\pm 6.17$  & 0.0$\pm 0.0$  &  0.0 & 7.26 & 2.67\\
		& -E-  & 1.0$\pm 0.0$  & 8$\pm 4.99$  &  0.0 & 1.0 & 0.37\\
		& \setrow{\bfseries}-Er- & 12.44$\pm 4.38$  & 6$\pm 5.73$  &  3.7 & 12.4 & 4.58\\
		\bottomrule
	\end{tabular}
\end{table}

\begin{table}[h!]
	\caption{The estimates of the rates for $L = 1$, $d_l = 100$, $m = 1000$, $n = 2$, $h_{\rm{epoch}} = 1$, and \emph{cluster std }= 0.75}
	\label{tab::L1std75}
	\centering
	\begin{tabular}{>{\rowmac}l>{\rowmac}l>{\rowmac}l>{\rowmac}l>{\rowmac}l>{\rowmac}l>{\rowmac}l<{\clearrow}}
		\toprule
		& & \multicolumn{3}{l}{Optimal values}              
		& \multicolumn{2}{l}{Pre-defined values} \\
		\cmidrule(r){2-7}
		LIGHT & Configuration   & m($r$)$\pm$ sd($r$)   & m($E$)$\pm$ sd($E$)  & $H$ & $E^*$ & $H^*$\\
		\midrule
		-V & -r- & 9.65$\pm 5.99$  & 0.0$\pm 0.0$  &  0.0 & 4.83 &  2.41\\
		& -E-  & 1.0$\pm 0.0$  & 5.6$\pm 5.4$  &  0.0 &  0.5 & 0.25\\
		& \setrow{\bfseries}-Er- & 12$\pm 2.65$  & 10.4$\pm 5.4$   & 1.42 & 6.02 & 3.01\\
		\midrule
		-G & -r- & 7.66$\pm 5.12$  & 0.0$\pm 0.0$  &  0.0 & 7.66 & 2.82\\
		& -E-  & 1.0$\pm 0.0$  & 8$\pm 7.06$  &  0.0 & 1.0 & 0.37\\
		& \setrow{\bfseries}-Er- & 11.64$\pm 3.48$  & 9.6$\pm 6.31$  &   4.2 & 11.64 & 4.28\\
		\bottomrule
	\end{tabular}
\end{table}

\begin{table}[h!]
	\caption{The estimates of the rates for $L = 2$, $d_l = 100$, $m = 1000$, $n = 2$, $h_{\rm{epoch}} = 1$, and \emph{cluster std }= 0.25}
	\label{tab::L2std25}
	\centering
	\begin{tabular}{>{\rowmac}l>{\rowmac}l>{\rowmac}l>{\rowmac}l>{\rowmac}l>{\rowmac}l>{\rowmac}l<{\clearrow}}
		\toprule
		& & \multicolumn{3}{l}{Optimal values}              
		& \multicolumn{2}{l}{Pre-defined values} \\
		\cmidrule(r){2-7}
		LIGHT & Configuration   & m($r$)$\pm$ sd($r$)   & m($E$)$\pm$ sd($E$)  & $H$ & $E^*$ & $H^*$\\
		\midrule
		-V & -r- & 7.24$\pm 5.87$  & 0.0$\pm 0.0$  &  0.0 & 3.63 &  1.82\\
		& -E-  & 1.0$\pm 0.0$  & 7.6$\pm 6.38$  & 7.6  & 0.5 & 0.25\\
		& \setrow{\bfseries}-Er- & 8.86$\pm 3.14$  & 9.2$\pm 5$  &  0.0 & 4.43 & 2.21\\
		\midrule
		-G & -r- & 9.65$\pm 5.03$  & 0.0$\pm 0.0$  &  0.0 & 9.65 & 3.55\\
		& -E-  & 1.0$\pm 0.0$  & 9.6$\pm 5.4$  &  0.0 & 1.0 & 0.37\\
		& \setrow{\bfseries}-Er- & 12.44$\pm 3.96$  & 6.4$\pm 4.7$  &  3.83 & 12.44 & 4.58\\
		\bottomrule
	\end{tabular}
\end{table}

\begin{table}[h!]
	\caption{The estimates of the rates for $L = 2$, $d_l = 100$, $m = 1000$, $n = 2$, $h_{\rm{epoch}} = 1$, and \emph{cluster std }= 0.50}
	\label{tab::L2std50}
	\centering
	\begin{tabular}{>{\rowmac}l>{\rowmac}l>{\rowmac}l>{\rowmac}l>{\rowmac}l>{\rowmac}l>{\rowmac}l<{\clearrow}}
		\toprule
		& & \multicolumn{3}{l}{Optimal values}              
		& \multicolumn{2}{l}{Pre-defined values} \\
		\cmidrule(r){2-7}
		LIGHT & Configuration   & m($r$)$\pm$ sd($r$)   & m($E$)$\pm$ sd($E$)  & $H$ & $E^*$ & $H^*$\\
		\midrule
		-V & -r- & 8.86$\pm 5.24$  & 0.0$\pm 0.0$  &  0 & 4.43 &  2.21\\
		& -E-  & 1.0$\pm 0.0$  & 8$\pm 7.3$  & 0.0 &  0.5 & 0.25\\
		& \setrow{\bfseries}-Er- & 10.85$\pm 3.78$  & 7.6$\pm 5.8$  &   2.27 & 5.42 & 2.71\\
		\midrule
		-G & -r- & 8.06$\pm 4.6$  & 0.0$\pm 0.0$  &  0.0 & 8.06 & 2.97\\
		& -E-  & 1.0$\pm 0.0$  & 6.8$\pm 6.27$  &  0.0 & 1.0 & 0.37\\
		& \setrow{\bfseries}-Er- & 10.85$\pm 4.22$  & 10$\pm 5.08$  &   3.98 & 10.85 & 3.99\\
		\bottomrule
	\end{tabular}
\end{table}

\begin{table}[h!]
	\caption{The estimates of the rates for $L = 2$, $d_l = 100$, $m = 1000$, $n = 2$, $h_{\rm{epoch}} = 1$, and \emph{cluster std }= 0.75}
	\label{tab::L2std75}
	\centering
	\begin{tabular}{>{\rowmac}l>{\rowmac}l>{\rowmac}l>{\rowmac}l>{\rowmac}l>{\rowmac}l>{\rowmac}l<{\clearrow}}
		\toprule
		& & \multicolumn{3}{l}{Optimal values}              
		& \multicolumn{2}{l}{Pre-defined values} \\
		\cmidrule(r){2-7}
		LIGHT & Configuration   & m($r$)$\pm$ sd($r$)   & m($E$)$\pm$ sd($E$)  & $H$ & $E^*$ & $H^*$\\
		\midrule
		-V & -r- & 8.06$\pm 6.22$  & 0.0$\pm 0.0$  &  0.0 & 4.03 &  2.01\\
		& -E-  & 1.0$\pm 0.0$  & 6.8$\pm 5.98$  & 0.0 &  0.5 & 0.25\\
		& \setrow{\bfseries}-Er- & 10.85$\pm 8$  & 7.06$\pm 5.8$  &  2.1 & 5.42 & 2.71\\
		\midrule
		-G & -r- & 8.06$\pm 4.2$  & 0.0$\pm 0.0$  &  0.0 & 8.06 & 2.97\\
		& -E-  & 1.0$\pm 0.0$  & 8.4$\pm 6.92$  &  0.0 & 1.0 & 0.37\\
		& \setrow{\bfseries}-Er- & 11.24$\pm 4.11$  & 7.6$\pm 6.65$  &  3.87 & 11.24 & 4.14\\
		\bottomrule
	\end{tabular}
\end{table}

\begin{table}[h!]
	\caption{The estimates of the rates for $L = 3$, $d_l = 100$, $m = 1000$, $n = 2$, $h_{\rm{epoch}} = 1$, and \emph{cluster std }= 0.25}
	\label{tab::L3std25}
	\centering
	\begin{tabular}{>{\rowmac}l>{\rowmac}l>{\rowmac}l>{\rowmac}l>{\rowmac}l>{\rowmac}l>{\rowmac}l<{\clearrow}}
		\toprule
		& & \multicolumn{3}{l}{Optimal values}              
		& \multicolumn{2}{l}{Pre-defined values} \\
		\cmidrule(r){2-7}
		LIGHT & Configuration   & m($r$)$\pm$ sd($r$)   & m($E$)$\pm$ sd($E$)  & $H$ & $E^*$ & $H^*$\\
		\midrule
		-V & -r- & 10.05$\pm 3.38$  & 0.0$\pm 0.0$  &  0.0 & 5.03 &  2.51\\
		& -E-  & 1.0$\pm 0.0$  & 6$\pm 6.32$  & 0.0 &  0.5 & 0.25\\
		& \setrow{\bfseries}-Er- & 12.84$\pm 4.11$  & 6.4$\pm 5.4$  &  3.2 & 6.42 & 3.2\\
		\midrule
		-G & -r- & 4.48$\pm 5.12$  & 0.0$\pm 0.0$  &  0.0 & 4.48 & 1.65\\
		& -E-  & 1.0$\pm 0.0$  & 5.6$\pm 4.3$  &  0.02 & 1.0 & 0.37\\
		& \setrow{\bfseries}-Er- & 8.06$\pm 4.97$  & 6.4$\pm 5.72$  &   2.89 & 8.06 & 2.97\\
		\bottomrule
	\end{tabular}
\end{table}

\begin{table}[h!]
	\caption{The estimates of the rates for $L = 3$, $d_l = 100$, $m = 1000$, $n = 2$, $h_{\rm{epoch}} = 1$, and \emph{cluster std }= 0.5}
	\label{tab::L3std50}
	\centering
	\begin{tabular}{>{\rowmac}l>{\rowmac}l>{\rowmac}l>{\rowmac}l>{\rowmac}l>{\rowmac}l>{\rowmac}l<{\clearrow}}
		\toprule
		& & \multicolumn{3}{l}{Optimal values}              
		& \multicolumn{2}{l}{Pre-defined values} \\
		\cmidrule(r){2-7}
		LIGHT & Configuration   & m($r$)$\pm$ sd($r$)   & m($E$)$\pm$ sd($E$)  & $H$ & $E^*$ & $H^*$\\
		\midrule
		-V & -r- & 10.05$\pm 3.38$  & 0.0$\pm 0.0$  &  0 & 3.83 &  1.92\\
		& -E-  & 1.0$\pm 0.0$  & 6.8$\pm 5.98$  & 0.0 &  0.5 & 0.25\\
		& \setrow{\bfseries}-Er- & 10.05$\pm 2.81$  & 8.0$\pm 6.25$  &  1.63 & 5.025 & 2.51\\
		\midrule
		-G & -r- & 10.05$\pm 3.87$  & 0.0$\pm 0.0$  &  0.0 & 10.0 & 3.7\\
		& -E-  & 1.0$\pm 0.0$  & 9.2$\pm 5.35$  &  0.0 & 1.0 & 0.37\\
		& \setrow{\bfseries}-Er- & 12.44$\pm 2.26$  & 9.6$\pm 5.4$  &  4.44 & 12.44 & 4.58\\
		\bottomrule
	\end{tabular}
\end{table}

\begin{table}[h!]
	\caption{The estimates of the rates for $L = 3$, $d_l = 100$, $m = 1000$, $n = 2$, $h_{\rm{epoch}} = 1$, and \emph{cluster std }= 0.75}
	\label{tab::L3std75}
	\centering
	\begin{tabular}{>{\rowmac}l>{\rowmac}l>{\rowmac}l>{\rowmac}l>{\rowmac}l>{\rowmac}l>{\rowmac}l<{\clearrow}}
		\toprule
		& & \multicolumn{3}{l}{Optimal values}              
		& \multicolumn{2}{l}{Pre-defined values} \\
		\cmidrule(r){2-7}
		LIGHT & Configuration   & m($r$)$\pm$ sd($r$)   & m($E$)$\pm$ sd($E$)  & $H$ & $E^*$ & $H^*$\\
		\midrule
		-V & -r- & 6.47$\pm 6.82$  & 0.0$\pm 0.0$  &  0.0 & 3.23 &  1.62\\
		& -E-  & 1.0$\pm 0.0$  & 7.2$\pm 5.59$  & 0.0 &  0.5 & 0.25\\
		& \setrow{\bfseries}-Er- & 11.24$\pm 2.52$  & 11.6$\pm 4.4$  &  0 & 5.62 & 2.81\\
		\midrule
		-G & -r- & 7.67$\pm 6.07$  & 0.0$\pm 0.0$  &  0.0 & 7.66 & 2.82\\
		& -E-  & 1.0$\pm 0.0$  & 7.6$\pm 5.15$  &  0.0 & 1.0 & 0.37\\
		& \setrow{\bfseries}-Er- & 14.03$\pm 3.87$  & 8.8$\pm 6.75$  &  4.7 & 14.03 & 5.16\\
		\bottomrule
	\end{tabular}
\end{table}

\begin{figure}[h!]
	\centering
	\includegraphics[width=0.85\columnwidth]{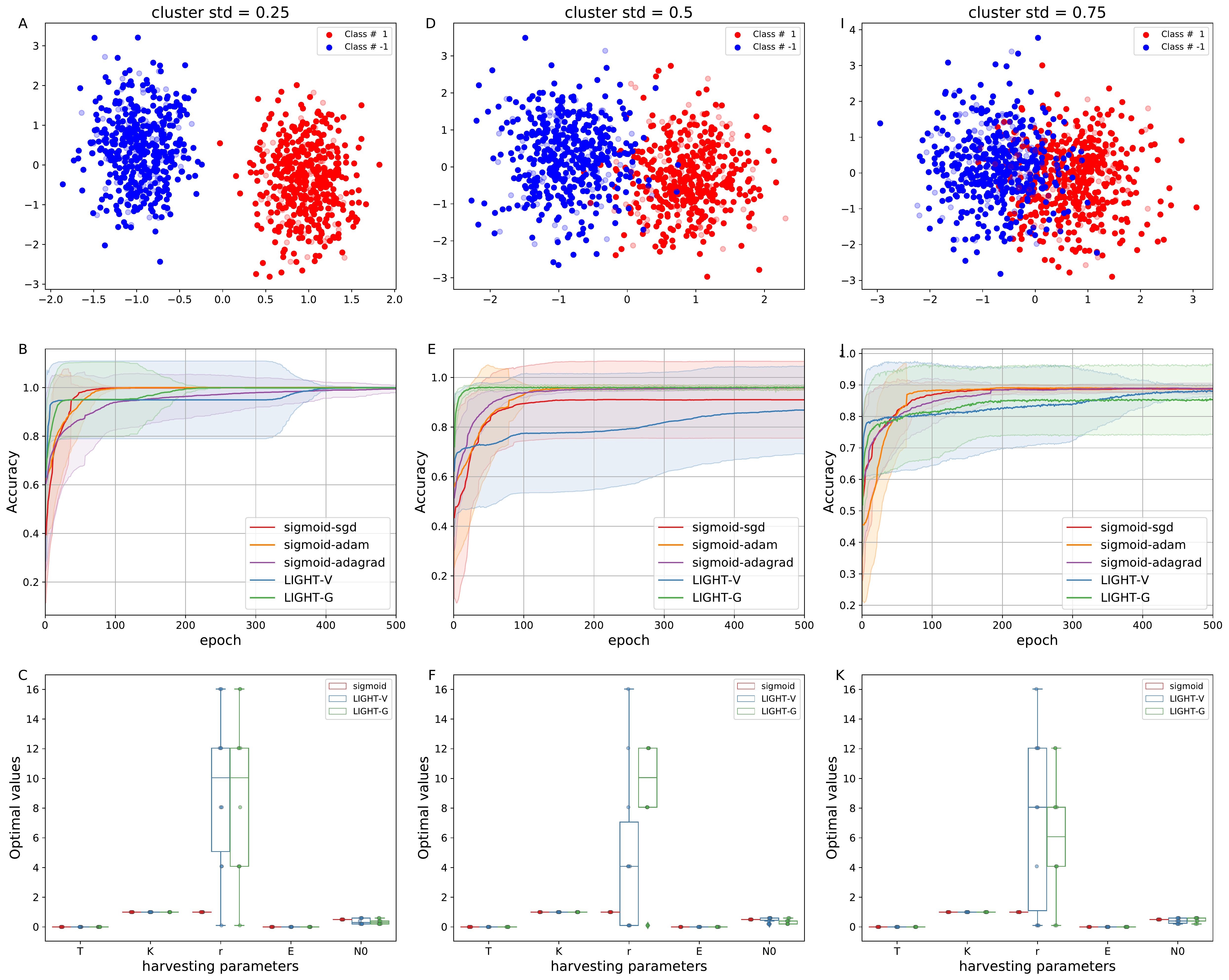}
	\caption{The accuracy curves on testing for $L = 0$, $d_l = 100$, $m = 1000$, $n = 2$, $h_{\rm{epoch}} = 1$, and \emph{cluster std} = \{0.25, 0.5, 0.75\}: The \textbf{-r-} configuration}
	\label{fig::L0std-r}
\end{figure}

\begin{figure}[h!]
	\centering
	\includegraphics[width=0.85\columnwidth]{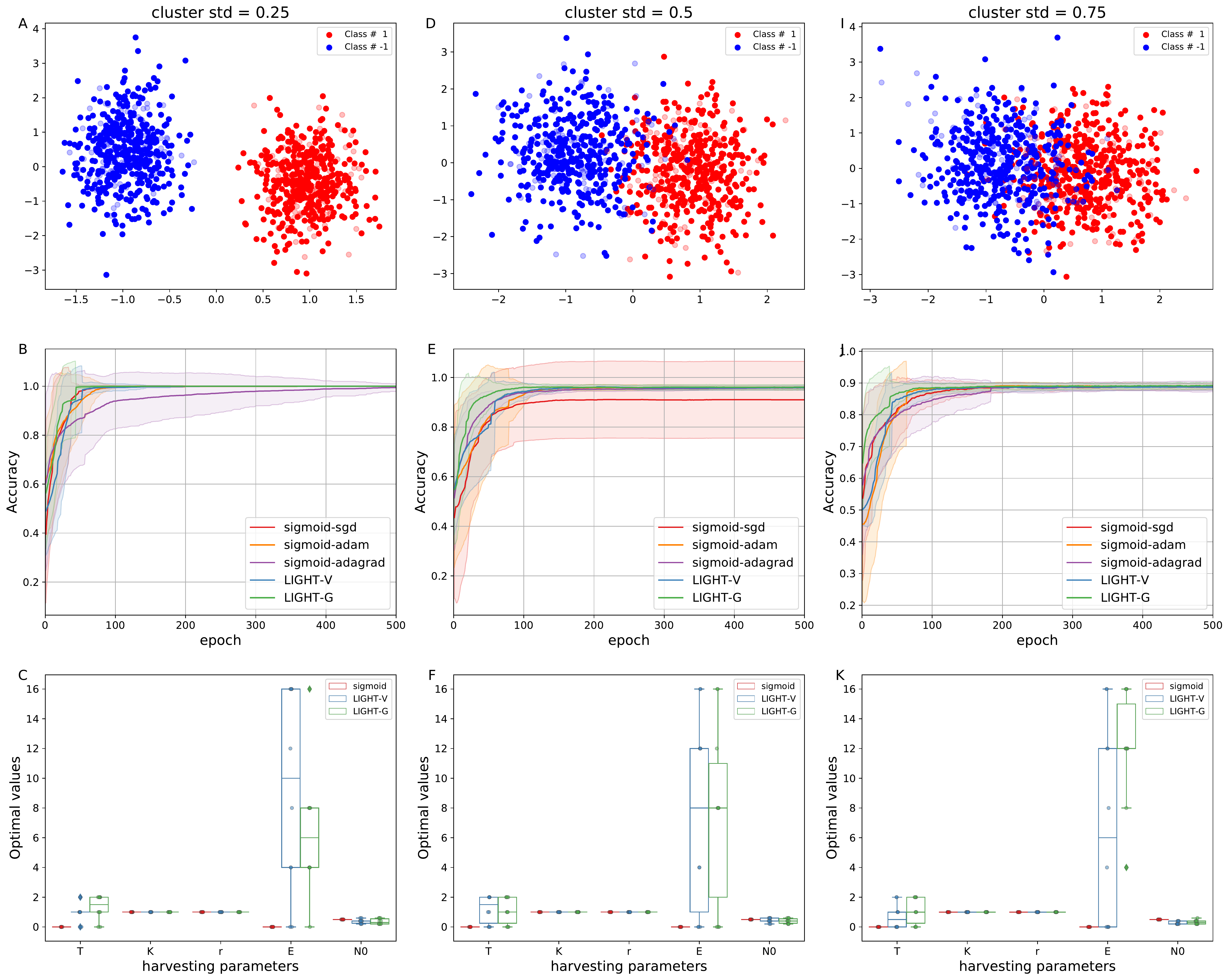}
	\caption{The accuracy curves on testing for $L = 0$, $d_l = 100$, $m = 1000$, $n = 2$, $h_{\rm{epoch}} = 1$, and \emph{cluster std} = \{0.25, 0.5, 0.75\}: The \textbf{ -E-} configuration}
	\label{fig::L0std-E}
\end{figure}

\begin{figure}[h!]
	\centering
	\includegraphics[width=0.85\columnwidth]{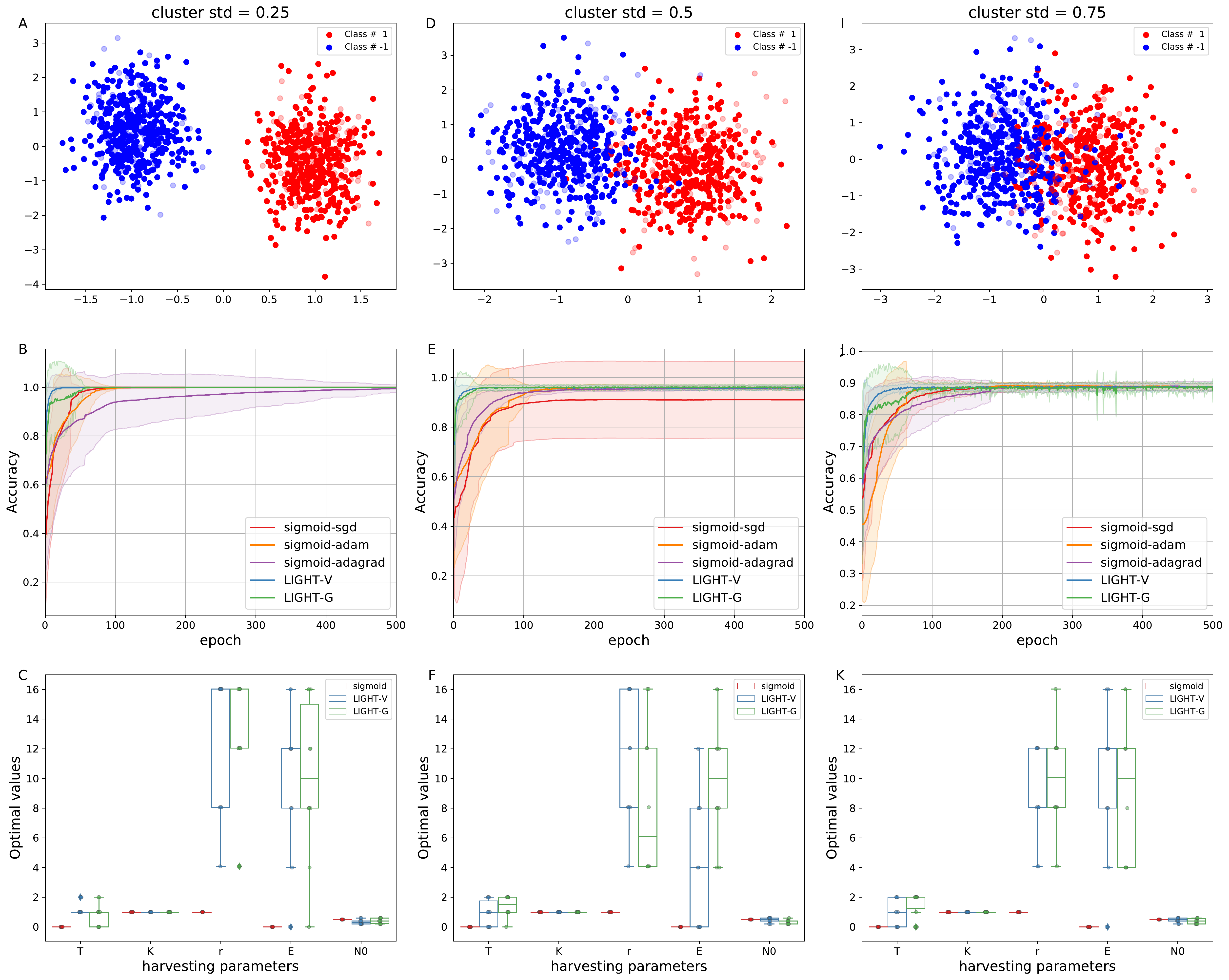}
	\caption{The accuracy curves on testing for $L = 0$, $d_l = 100$, $m = 1000$, $n = 2$, $h_{\rm{epoch}} = 1$, and \emph{cluster std }= \{0.25, 0.5, 0.75\}: The \textbf{-Er-} configuration}
	\label{fig::L0std-Er}
\end{figure}

\begin{figure}[h!]
	\centering
	\includegraphics[width=0.85\columnwidth]{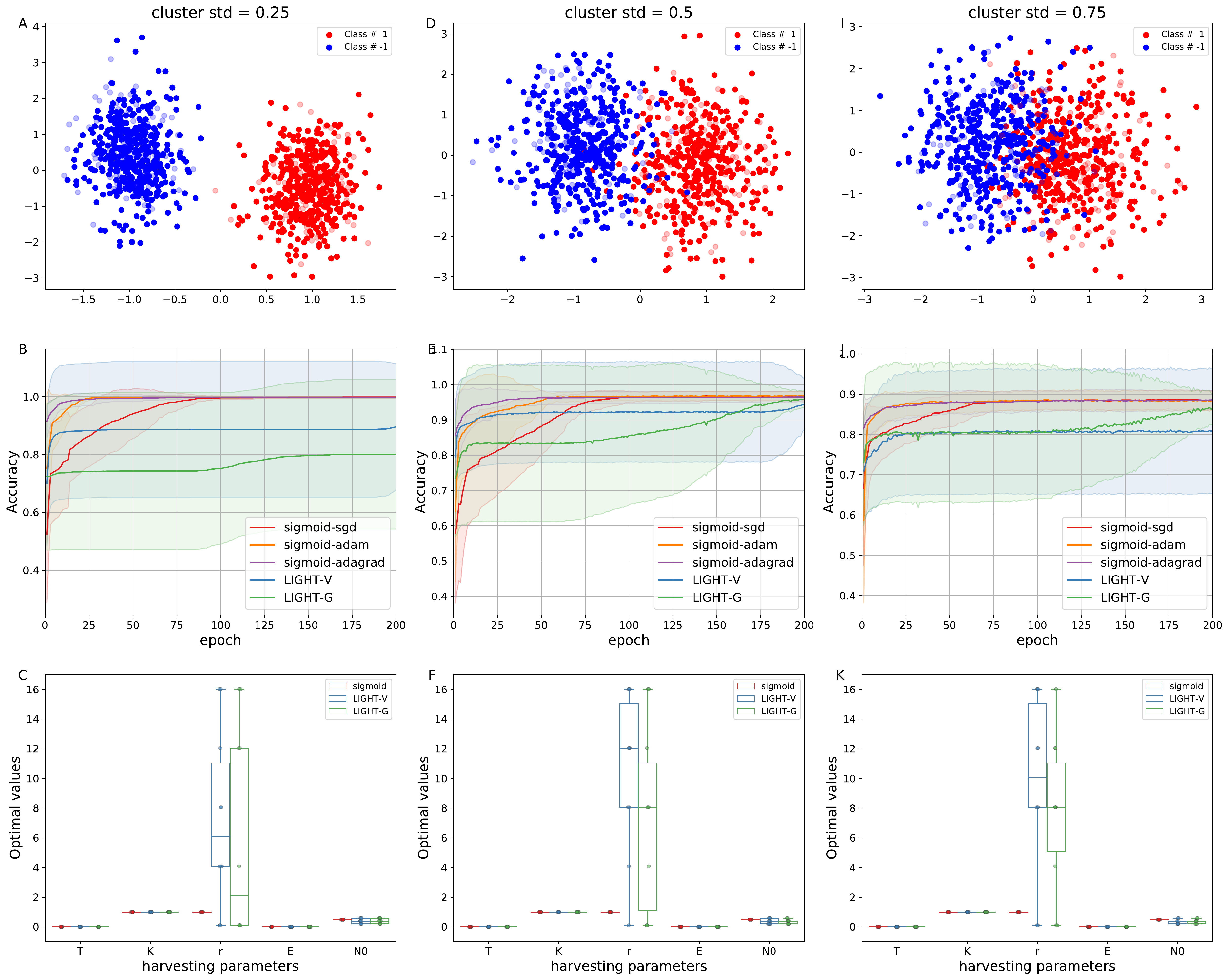}
	\caption{The accuracy curves on testing for $L = 1$, $d_l = 100$, $m = 1000$, $n = 2$, $h_{\rm{epoch}} = 1$, and \emph{cluster std} = \{0.25, 0.5, 0.75\}: The \textbf{-r-} configuration}
	\label{fig::L1std-r}
\end{figure}

\begin{figure}[h!]
	\centering
	\includegraphics[width=0.85\columnwidth]{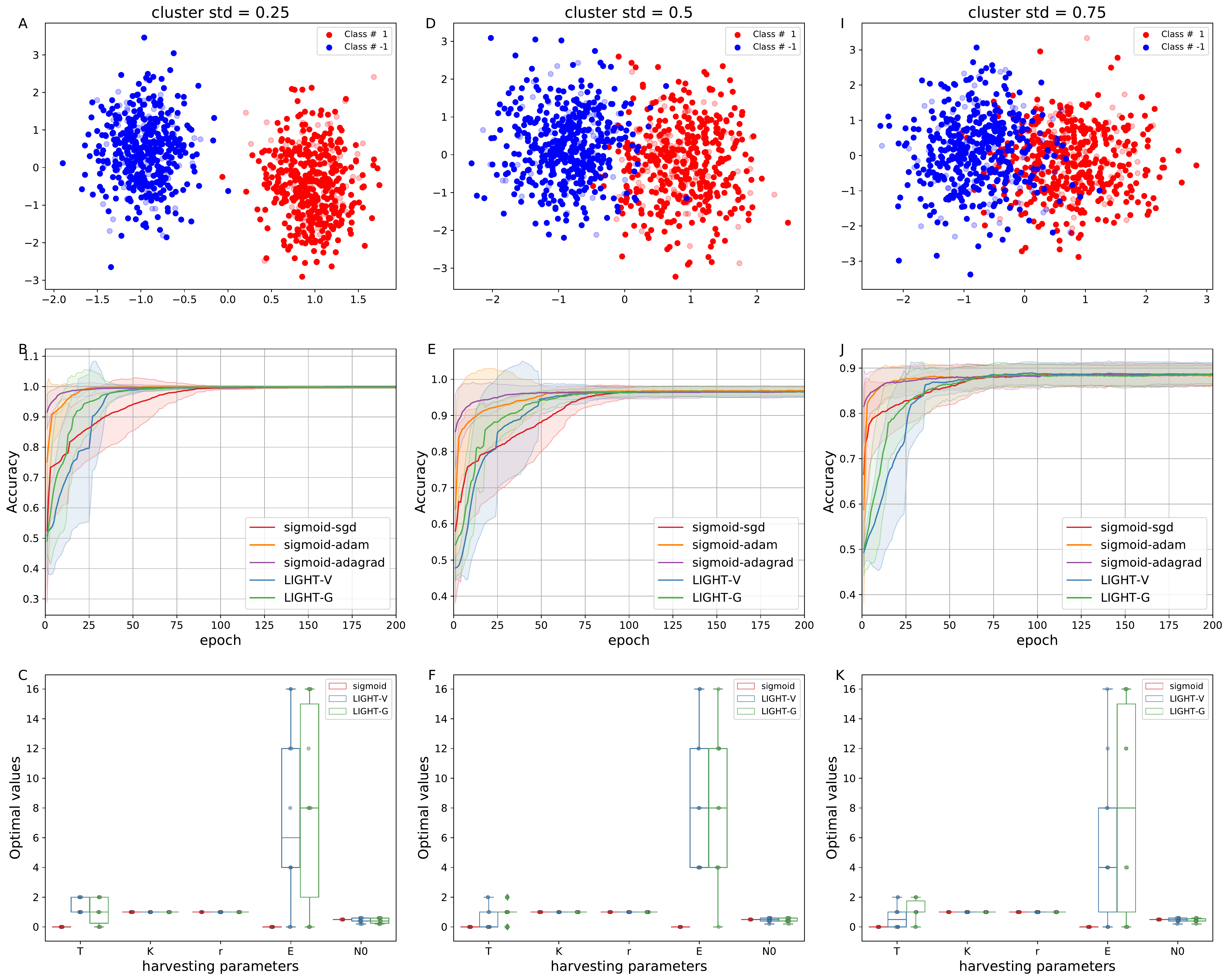}
	\caption{The accuracy curves on testing for $L = 1$, $d_l = 100$, $m = 1000$, $n = 2$, $h_{\rm{epoch}} = 1$, and \emph{cluster std} = \{0.25, 0.5, 0.75\}: The \textbf{-E-} configuration}
	\label{fig::L1std-E}
\end{figure}

\begin{figure}[h!]
	\centering
	\includegraphics[width=0.85\columnwidth]{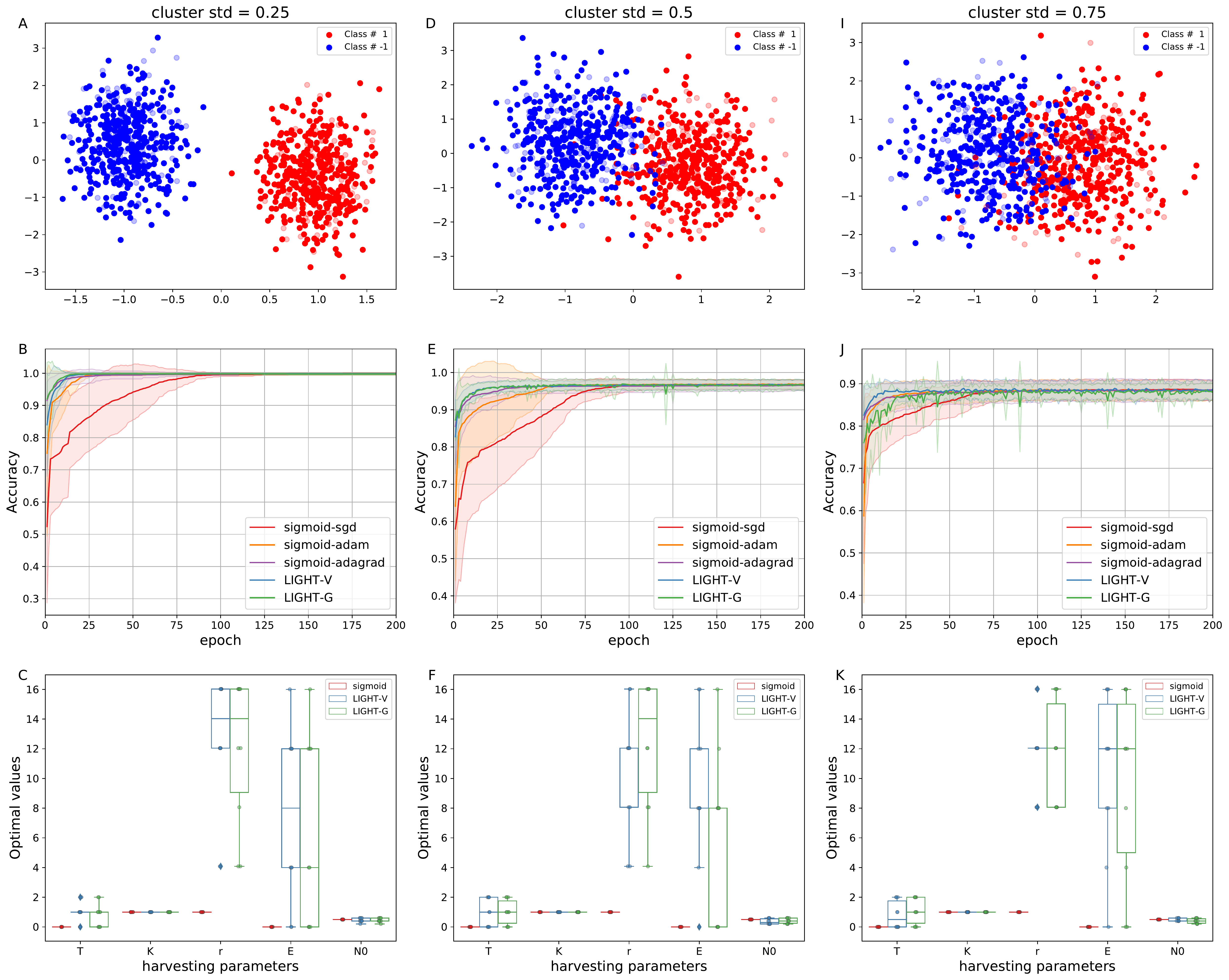}
	\caption{The accuracy curves on testing for $L = 1$, $d_l = 100$, $m = 1000$, $n = 2$, $h_{\rm{epoch}} = 1$, and \emph{cluster std} = \{0.25, 0.5, 0.75\}: The \textbf{-Er-} configuration}
	\label{fig::L1std-Er}
\end{figure}

\begin{figure}[h!]
	\centering
	\includegraphics[width=0.85\columnwidth]{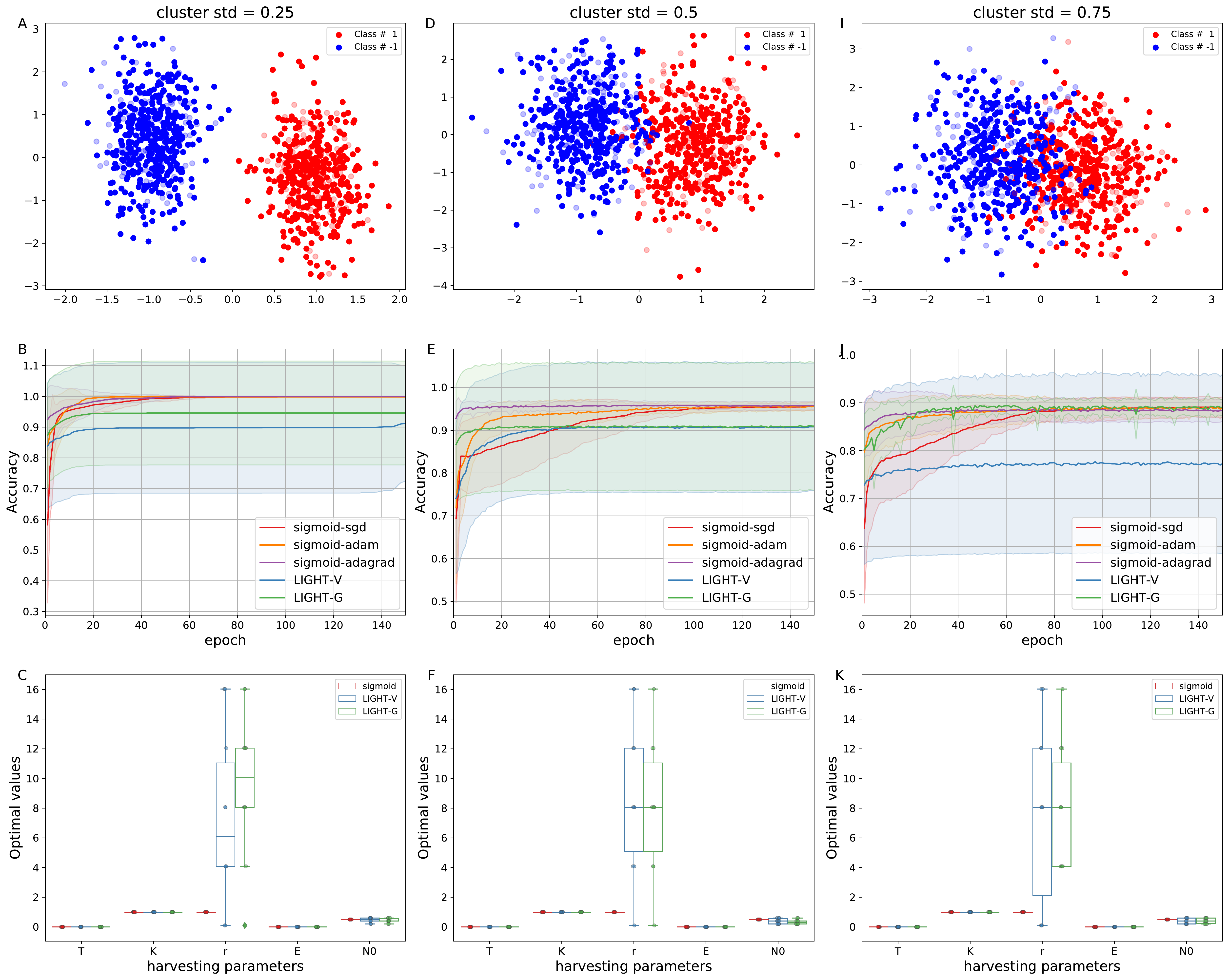}
	\caption{The accuracy curves on testing for $L = 2$, $d_l = 100$, $m = 1000$, $n = 2$, $h_{\rm{epoch}} = 1$, and \emph{cluster std} = \{0.25, 0.5, 0.75\}: The \textbf{-r-} configuration}
	\label{fig::L2std-r}
\end{figure}

\begin{figure}[h!]
	\centering
	\includegraphics[width=0.85\columnwidth]{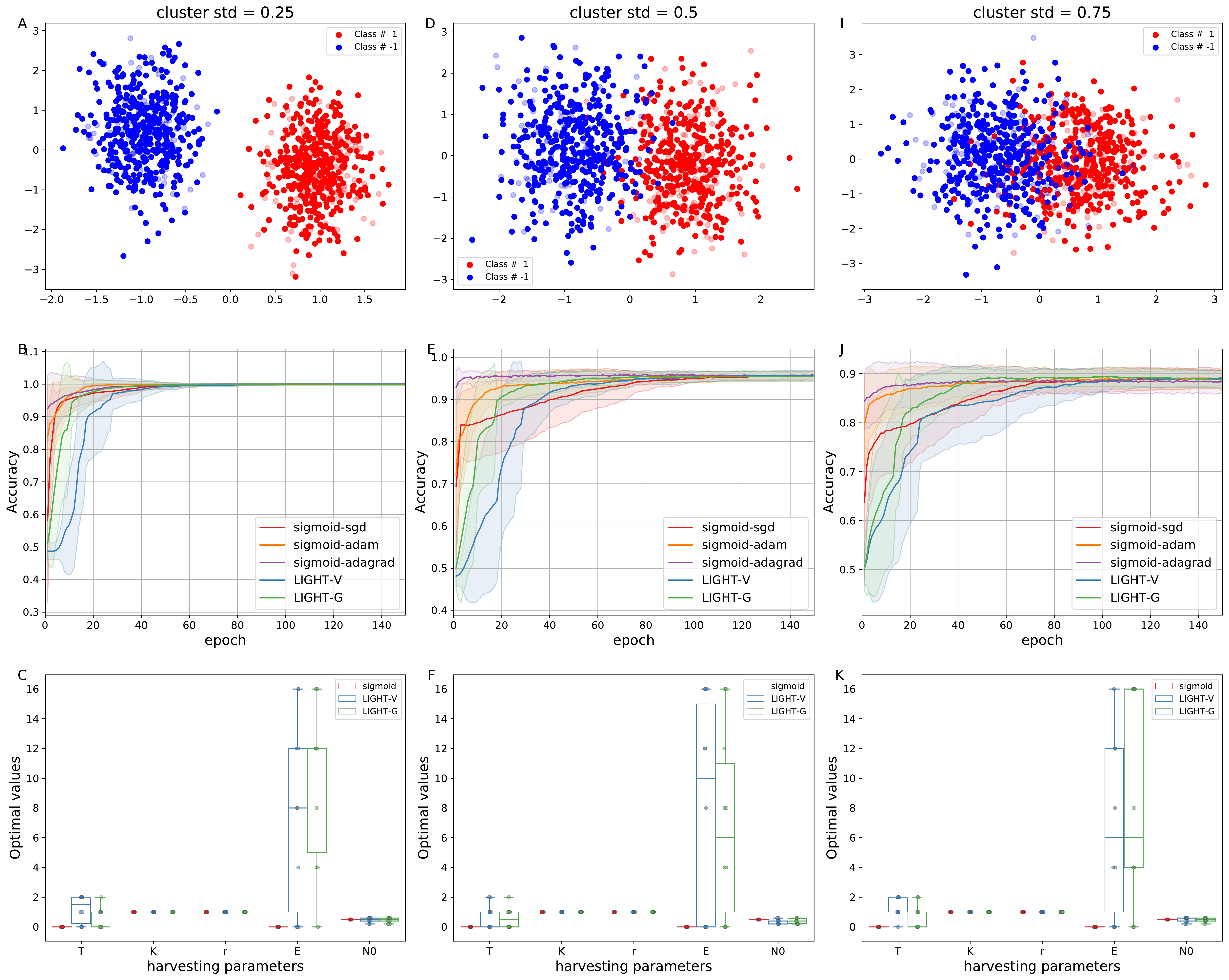}
	\caption{The accuracy curves on testing for $L = 2$, $d_l = 100$, $m = 1000$, $n = 2$, $h_{\rm{epoch}} = 1$, and \emph{cluster std} = \{0.25, 0.5, 0.75\}: The \textbf{-E-} configuration}
	\label{fig::L2std-E}
\end{figure}

\begin{figure}[h!]
	\centering
	\includegraphics[width=0.85\columnwidth]{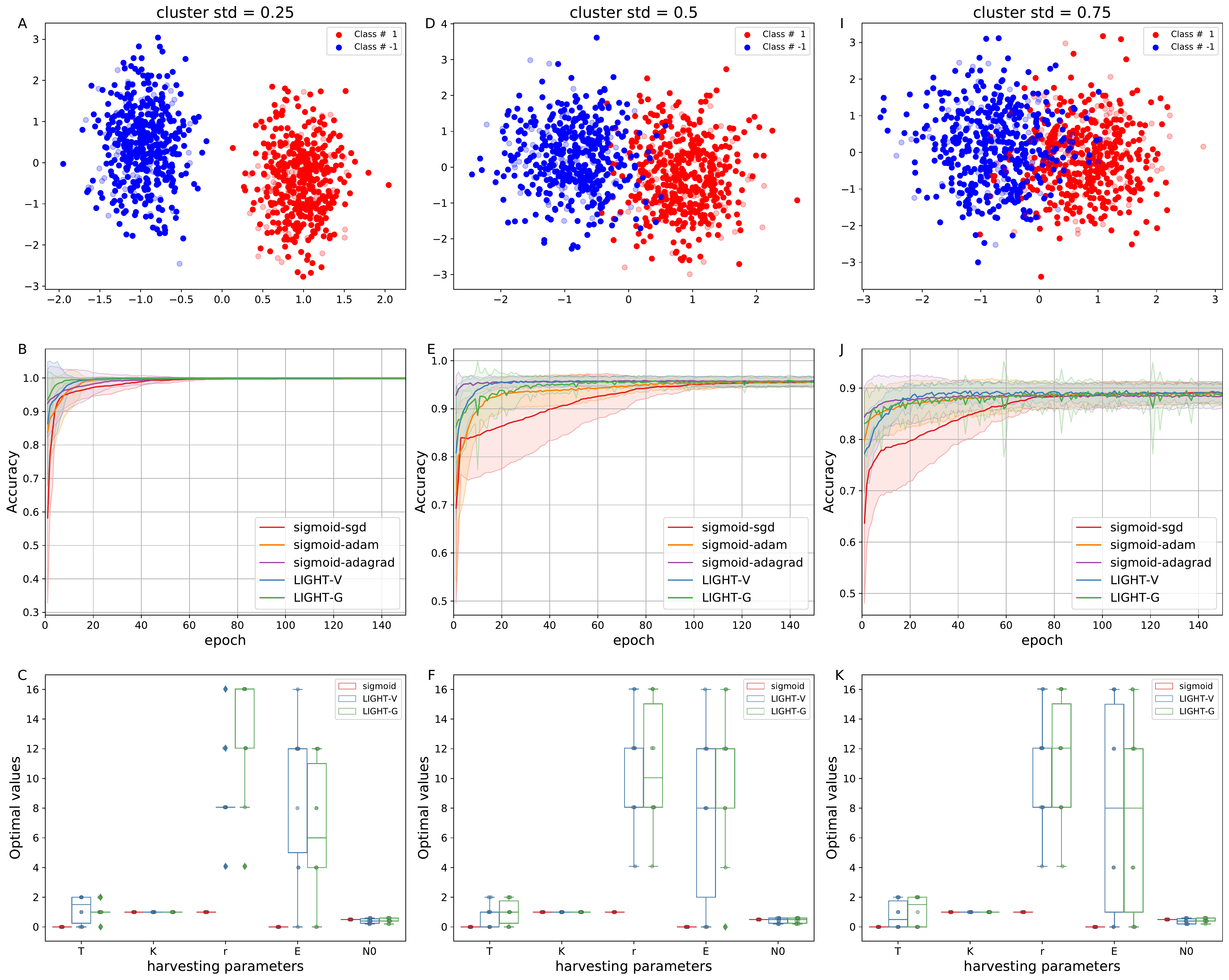}
	\caption{The accuracy curves on testing for $L = 2$, $d_l = 100$, $m = 1000$, $n = 2$,  $h_{\rm{epoch}} = 1$, and \emph{cluster std = \{0.25, 0.5, 0.75\}}: The \textbf{-Er-} configuration}
	\label{fig::L2std-Er}
\end{figure}

\begin{figure}[h!]
	\centering
	\includegraphics[width=0.85\columnwidth]{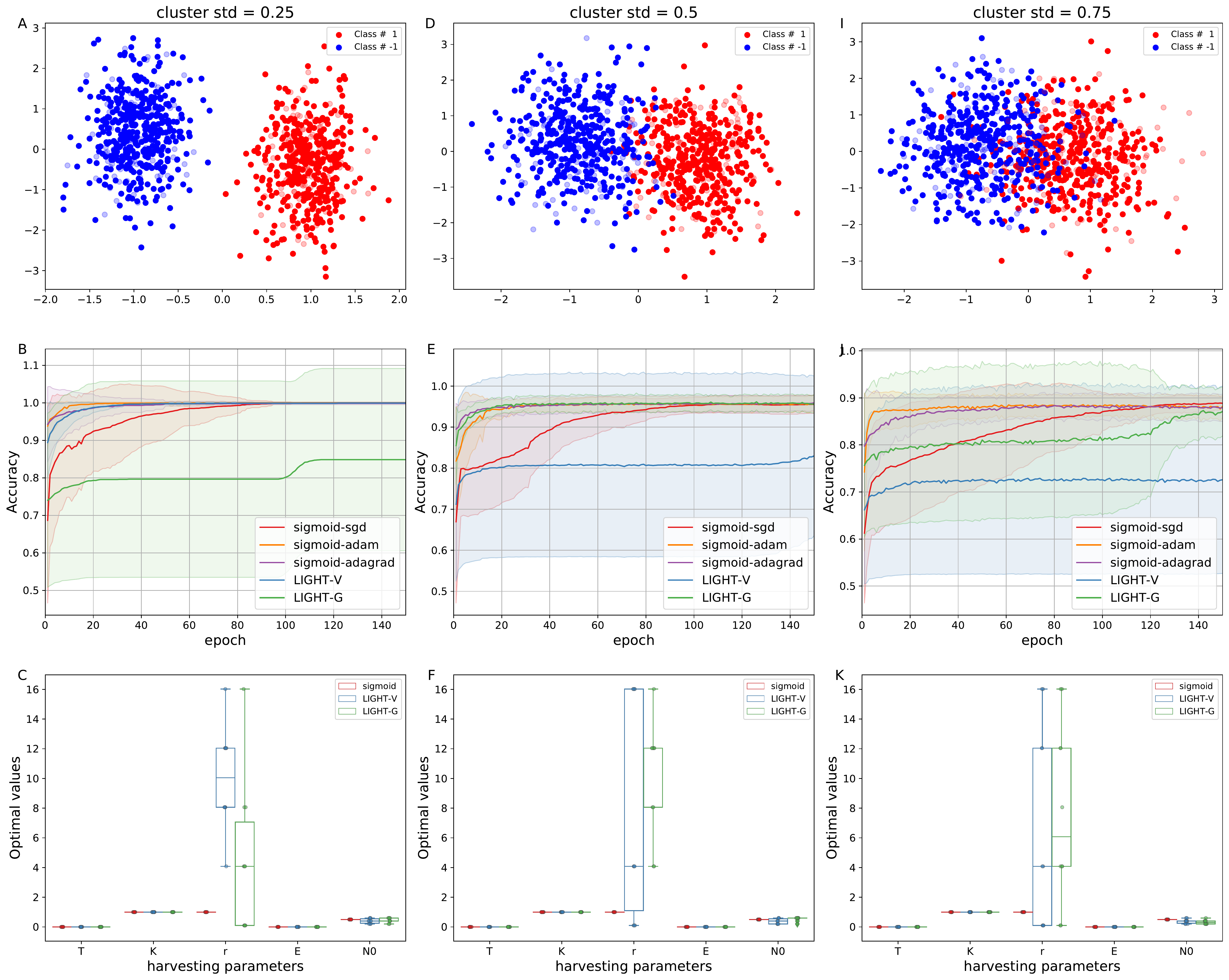}
	\caption{The accuracy curves on testing for $L = 3$, $d_l = 100$, $m = 1000$, $n = 2$,  $h_{\rm{epoch}} = 1$, and \emph{cluster std }= \{0.25, 0.5, 0.75\}: The \textbf{-r-} configuration}
	\label{fig::L3std-r}
\end{figure}

\begin{figure}[h!]
	\centering
	\includegraphics[width=0.85\columnwidth]{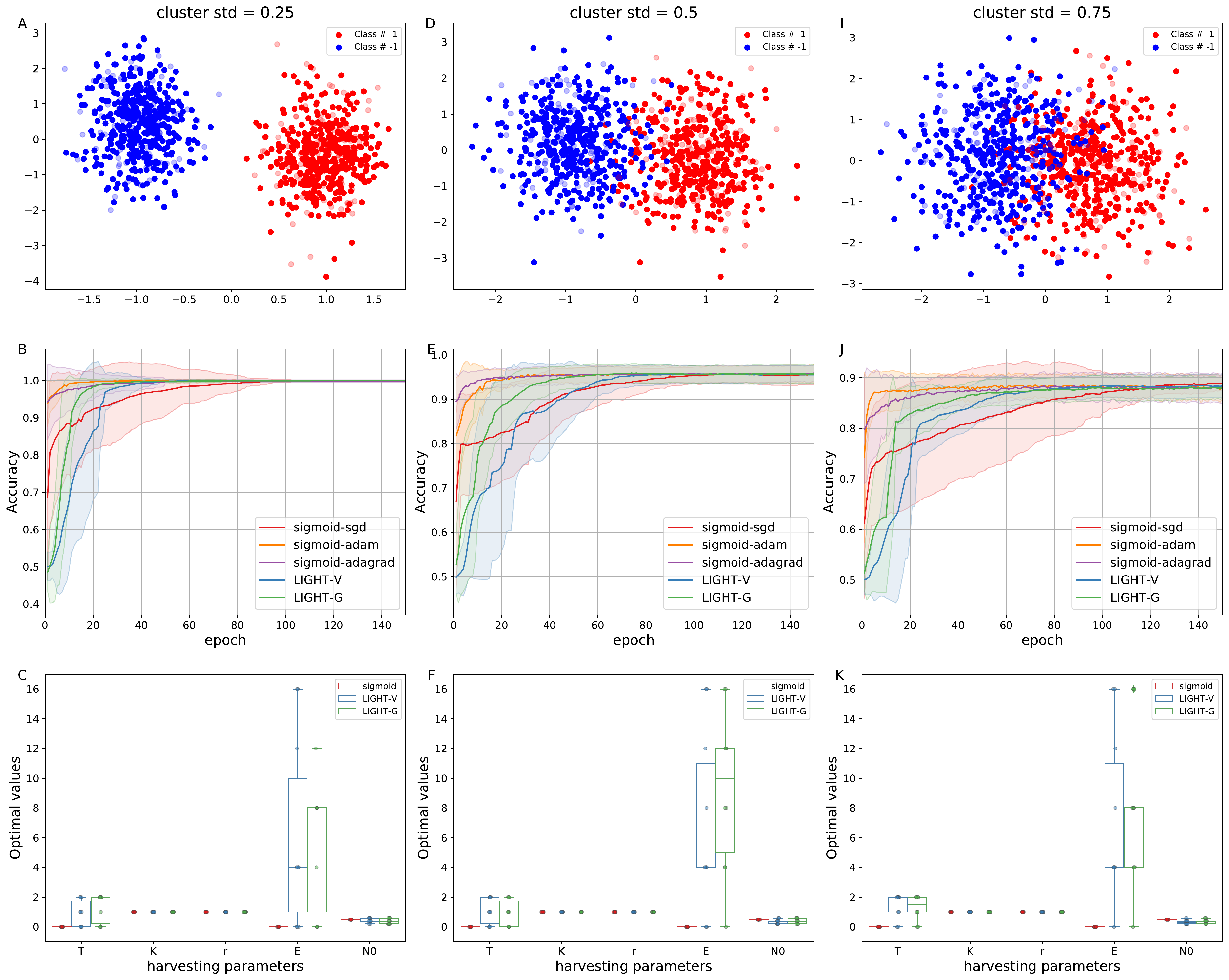}
	\caption{The accuracy curves on testing for $L = 3$, $d_l = 100$, $m = 1000$, $n = 2$, $h_{\rm{epoch}} = 1$, and \emph{cluster std }= \{0.25, 0.5, 0.75\}: The \textbf{-E-} configuration}
	\label{fig::L3std-E}
\end{figure}

\begin{figure}[h!]
	\centering
	\includegraphics[width=0.85\columnwidth]{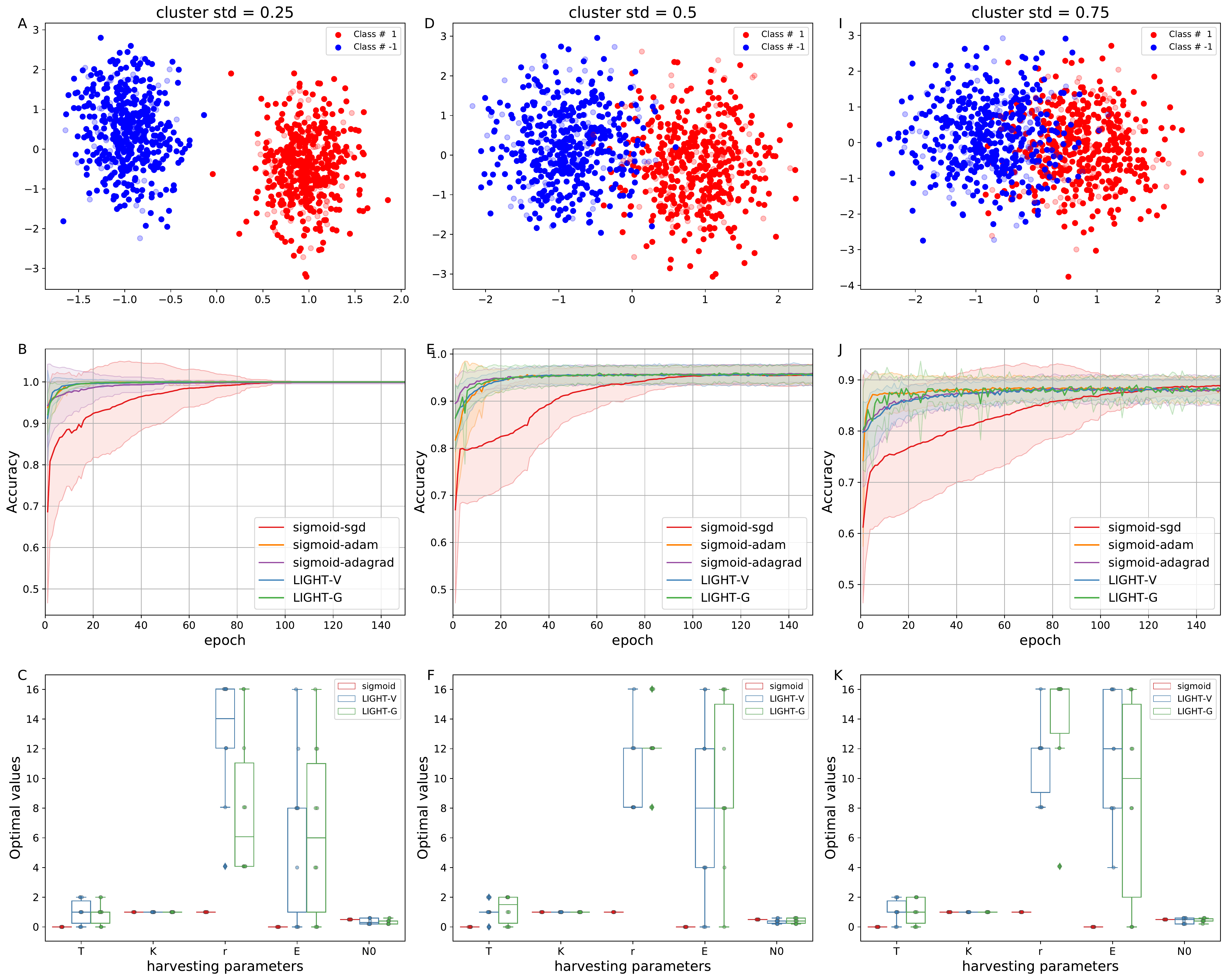}
	\caption{The accuracy curves on testing for $L = 3$, $d_l = 100$, $m = 1000$, $n = 2$,  $h_{\rm{epoch}} = 1$  and \emph{cluster std = \{0.25, 0.5, 0.75\}}: The \textbf{-Er-} configuration}
	\label{fig::L3std-Er}
\end{figure}

\clearpage
\begin{figure}[h!]
	\centering
	\includegraphics[width=0.95\columnwidth]{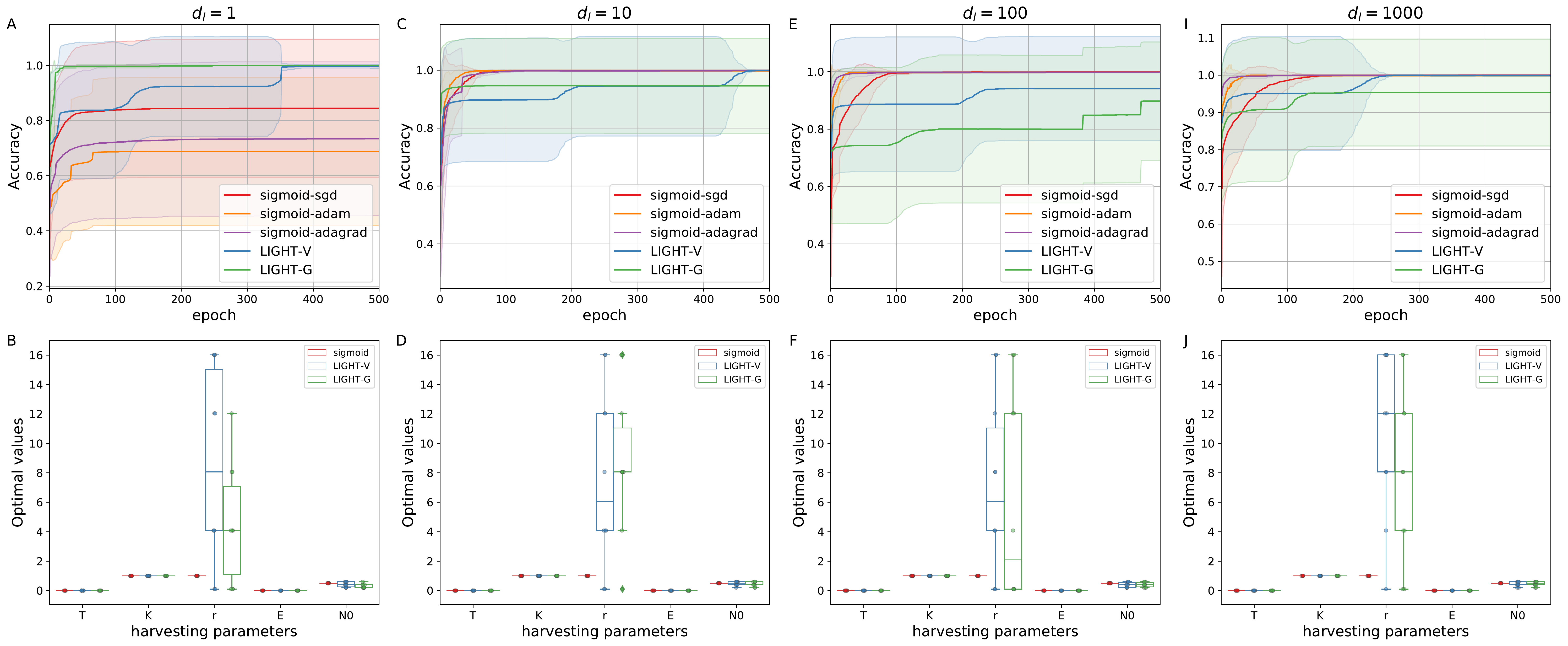}
	\caption{The accuracy curves on testing for $L = 1$, $m = 1000$, $n = 2$,  \emph{cluster std} = 0.25, $h_{\rm{epoch}} = 1$, and $d_l = \{1, 10, 100, 1000\}$: The \textbf{-r-} configuration}
	\label{fig::dl-r}
\end{figure}

\begin{figure}[h!]
	\centering
	\includegraphics[width=0.95\columnwidth]{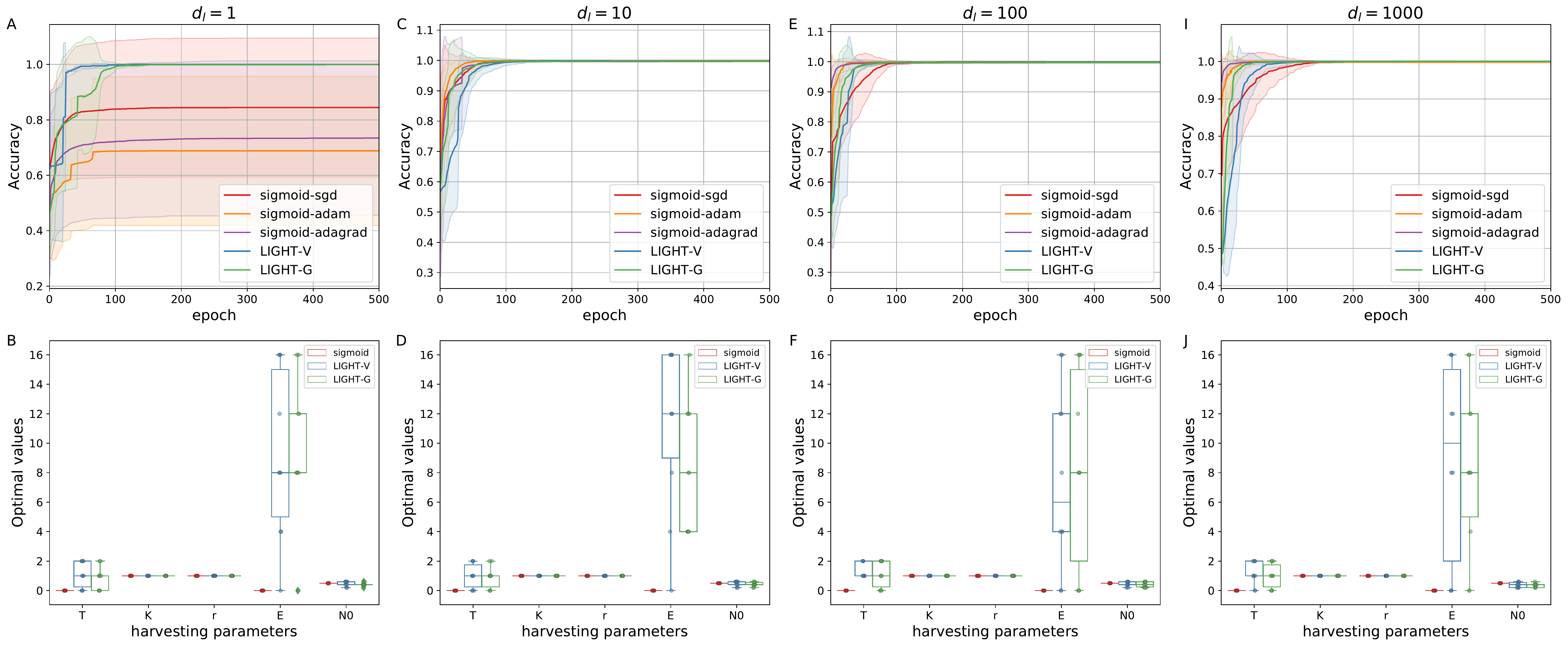}
	\caption{The accuracy curves on testing for $L = 1$, $m = 1000$,  $n = 2$,  \emph{cluster std} = 0.25, $h_{\rm{epoch}} = 1$, and $d_l = \{1, 10, 100, 1000\}$: The \textbf{-E-} configuration}
	\label{fig::dl-E}
\end{figure}

\begin{figure}[h!]
	\centering
	\includegraphics[width=0.95\columnwidth]{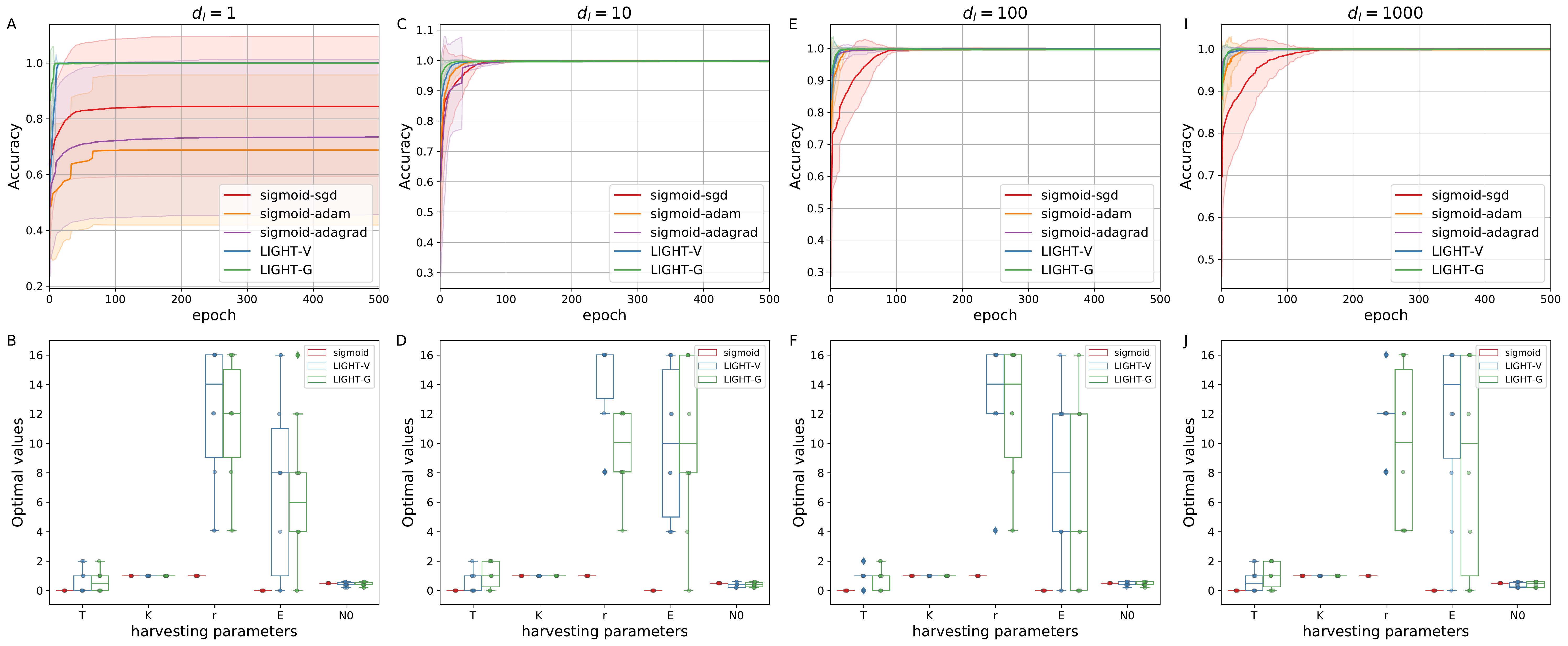}
	\caption{The accuracy curves on testing for $L = 1$, $m = 1000$, $n = 2$,  \emph{cluster std} = 0.25, $h_{\rm{epoch}} = 1$, and $d_l = \{1, 10, 100, 1000\}$: The \textbf{-Er-} configuration}
	\label{fig::dl-Er}
\end{figure}

\begin{figure}[h!]
	\centering
	\includegraphics[width=0.95\columnwidth]{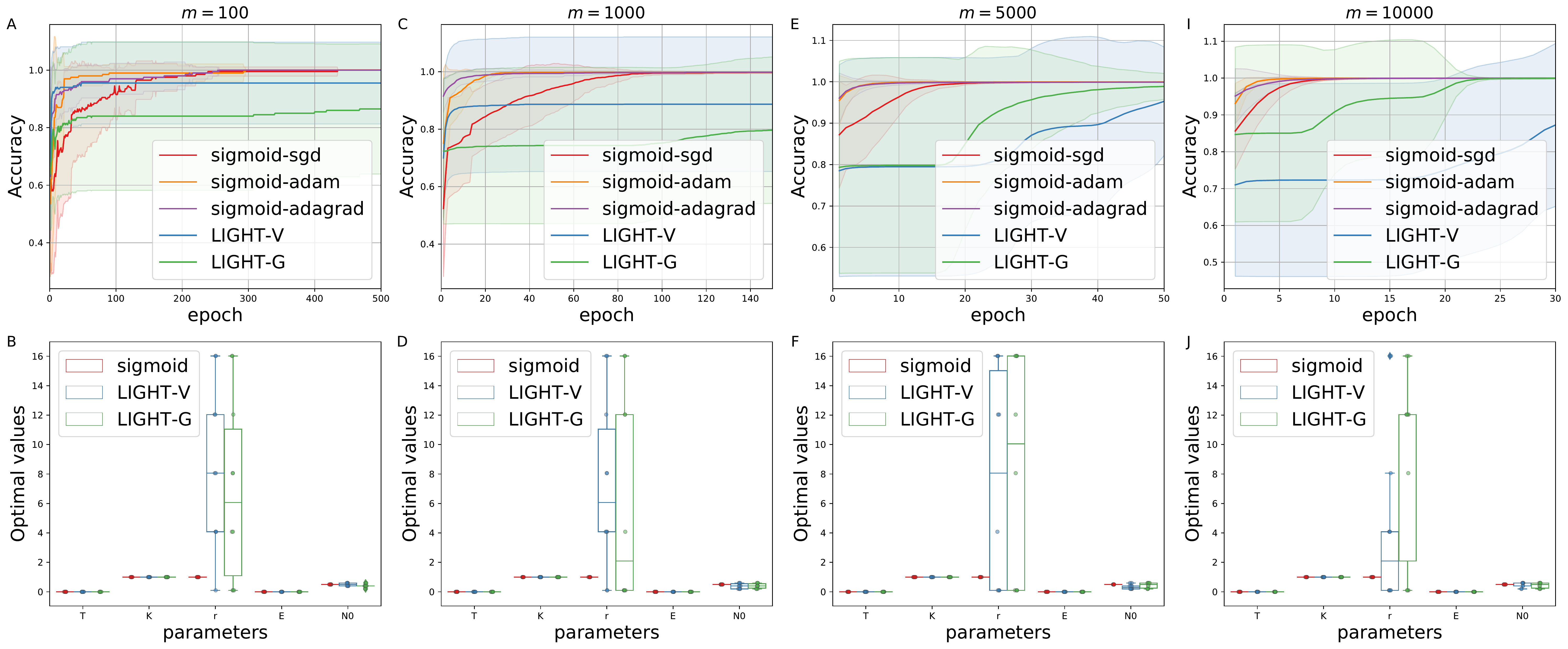}
	\caption{The accuracy curves on testing for $L = 1$, $d_l = 100$, $n = 2$, \emph{cluster std} = 0.25, $h_{\rm{epoch}} = 1$, and $m = \{100, 1000, 5000, 10000\}$: The \textbf{-r-} configuration}
	\label{fig::N-r}
\end{figure}

\begin{figure}[h!]
	\centering
	\includegraphics[width=0.95\columnwidth]{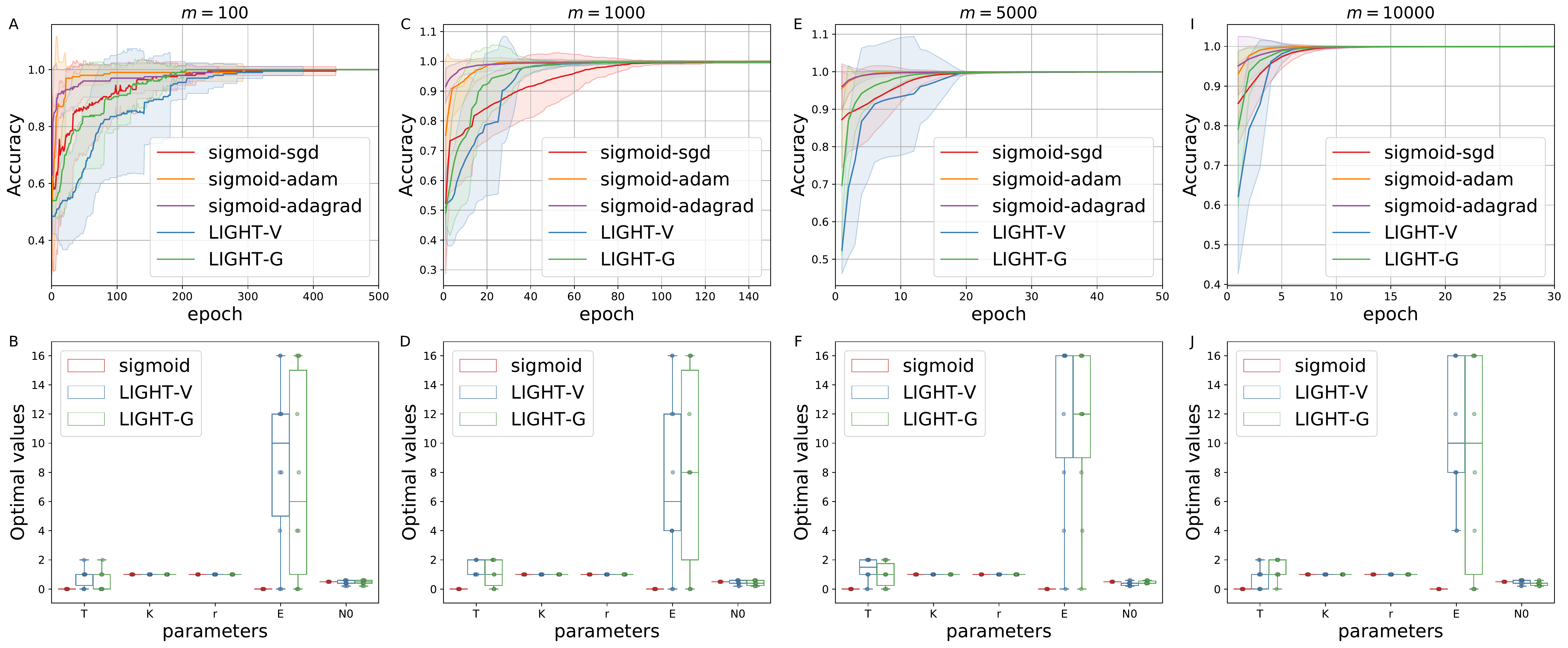}
	\caption{The accuracy curves on testing for $L = 1$, $d_l = 100$, $n = 2$, \emph{cluster std} = 0.25, $h_{\rm{epoch}} = 1$, and $m = \{100, 1000, 5000, 10000\}$: The \textbf{-E-} configuration}
	\label{fig::N-E}
\end{figure}

\begin{figure}[h!]
	\centering
	\includegraphics[width=0.95\columnwidth]{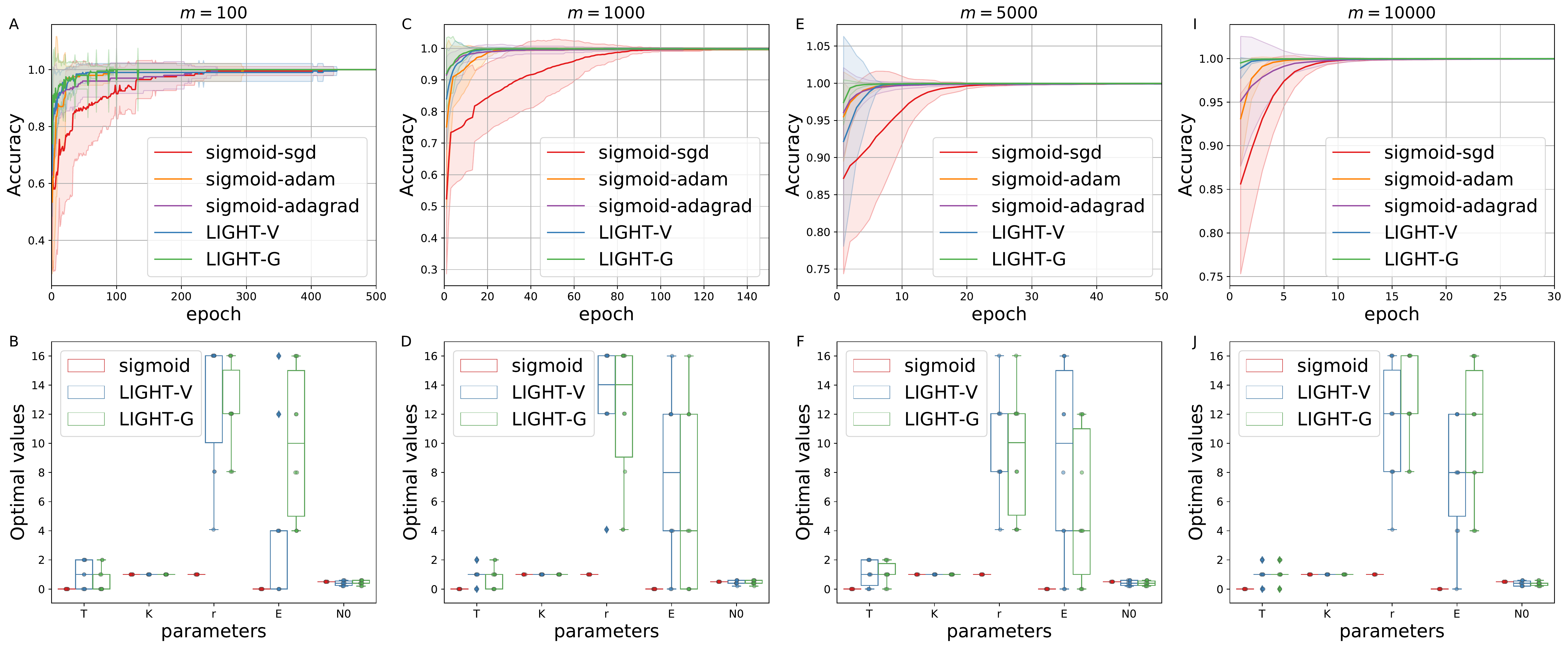}
	\caption{The accuracy curves on testing for $L = 1$, $d_l = 100$, $n = 2$, \emph{cluster std} = 0.25, $h_{\rm{epoch}} = 1$, and $m = \{100, 1000, 5000, 10000\}$: The \textbf{-Er-} configuration}
	\label{fig::N-Er}
\end{figure}

\begin{figure}[h!]
	\centering
	\includegraphics[width=0.95\columnwidth]{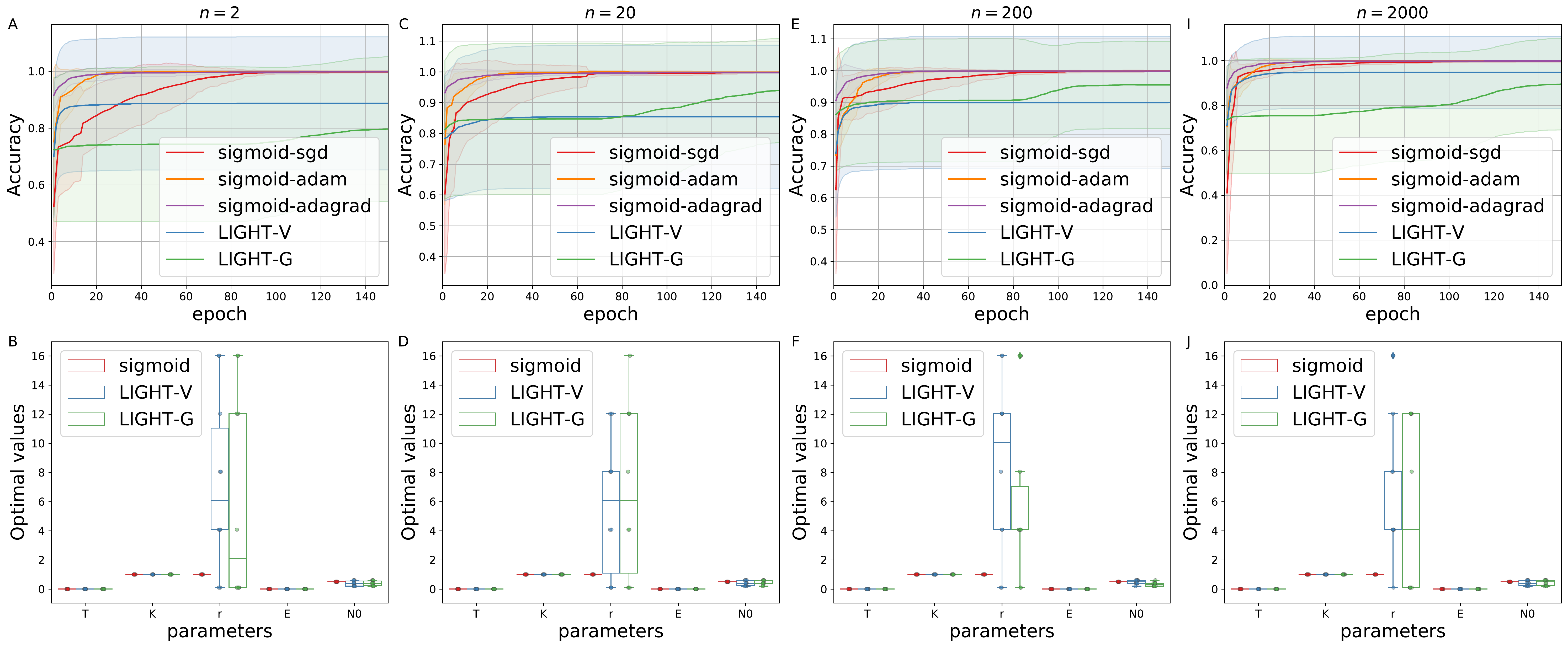}
	\caption{The accuracy curves on testing for $L = 1$, $d_l = 100$, $m = 1000$, \emph{cluster std} = 0.25, $h_{\rm{epoch}} = 1$, and $n = \{2, 20, 200,  2000\}$: The \textbf{-r-} configuration}
	\label{fig::M-r}
\end{figure}

\begin{figure}[h!]
	\centering
	\includegraphics[width=0.95\columnwidth]{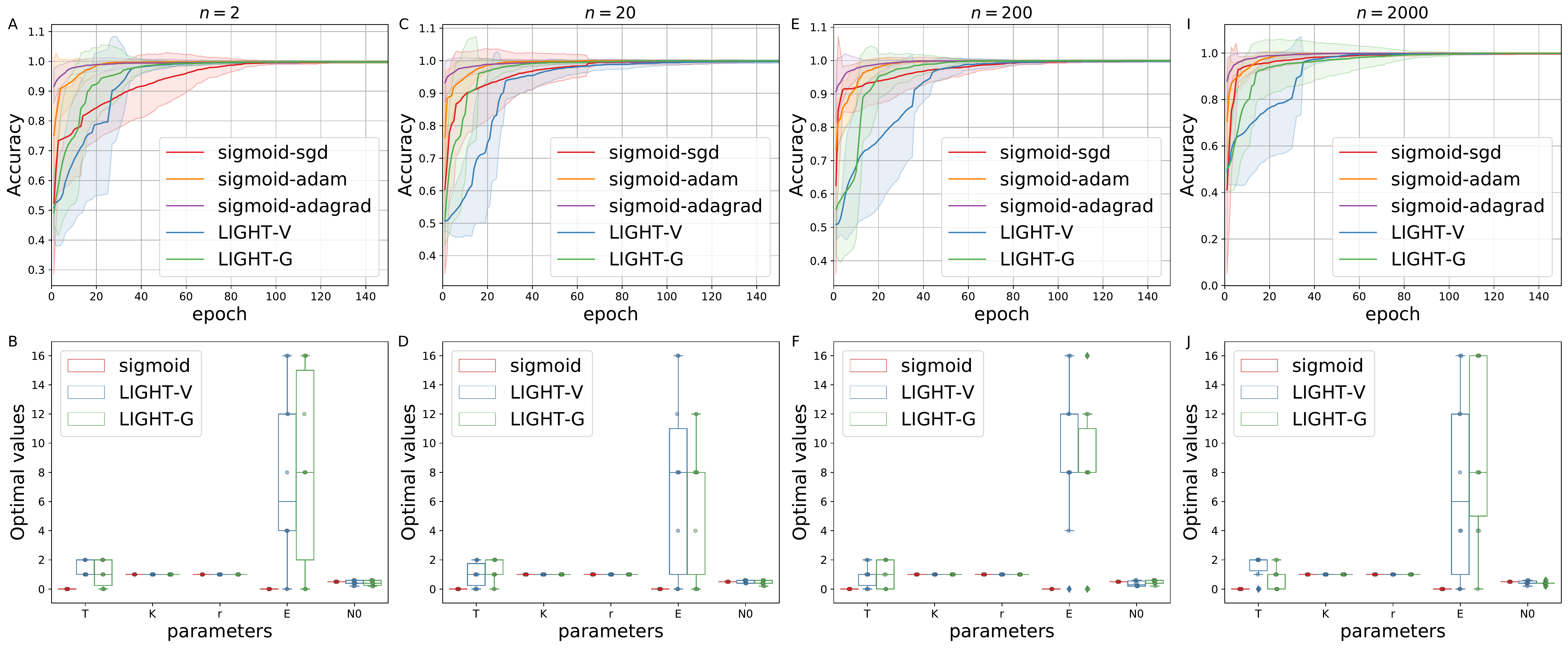}
	\caption{The accuracy curves on testing for $L = 1$, $d_l = 100$, $m = 1000$, \emph{cluster std} = 0.25, $h_{\rm{epoch}} = 1$, and $n = \{2, 20, 200,  2000\}$: The \textbf{-E-} configuration}
	\label{fig::M-E}
\end{figure}

\begin{figure}[h!]
	\centering
	\includegraphics[width=0.95\columnwidth]{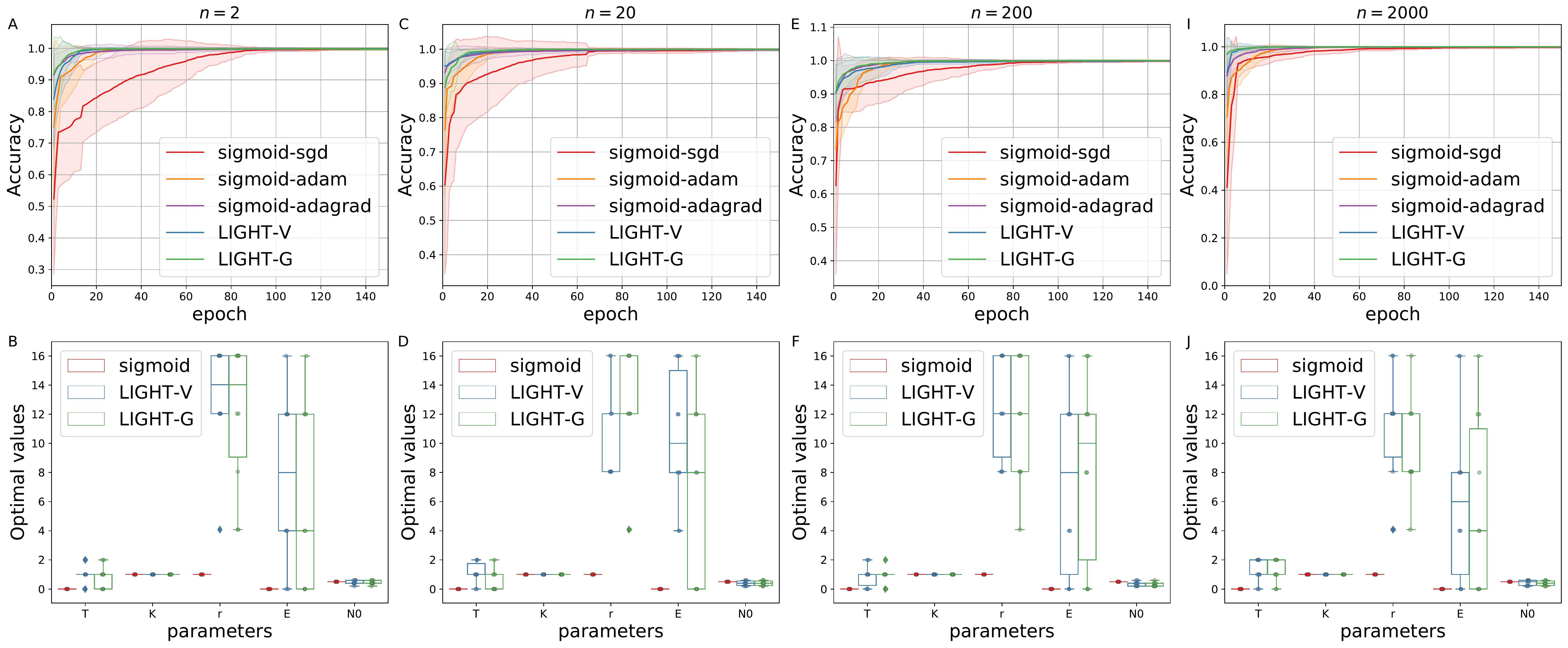}
	\caption{The accuracy curves on testing for $L = 1$, $d_l = 100$, $m = 1000$, \emph{cluster std} = 0.25, $h_{\rm{epoch}} = 1$, and $n = \{2, 20, 200,  2000\}$: The \textbf{-Er-} configuration}
	\label{fig::M-Er}
\end{figure}

\begin{figure}[h!]
	\centering
	\includegraphics[width=0.95\columnwidth]{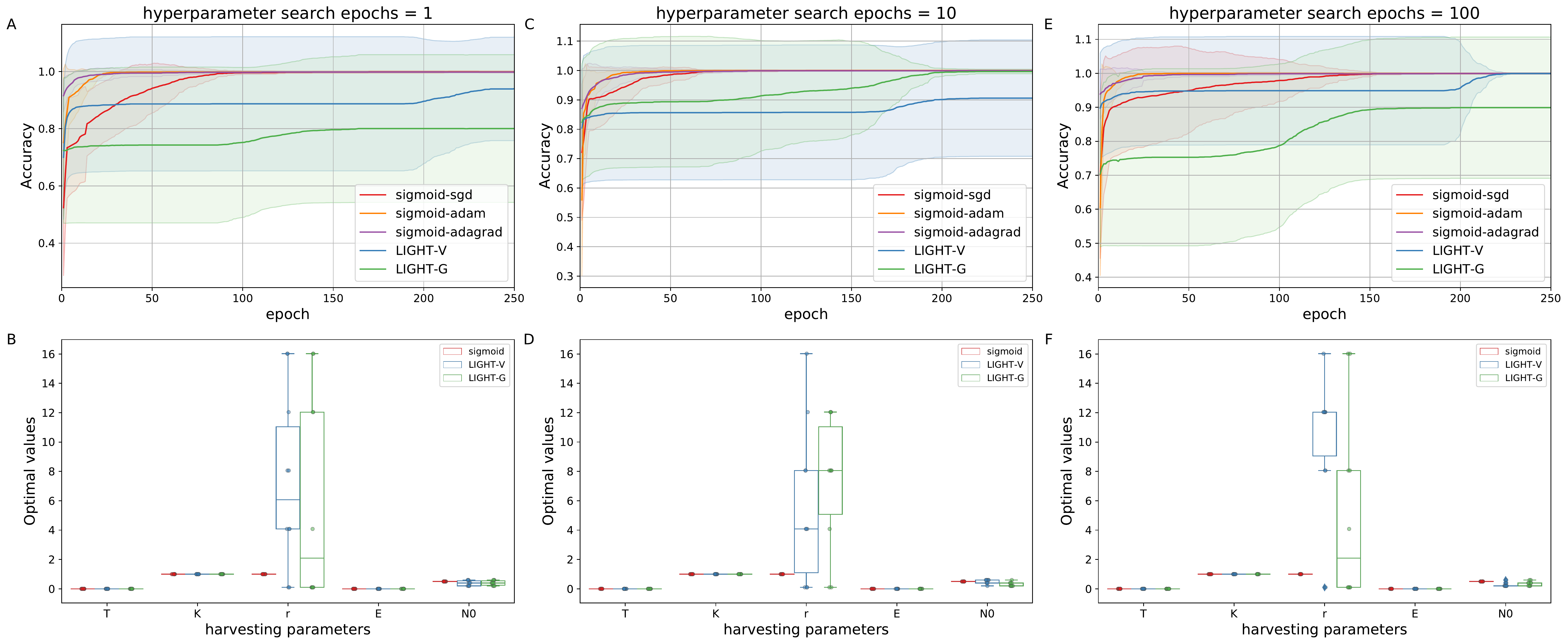}
	\caption{The accuracy curves on testing for $L = 1$, $d_l = 100$, $m = 1000$, $n=2$, \emph{cluster std} = 0.25, and $h_{\rm{epoch}} = \{1, 10, 100\}$: The \textbf{-r-} configuration}
	\label{fig::hyper-r}
\end{figure}

\begin{figure}[h!]
	\centering
	\includegraphics[width=0.95\columnwidth]{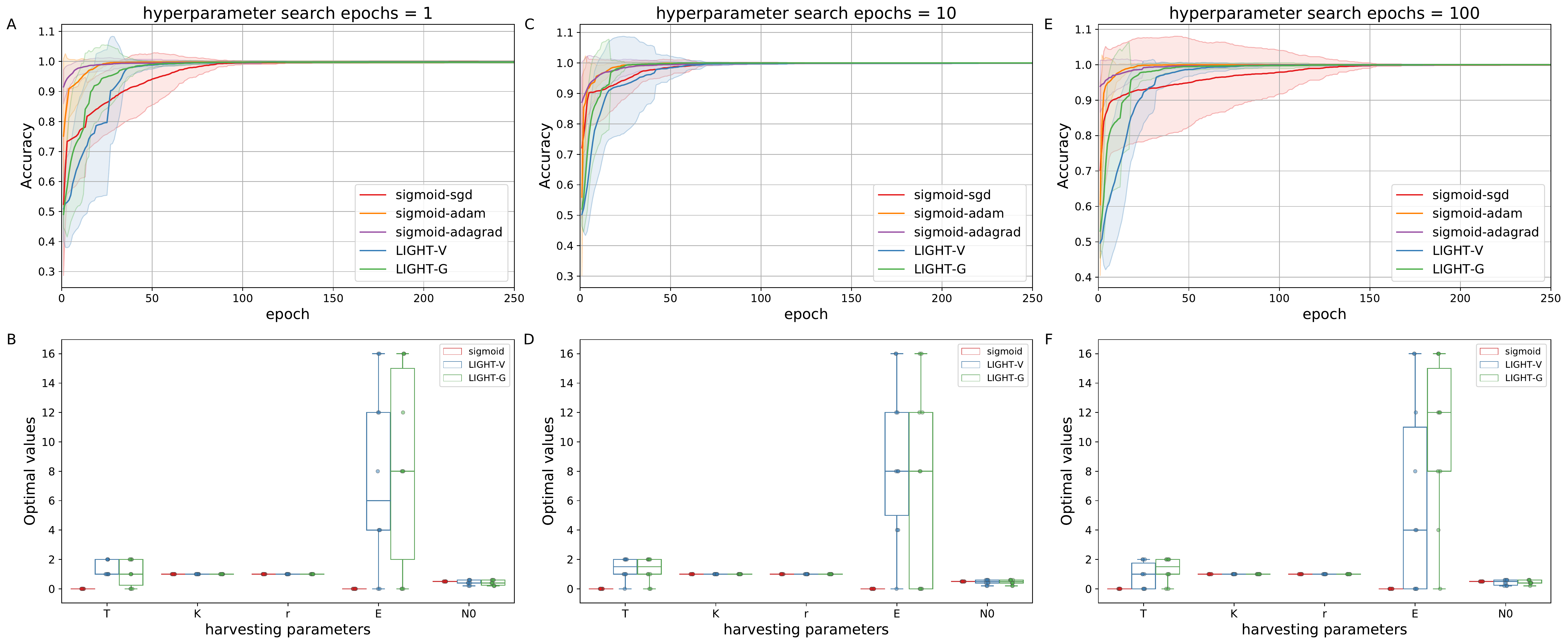}
	\caption{The accuracy curves on testing for $L = 1$, $d_l = 100$, $m = 1000$, $n=2$,  \emph{cluster std} = 0.25, and $h_{\rm{epoch}} = \{1, 10, 100\}$: The \textbf{-E-} configuration}
	\label{fig::hyper-E}
\end{figure}

\begin{figure}[h!]
	\centering
	\includegraphics[width=0.95\columnwidth]{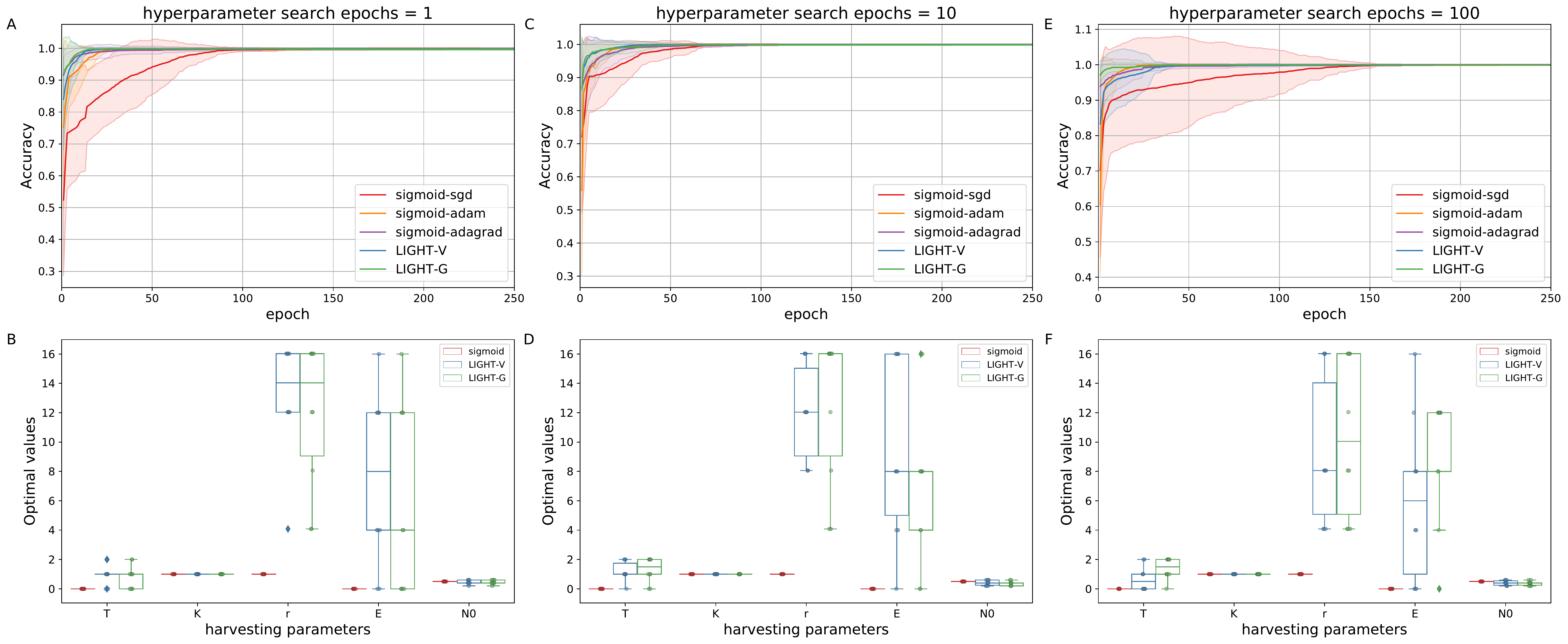}
	\caption{The accuracy curves on testing for $L = 1$, $d_l = 100$, $m = 1000$, $n=2$,  \emph{cluster std} = 0.25, and $h_{\rm{epoch}} = \{1, 10, 100\}$: The \textbf{-Er-} configuration}
	\label{fig::hyper-Er}
\end{figure}

\clearpage
\renewcommand\thefigure{\thesection.\arabic{figure}}  
\renewcommand\thetable{\thesection.\arabic{table}}  
\section{Summary plots}  
\setcounter{figure}{0}
\setcounter{table}{0}
\label{exp::sum}

\begin{figure}[h!]
	\centering
	\includegraphics[width=0.96\columnwidth]{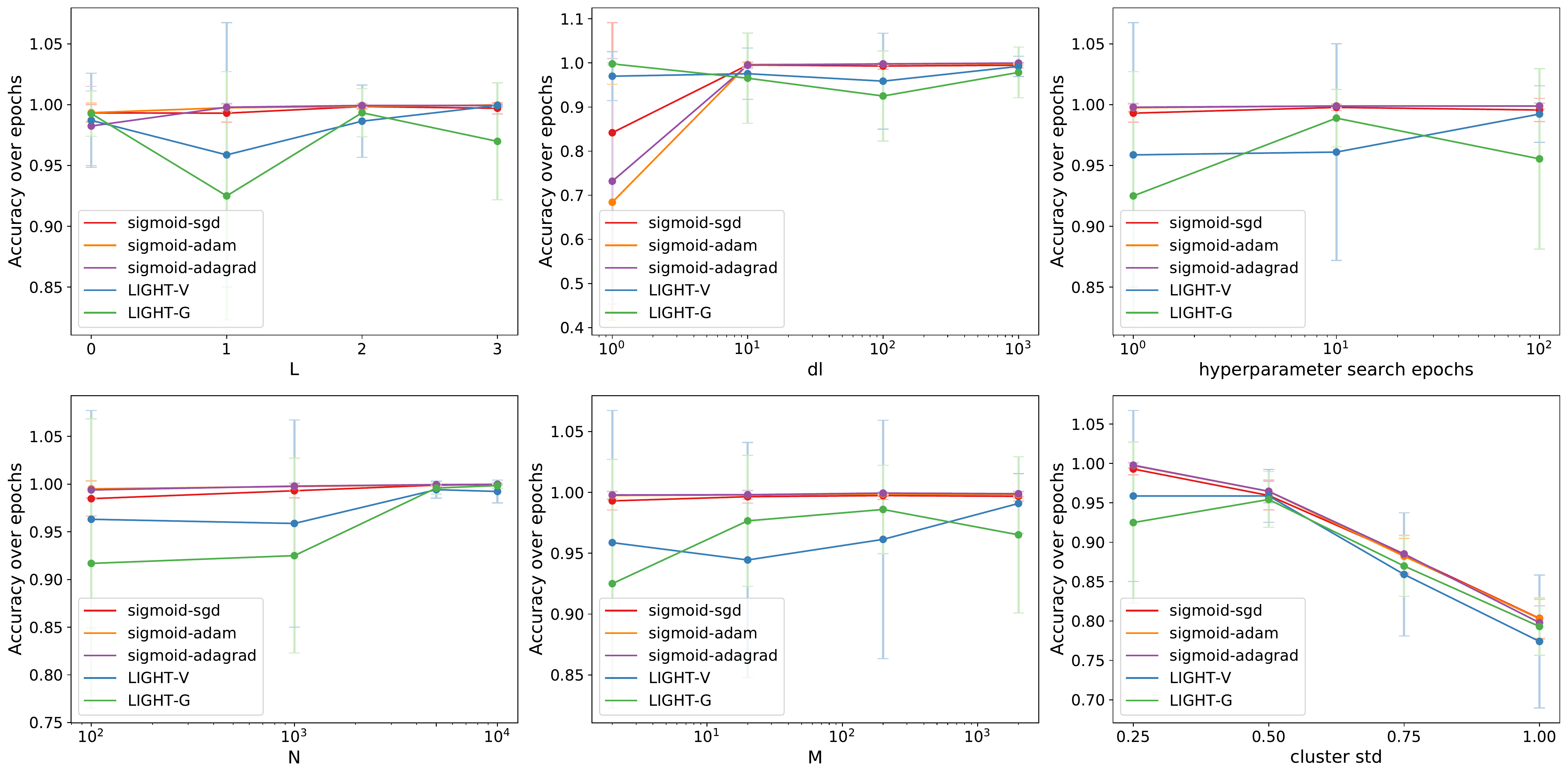}
	\caption{The accuracy curves on testing over epochs: The \textbf{-r-} configuration}
	\label{fig::sum:r}
\end{figure}

\begin{figure}[h!]
	\centering
	\includegraphics[width=0.96\columnwidth]{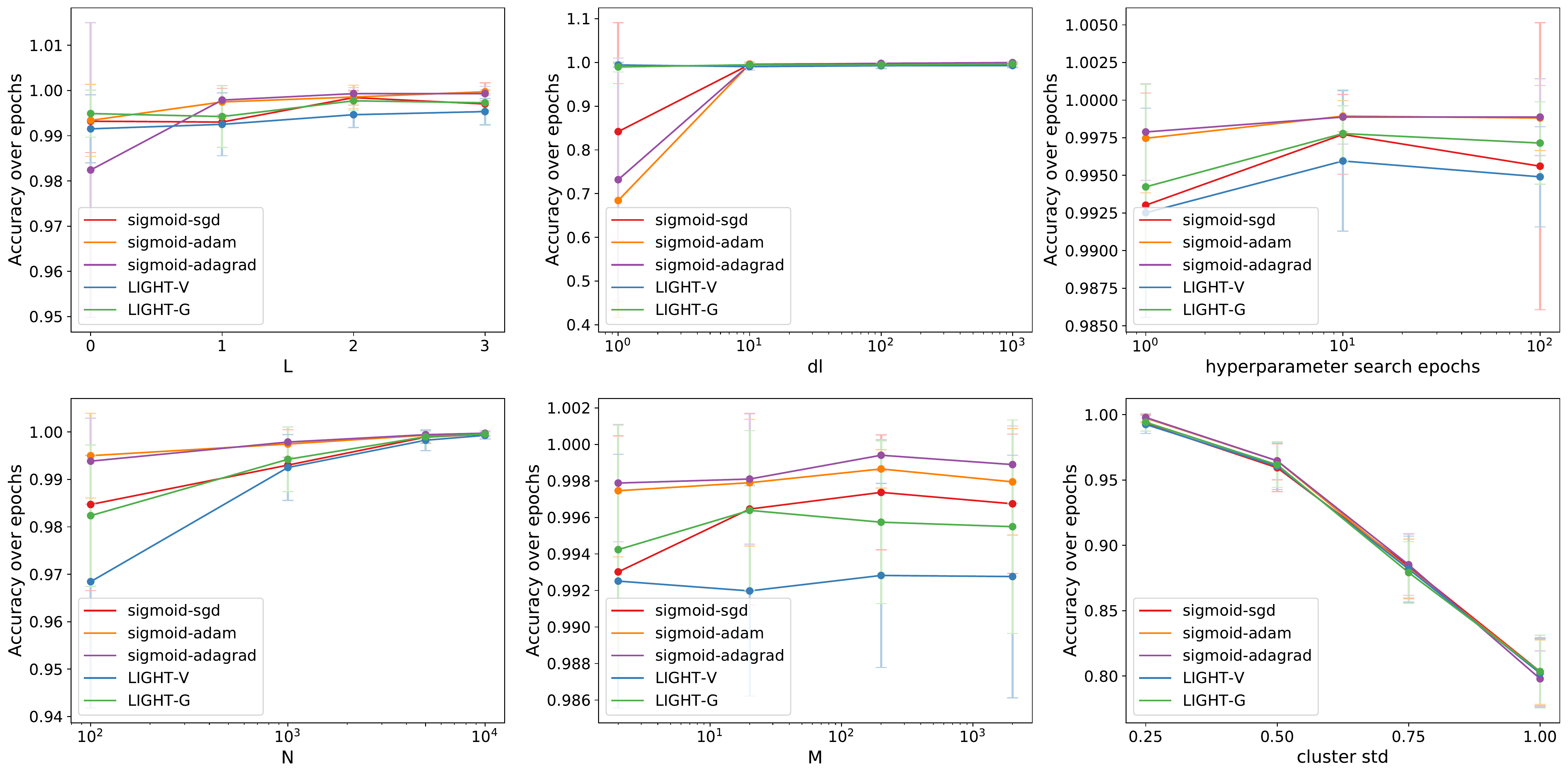}
	\caption{The accuracy curves on testing over epochs: The \textbf{-E-} configuration}
	\label{fig::sum:e}
\end{figure}

We summarized the quantitative  results given in Appendix~\ref{exp::syn} with a set of plots for each configuration, varying  $L = \{0, 1, 2, 3\}$, $d_l = \{1, 10, 100, 1000\}$, $h_{\rm{epoch}} = \{1, 10, 100\}$, $m = \{100, 1000, 5000, 10000\}$, $n = \{2, 20, 200, 2000\}$, and \emph{cluster std} = \{0.25, 0.5, 0.75, 1\} (see Figure \ref{fig::sum:r} and \ref{fig::sum:e}). The fixed values for plotting 2D graphs are $L = 1$, $d_l = 100$, $h_{\rm{epoch}} = 1$, $m = 1000$, $n = 2$, \emph{cluster std} = 0.25.
The summary plot for the -Er- configuration is given in the paper (see Figure~\ref{fig::sum:er}). 

From the summary plots, we can see that only the combination of \emph{per} capita growth and harvesting rates (-Er) is beneficial for optimization processes while inducing each of them separately (-r- and -E-) seems less competitive to both default and adaptive optimizers. We can also observe that the -E- configuration is more influential than the -r- configuration. This can be explained as follows. The -E- configuration regulates inductive biases by squeezing accuracy learning curves to the top while the -r- configuration is responsible for convergence rates and moves the curves to the left (see Figure~\ref{fig:config}).

\clearpage
\renewcommand\thefigure{\thesection.\arabic{figure}}  
\renewcommand\thetable{\thesection.\arabic{table}}  
\section{Experimental datasets}  
\setcounter{figure}{0}
\setcounter{table}{0}
\label{exp::exp}

\begin{wraptable}{r}{7cm}
	\caption{A  brief description of the datasets}
	\label{tab1}
	\begin{center}
		\begin{tabular}{lccc}
			\toprule
			Dataset &$m$ &$n$ \\
			\midrule
			\emph{pima indians} & 768 & 8 \\
			\emph{breast cancer wisc} & 699 & 9 \\
			\emph{heart statlog} & 270 & 13 \\
			\emph{mnist}  & 70000 & 784 \\
			\emph{fashion mnist} & 70000 &  784\\
			\emph{cifar10} & 60000 & 1024  \\
			\bottomrule
		\end{tabular}
	\end{center}
\end{wraptable}

\begin{figure}[h!]
	\centering
	\includegraphics[width=0.95\columnwidth]{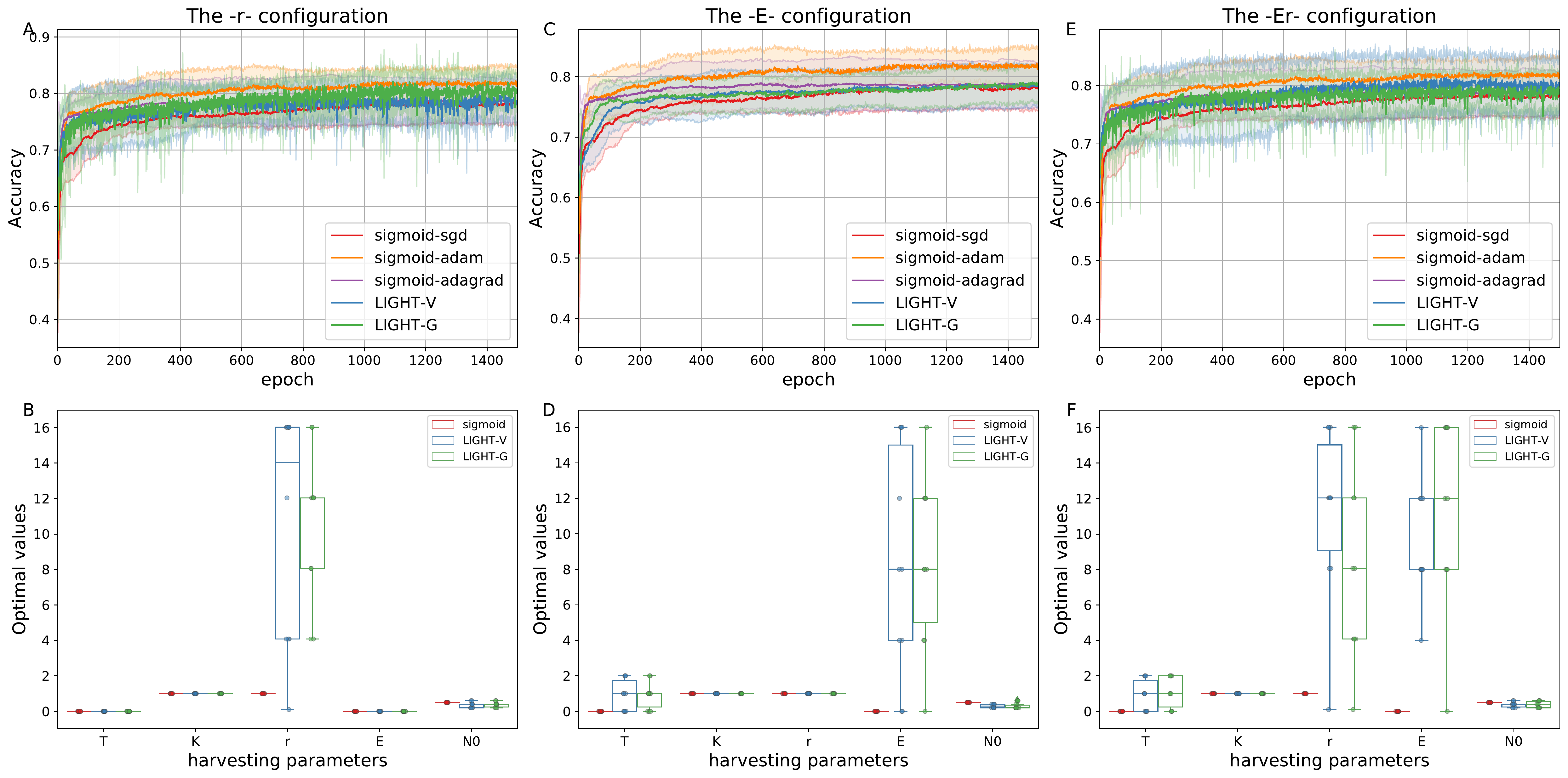}
	\caption{Accuracy on \emph{pima indians} for $L=1$, $d_l = 10$, $m = 768$, $n = 8$, and $h_{\rm{epoch}} = 1$}
	\label{fig::pima}
\end{figure}

\begin{figure}[h!]
	\centering
	\includegraphics[width=0.95\columnwidth]{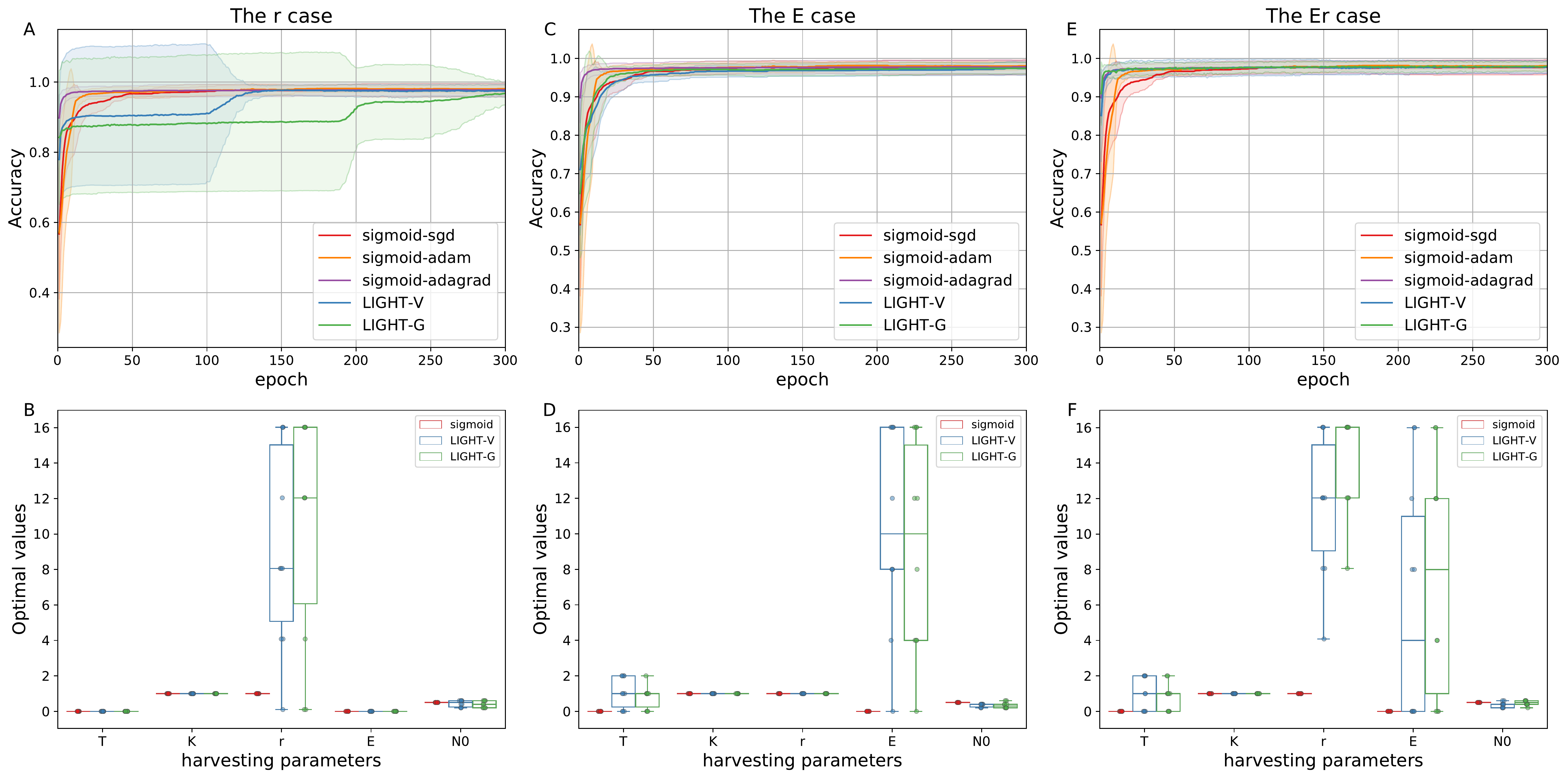}
	\caption{Accuracy on \emph{breast cancer wisc} for $L=1$, $d_l = 10$, $m = 699$, $n = 9$, and $h_{\rm{epoch}} = 1$}
	\label{fig::breast}
\end{figure}

\begin{figure}[h!]
	\centering
	\includegraphics[width=0.95\columnwidth]{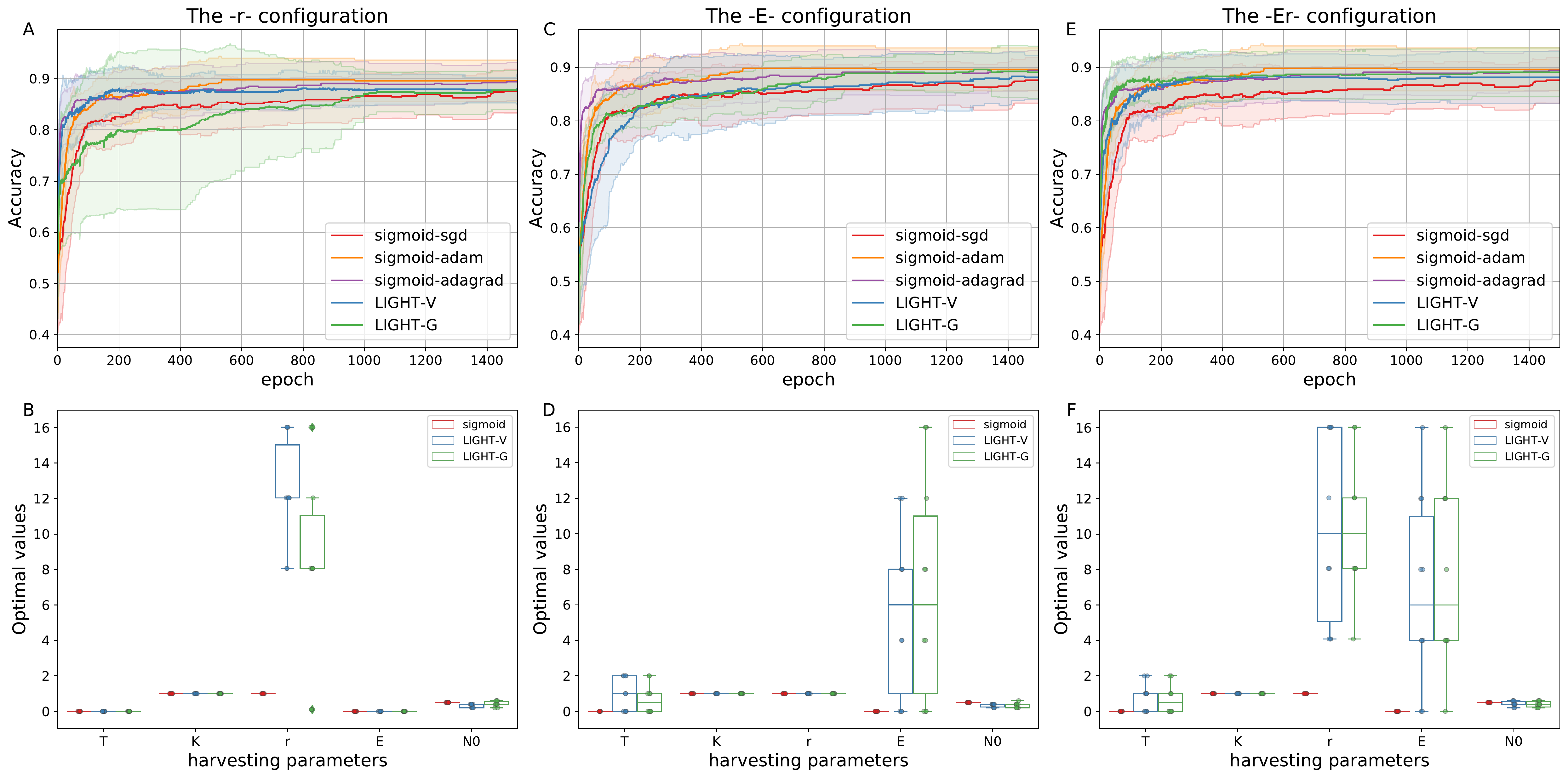}
	\caption{Accuracy on \emph{heart statlog} for $L=1$, $d_l = 10$, $m = 270$, $n = 13$, and $h_{\rm{epoch}} = 1$}
	\label{fig::heart}
\end{figure}

\begin{figure}[h!]
	\centering
	\includegraphics[width=0.95\columnwidth]{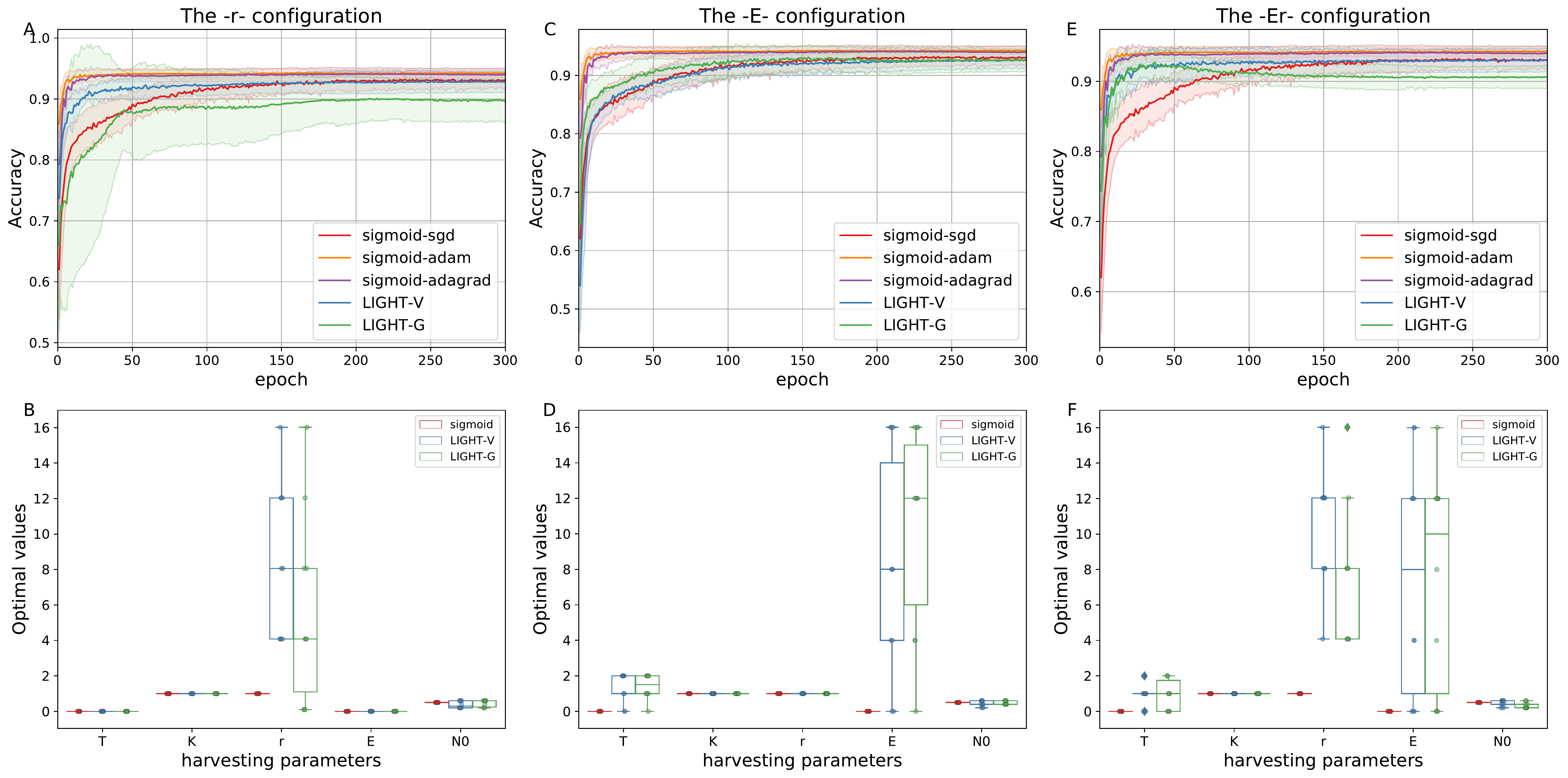}
	\caption{Accuracy on \emph{mnist} for $L=1$, $d_l = 10$, $m = 1000$, $n = 784$, and $h_{\rm{epoch}} = 1$}
	\label{fig::mnist}
\end{figure}

\begin{figure}[h!]
	\centering
	\includegraphics[width=0.95\columnwidth]{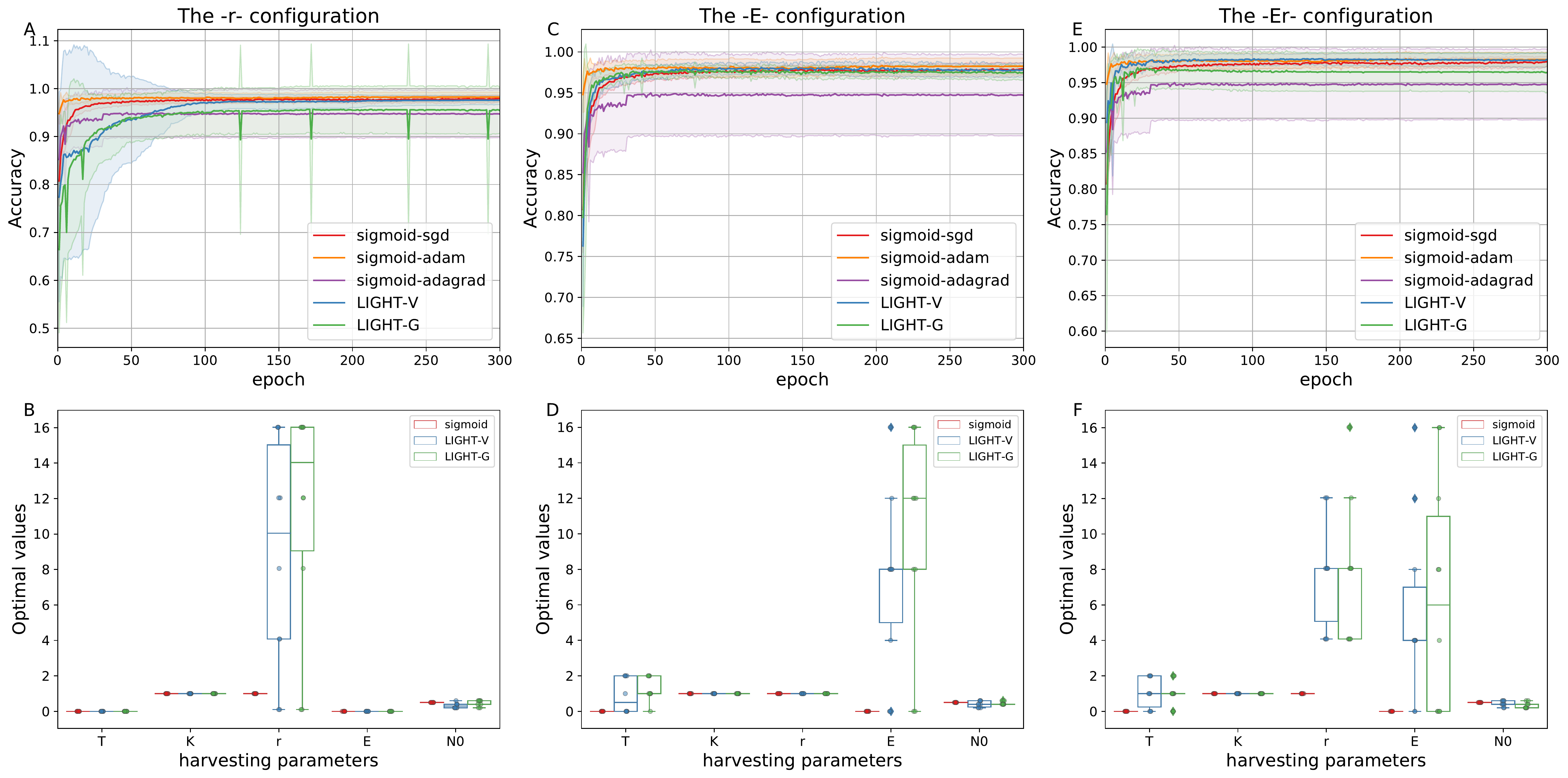}
	\caption{Accuracy on \emph{fashion mnist} for $L=1$, $d_l = 10$, $m = 1000$, $n = 784$, and $h_{\rm{epoch}} = 1$}
	\label{fig::fmnist}
\end{figure}

\begin{figure}[h!]
	\centering
	\includegraphics[width=0.95\columnwidth]{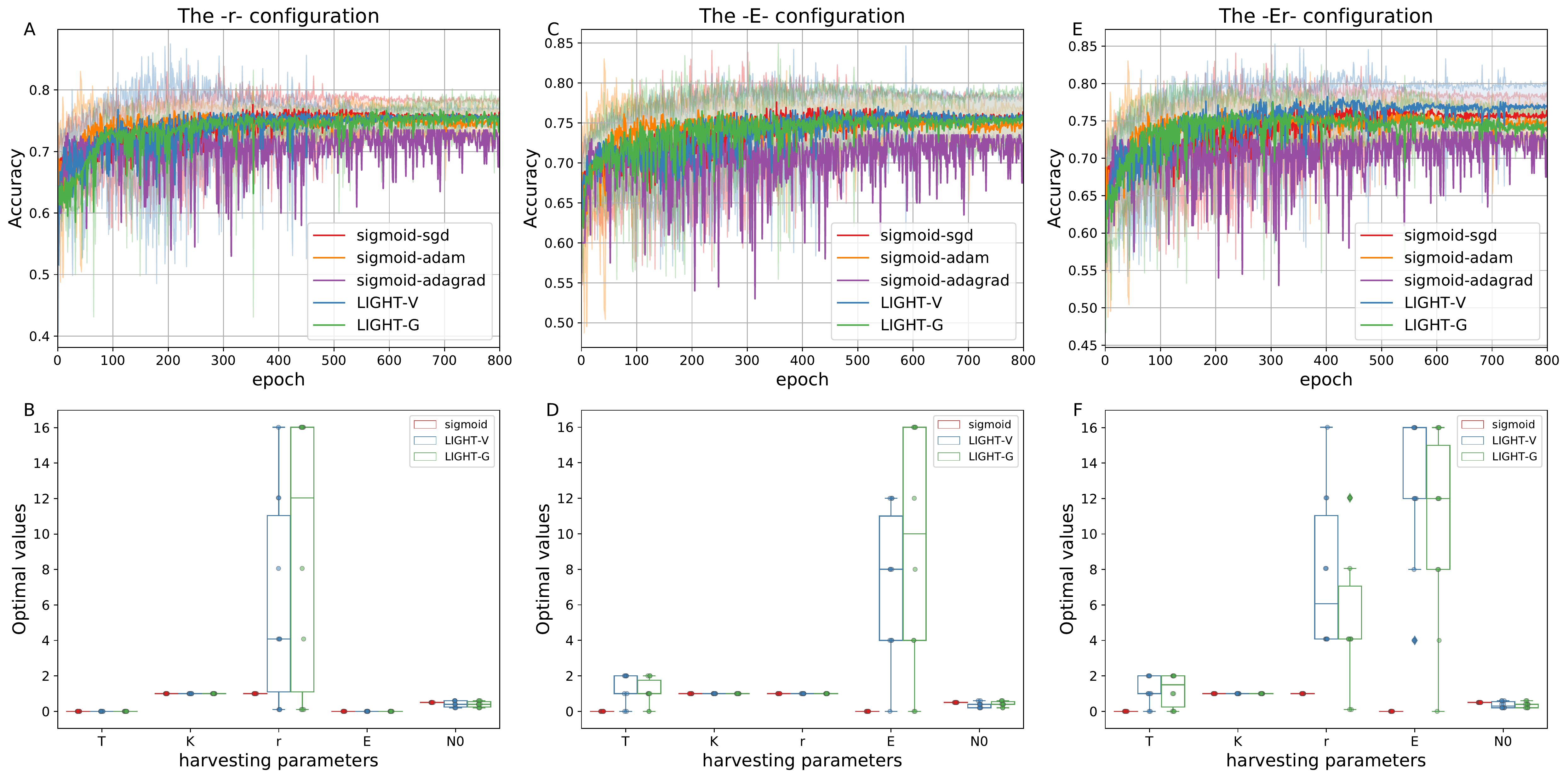}
	\caption{Accuracy on \emph{cifar10} for $L=1$, $d_l = 10$, $m = 1000$, $n = 1024$, and $h_{\rm{epoch}} = 1$}
	\label{fig::cifar10}
\end{figure}

Table \ref{tab1} describes the experimental datasets available from UCI Machine Learning repository with the number of examples $m$ and the number of features $n$. The image datasets (\emph{mnist}, \emph{fashion mnist}, and \emph{cifar10}) were binarized with regard to \eqref{eq::01}. For simplicity, the \emph{cifar10} images were additionally converted into the grayscale. For the last three datasets, we also randomly sampled 1000 examples from the train subset and 200 examples from the test subset for further analysis. The accuracy curves on testing are shown in Figures~\ref{fig::pima}-\ref{fig::cifar10}. Tables~\ref{tab::pima}-\ref{tab::cifar10} report the estimates of the \emph{per} capita growth and harvesting rates. The values biased toward the correct proportion between $r$  and $E$ ($E \leq \frac{r}{2}$ for LIGHT-V and $E \leq  r$ for LIGHT-G) are highlighted in bold.

As can be seen, these values guarantee the highest harvesting rate $H^*$.
In some cases, we can also see a favorable situation when $r < E$ but harvesting still increases the growth rate $r$ that also results in higher $H^*$.
Considering high levels of sd for both $r$ and $E$, we highlighted these values as well. 
For the chosen configuration of the network with $L = 1$, $d_l = 10$, the results can be also in favor of the -E- configuration (see Figure~\ref{fig::sum:e} in comparison with Figure~\ref{fig::sum:r}). We see such results on \emph{fashion mnist} and \emph{cifar10} (see Tables \ref{tab::fmnist}, \ref{tab::cifar10}). This means that the model requires either more deep and wide architectures or higher values of $r$. 

\begin{table}[h!]
	\caption{The estimates of the rates on \emph{pima indians} for $L=1$, $d_l = 10$, $m = 768$, $n = 8$, and $h_{\rm{epoch}} = 1$}
	\label{tab::pima}
	\centering
	\begin{tabular}{>{\rowmac}l>{\rowmac}l>{\rowmac}l>{\rowmac}l>{\rowmac}l>{\rowmac}l>{\rowmac}l<{\clearrow}}
		\toprule
		& & \multicolumn{3}{l}{Optimal values}              
		& \multicolumn{2}{l}{Pre-defined values} \\
		\cmidrule(r){2-7}
		LIGHT & Configuration   & m($r$)$\pm$ sd($r$)   & m($E$)$\pm$ sd($E$)  & $H$ & $E^*$ & $H^*$\\
		\midrule
		-V & -r- & 10.45$\pm 6.55$  & 0.0$\pm 0.0$  &   0 & 5.22 &  2.61\\
		& -E-  & 1.0$\pm 0.0$  & 8.4$\pm 6.38$  & 0.0 &  0.5 & 0.25\\
		& \setrow{\bfseries}-Er- & 11.24$\pm 4.89$  & 9.6$\pm 3.37$  &   1.4 & 5.62 & 2.81\\
		\midrule
		-G & \setrow{\bfseries}-r- & 10.45$\pm 4.28$  & 0.0$\pm 0.0$  &   0.0 & 10.45 & 3.84\\
		& -E-  & 1.0$\pm 0.0$  & 8.4$\pm 4.79$  &  0.0 & 1.0 & 0.37\\
		& -Er- & 8.46$\pm 5.45$  & 11.2$\pm 5.27$  &  2.98 & 8.46 & 3.11\\
		\bottomrule
	\end{tabular}
\end{table}

\begin{table}[h!]
	\caption{The estimate of the rates on \emph{breast cancer wisc} for $L=1$, $d_l = 10$, $m = 699$, $n = 9$, and $h_{\rm{epoch}} = 1$}
	\label{tab::breast}
	\centering
	\begin{tabular}{>{\rowmac}l>{\rowmac}l>{\rowmac}l>{\rowmac}l>{\rowmac}l>{\rowmac}l>{\rowmac}l<{\clearrow}}
		\toprule
		& & \multicolumn{3}{l}{Optimal values}              
		& \multicolumn{2}{l}{Pre-defined values} \\
		\cmidrule(r){2-7}
		LIGHT & Configuration   & m($r$)$\pm$ sd($r$)   & m($E$)$\pm$ sd($E$)  & $H$ & $E^*$ & $H^*$\\
		\midrule
		-V & -r- & 9.25$\pm 5.64$  & 0.0$\pm 0.0$  &  0 & 4.63 &  2.31\\
		& -E-  & 1.0$\pm 0.0$  & 10.4$\pm 5.72$  & 0.0 &  0.5 & 0.25\\
		& \setrow{\bfseries}-Er- & 11.64$\pm 3.96$  & 6.0$\pm 6.86$  &   2.9 & 5.82 & 2.91\\
		\midrule
		-G & - r- & 10.45$\pm 6.55$  & 0.0$\pm 0.0$  &  0.0 & 10.45 & 3.84\\
		& -E-  & 1.0$\pm 0.0$  & 9.2$\pm 5.98$  &  0.0 & 1.0 & 0.37\\
		& \setrow{\bfseries}-Er- & 14.03$\pm 2.81$  & 7.6$\pm 6.65$  &   4.42 & 14.03 & 5.16\\
		\bottomrule
	\end{tabular}
\end{table}

\begin{table}[h!]
	\caption{The estimate of the rates on \emph{heart statlog} for $L=1$, $d_l = 10$, $m = 270$, $n = 13$, and $h_{\rm{epoch}} = 1$}
	\label{tab::heart}
	\centering
	\begin{tabular}{>{\rowmac}l>{\rowmac}l>{\rowmac}l>{\rowmac}l>{\rowmac}l>{\rowmac}l>{\rowmac}l<{\clearrow}}
		\toprule
		& & \multicolumn{3}{l}{Optimal values}              
		& \multicolumn{2}{l}{Pre-defined values} \\
		\cmidrule(r){2-7}
		LIGHT & Configuration   & m($r$)$\pm$ sd($r$)   & m($E$)$\pm$ sd($E$)  & $H$ & $E^*$ & $H^*$\\
		\midrule
		-V & \setrow{\bfseries}-r- & 12.44$\pm 2.94$  & 0.0$\pm 0.0$  &  0 & 6.22 &  3.11\\
		& -E-  & 1.0$\pm 0.0$  & 5.6$\pm 4.7$  &  0.0 &  0.5 & 0.25\\
		& -Er- & 10.45$\pm 5.37$  & 6.8$\pm 5.35$  &   2.37 & 5.22 & 2.61\\
		\midrule
		-G & -r- & 8.46$\pm 5.45$  & 0.0$\pm 0.0$  &  0.0 & 8.46 & 3.11\\
		& -E-  & 1.0$\pm 0.0$  & 6.8$\pm 6.27$  &  0.0 & 1.0 & 0.37\\
		& \setrow{\bfseries}-Er- & 10.45$\pm 3.85$  & 7.6$\pm 5.15$  &   3.67 & 10.45 & 3.84\\
		\bottomrule
	\end{tabular}
\end{table}

\begin{table}[h!]
	\caption{The estimate of the rates on \emph{mnist} for $L=1$, $d_l = 10$, $m = 1000$, $n = 784$, and $h_{\rm{epoch}} = 1$}
	\label{tab::mnist}
	\centering
	\begin{tabular}{>{\rowmac}l>{\rowmac}l>{\rowmac}l>{\rowmac}l>{\rowmac}l>{\rowmac}l>{\rowmac}l<{\clearrow}}
		\toprule
		& & \multicolumn{3}{l}{Optimal values}              
		& \multicolumn{2}{l}{Pre-defined values} \\
		\cmidrule(r){2-7}
		LIGHT & Configuration   & m($r$)$\pm$ sd($r$)   & m($E$)$\pm$ sd($E$)  & $H$ & $E^*$ & $H^*$\\
		\midrule
		-V & -r- & 8.46$\pm 4.38$  & 0.0$\pm 0.0$  &  0 & 4.23 &  2.11\\
		& -E-  & 1.0$\pm 0.0$  & 8$\pm 6.25$  &  0.0 &  0.5 & 0.25\\
		& \setrow{\bfseries}-Er- & 10.45$\pm 3.36$  & 7.6$\pm 6.65$  &   2.07 & 5.22 & 2.61\\
		\midrule
		-G & -r- & 5.67$\pm 5.37$  & 0.0$\pm 0.0$  &  0.0 & 5.67 & 2.09\\
		& -E-  & 1.0$\pm 0.0$  & 10.4$\pm 5.72$  &  0.0 & 1.0 & 0.37\\
		& \setrow{\bfseries}-Er- & 7.66$\pm 3.96$  & 7.6$\pm 6.1$  &   2.82 & 7.66 & 2.82\\
		\bottomrule
	\end{tabular}
\end{table}

\begin{table}[h!]
	\caption{The estimate of the rates on \emph{fashion mnist} for $L=1$, $d_l = 10$, $m = 1000$, $n = 784$, and $h_{\rm{epoch}} = 1$}
	\label{tab::fmnist}
	\centering
	\begin{tabular}{>{\rowmac}l>{\rowmac}l>{\rowmac}l>{\rowmac}l>{\rowmac}l>{\rowmac}l>{\rowmac}l<{\clearrow}}
		\toprule
		& & \multicolumn{3}{l}{Optimal values}              
		& \multicolumn{2}{l}{Pre-defined values} \\
		\cmidrule(r){2-7}
		LIGHT & Configuration   & m($r$)$\pm$ sd($r$)   & m($E$)$\pm$ sd($E$)  & $H$ & $E^*$ & $H^*$\\
		\midrule
		-V & -r- & 8.86$\pm 6.44$  & 0.0$\pm 0.0$  &  0.0 & 4.43 &  2.21\\
		& -E-  & 1.0$\pm 0.0$  & 7.2$\pm 4.92$  & 0.0 &  0.5 & 0.25\\
		& \setrow{\bfseries}-Er- & 7.66$\pm 2.94$  & 5.6$\pm 5.06$  &   1.51 & 3.83 & 1.92\\
		\midrule
		-G & -r- & 11.24$\pm 6.44$  & 0.0$\pm 0.0$  &  0.0 & 11.24 & 4.14\\
		& \setrow{\bfseries}-E-  & 1.0$\pm 0.0$  & 10$\pm 6.04$  &  0.0 & 1.0 & 0.37\\
		& -Er- & 7.66$\pm 3.96$  & 6.4$\pm 6.59$  &  2.78 & 7.66 & 2.82\\
		\bottomrule
	\end{tabular}
\end{table}

\begin{table}[h!]
	\caption{The estimate of the rates on \emph{cifar10} for $L=1$, $d_l = 10$, $m = 1000$, $n = 1024$, and $h_{\rm{epoch}} = 1$}
	\label{tab::cifar10}
	\centering
	\begin{tabular}{>{\rowmac}l>{\rowmac}l>{\rowmac}l>{\rowmac}l>{\rowmac}l>{\rowmac}l>{\rowmac}l<{\clearrow}}
		\toprule
		& & \multicolumn{3}{l}{Optimal values}              
		& \multicolumn{2}{l}{Pre-defined values} \\
		\cmidrule(r){2-7}
		LIGHT & Configuration   & m($r$)$\pm$ sd($r$)   & m($E$)$\pm$ sd($E$)  & $H$ & $E^*$ & $H^*$\\
		\midrule
		-V & -r- & 6.07$\pm 5.7$  & 0.0$\pm 0.0$  &  0.0 & 3.04 &  1.52\\
		& -E-  & 1.0$\pm 0.0$  & 7.2$\pm 4.13$  & 0.0 &  0.5 & 0.25\\
		& \setrow{\bfseries}-Er- & 7.66$\pm 4.38$  & 12.4$\pm 3.98$  &  0.0 & 3.83 & 1.92\\
		\midrule
		-G & -r- & 9.25$\pm 7.52$  & 0.0$\pm 0.0$  &  0.0 & 9.25 & 3.4\\
		& \setrow{\bfseries}-E-  & 1.0$\pm 0.0$  & 9.2$\pm 6.81$  &  0.0 & 1.0 & 0.37\\
		& -Er- & 5.27$\pm 4.22$  & 10.4$\pm 5.4$  &  1.45 & 5.27 & 1.94\\
		\bottomrule
	\end{tabular}
\end{table}

\end{appendices}

\end{document}